%% file: paper.tex
\documentclass{opendatalab}

\usepackage[utf8]{inputenc}
\usepackage{xcolor}
\usepackage[most]{tcolorbox}
\usepackage{longtable}
\usepackage{listings}
\usepackage[numbers]{natbib}
\usepackage{enumitem}
\usepackage{graphicx}
\usepackage{xspace} 
\usepackage{hyperref}
\usepackage{amsmath}
\usepackage{amssymb}
\usepackage{capt-of}
\usepackage{pifont}
\usepackage{wrapfig}
\usepackage{titletoc}

\definecolor{codegreen}{rgb}{0,0.6,0}
\definecolor{codegray}{rgb}{0.5,0.5,0.5}
\definecolor{codepurple}{rgb}{0.58,0,0.82}
\definecolor{backcolour}{rgb}{0.95,0.95,0.92}
\definecolor{promptcolor}{HTML}{D1D0F2}
\definecolor{promptcolorheader}{HTML}{bdbcec}

\newcommand{\github}{\raisebox{-1.5pt}{\includegraphics[height=1.05em]{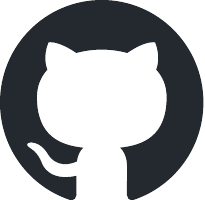}}\xspace}

\newcommand{\huggingface}{\raisebox{-1.5pt}{\includegraphics[height=1.05em]{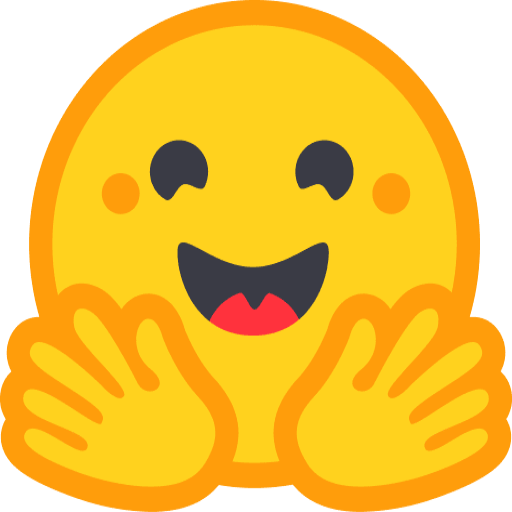}}\xspace}

\definecolor{promptcolor}{HTML}{E3F0FA}
\definecolor{promptcolorheader}{HTML}{B5D6ED}
\definecolor{prompttitletext}{HTML}{1B3A5C}

\newcommand{\xmark}{\ding{55}}
\newcommand{\cmark}{\ding{51}}
\newtcolorbox{promptbox}[1][]{
  enhanced,
  breakable,
  colback=blue!3,
  colframe=blue!35,
  colbacktitle=blue!10,
  coltitle=black,
  fonttitle=\bfseries,
  title={#1},
  boxrule=0.5pt,
  arc=2pt,
  left=5pt,
  right=5pt,
  top=5pt,
  bottom=5pt
}
\lstdefinestyle{promptstyle}{
    backgroundcolor=\color{backcolour},   
    commentstyle=\color{codegreen},
    keywordstyle=\color{magenta},
    numberstyle=\tiny\color{codegray},
    stringstyle=\color{codepurple},
    basicstyle=\ttfamily\footnotesize,
    breakatwhitespace=false,         
    breaklines=true,                 
    captionpos=b,                    
    keepspaces=true,                 
    numbers=left,                    
    numbersep=5pt,                  
    showspaces=false,                
    showstringspaces=false,
    showtabs=false,                  
    tabsize=2
}
\title{PaperFit: Vision-in-the-Loop Typesetting Optimization for Scientific Documents}

\author[1*]{Bihui Yu}
\author[1*]{Xinglong Xu}
\author[1*]{Junjie Jiang}
\author[3*]{Jiabei Cheng}
\author[1]{Caijun Jia}
\author[2]{Siyuan Li}
\author[2]{Conghui He}
\author[1]{Jingxuan Wei}
\author[2]{Cheng Tan}

\affiliation[1]{University of Chinese Academy of Sciences}
\affiliation[2]{Shanghai Artificial Intelligence Laboratory}
\affiliation[3]{School of Automation and Intelligent Sensing, Shanghai Jiao Tong University}

\abstract{
A LaTeX manuscript that compiles without error is not necessarily publication-ready. The resulting PDFs frequently suffer from misplaced floats, overflowing equations, inconsistent table scaling, widow and orphan lines, and poor page balance, forcing authors into repetitive compile-inspect-edit cycles. Rule-based tools are blind to rendered visuals, operating only on source code and log files. Text-only LLMs perform open-loop text editing, unable to predict or verify the two-dimensional layout consequences of their changes. Reliable typesetting optimization therefore requires a visual closed loop with verification after every edit. We formalize this problem as Visual Typesetting Optimization (VTO), the task of transforming a compilable LaTeX paper into a visually polished, page-budget-compliant PDF through iterative visual verification and source-level revision, and introduce a five-category taxonomy of typesetting defects to guide diagnosis. We present PaperFit, a vision-in-the-loop agent that iteratively renders pages, diagnoses defects, and applies constrained repairs. To benchmark VTO, we construct PaperFit-Bench with 200 papers across 10 venue templates and 13 defect types at different difficulty. Extensive experiments show that PaperFit outperforms all baselines by a large margin, establishing that bridging the gap from compilable source to publication-ready PDF requires vision-in-the-loop optimization and that VTO constitutes a critical missing stage in the document automation pipeline.
}

\date{\today}
\correspondence{Cheng Tan, \email{tancheng@pjlab.org.cn}
\\
\\
  \parbox{\linewidth}{\centering
    \github~\href{https://github.com/OpenRaiser/PaperFit}{\textbf{Code}} \quad
    \huggingface~\href{https://huggingface.co/datasets/OpenRaiser/PaperFit}{\textbf{Dataset}}
  }
}

\makeatletter
\def\blfootnote{\xdef\@thefnmark{}\@footnotetext}
\makeatother

\begin{document}

\maketitle
\blfootnote{*Equal contribution.}


\section{Introduction}
\label{sec:intro}

The past decade has witnessed remarkable progress in document automation. Format conversion tools such as Pandoc~\cite{pandoc} enable structural transformation from Word and Markdown to \LaTeX{}. Document understanding models~\cite{blecher_nougat_2023,wang_mineru_2024,datalab_marker_2024} can reconstruct \LaTeX{} source code from PDF files. Recent large language models (LLMs) can generate complete \LaTeX{} document frameworks directly from natural descriptions~\cite{saraiva2025rxiv,yadav_automated_2014}. We refer to this stage collectively as \emph{structural formatting}, whose primary objective is to produce compilable \texttt{.tex} files. However, compilation success does not guarantee visual quality. A syntactically valid \LaTeX{} project may still produce PDFs with misplaced floats, overflowing equations, inconsistent table scaling, widow and orphan lines, and poor page balance~\cite{mittelbach2004latex,knuth1984texbook}. The final page may contain excessive white space that makes the content appear incomplete, or spill into an extra half page that violates strict conference page limits. Currently, resolving these issues relies entirely on manual effort: researchers repeatedly compile the source, inspect the rendered PDF, identify visual defects, adjust the \texttt{.tex} file, and recompile. This compile--inspect--edit cycle, particularly intense in the final hours before submission deadlines, depends almost exclusively on visual judgment that no existing tool fully
automates~\cite{jiang_latte_2025}.

\begin{figure}[t]
    \centering
    \includegraphics[width=\linewidth]{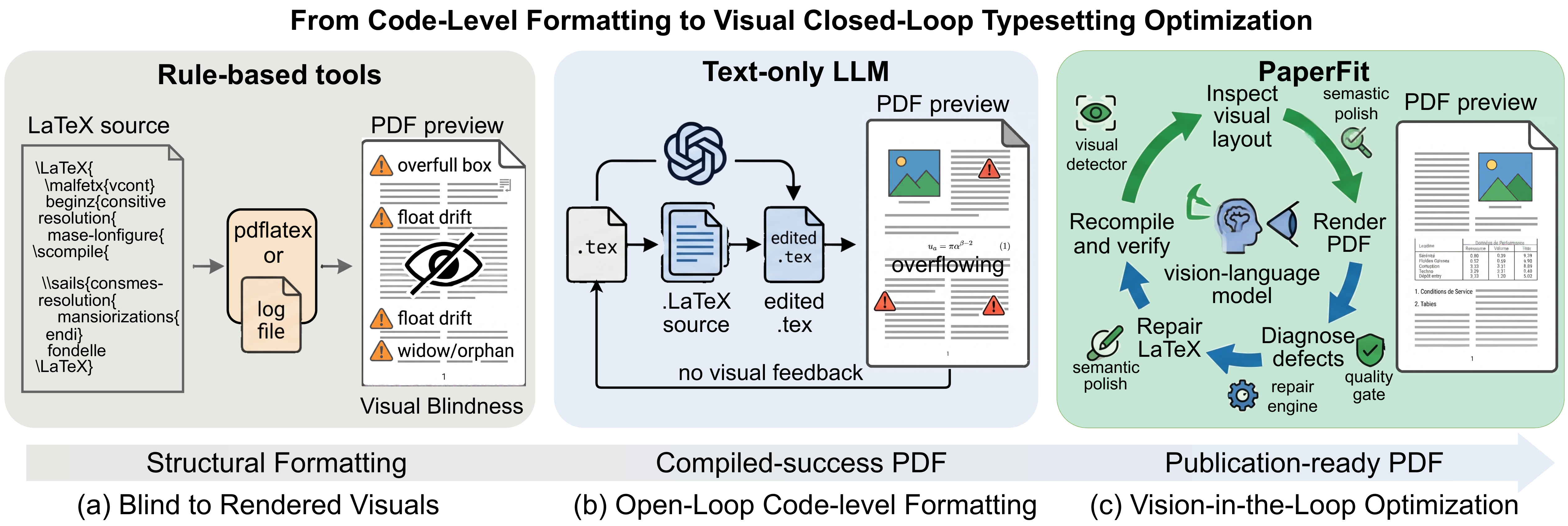}
    \caption{Comparison of typesetting optimization approaches: (a) Rule-based tools are blind to visuals; (b) Text-only LLMs operate in an open loop and cannot predict rendering outcomes; (c) Our PaperFit system establishes a visual closed-loop agent that mimics the iterative human workflow.}
    \label{fig1}
\end{figure}

Existing approaches fail to automate this process due to three fundamental limitations (Figure~\ref{fig1}): \textbf{(i) incomplete observability}. Rule-based tools and compilation logs provide only one-dimensional, code-level signals (Figure~\ref{fig1}a). They can detect overfull hbox warnings but cannot judge whether a minor overflow is visually significant, how figure placement affects reading flow, or how white space is distributed across a page. Typesetting quality is inherently a two-dimensional, spatial judgment that source code and logs alone cannot support. \textbf{(ii) unconstrained repair space}. When a model identifies a problem, it faces an enormous action space in which most options are pseudo-fixes: commands such as \verb|\vspace|, \verb|\resizebox|, and \verb|\newpage| produce compilable output but violate implicit typesetting norms by distorting typography, masking issues, or shifting defects elsewhere. Template files define formatting rules for fonts, margins, and headings, yet encode none of the repair preferences that distinguish a legitimate fix from a cosmetic workaround. \textbf{(iii) unverified cascading effects}. \LaTeX{} edits are highly non-local: a small change in figure width can trigger page-break rearrangements across the entire document. Text-only LLMs operate in an open loop (Figure~\ref{fig1}b), modifying source without rendering or inspecting the result, and thus cannot confirm whether an edit improves or degrades global layout. These challenges characterize typesetting as a closed-loop control problem requiring visual sensing, constrained action, and global verification after every edit.

The advancement of vision-language models (VLMs)~\cite{hurst2024gpt,team2023gemini,yang2025qwen3} has made it feasible to automate this closed loop: a model that can both interpret rendered pages and generate \LaTeX{} modifications can replicate the human compile--inspect--edit workflow. Naively providing page images to a VLM across multiple rounds is insufficient; without structured diagnosis, constrained repair, and gated validation, the model tends to introduce new defects or ignore page-budget constraints~\cite{madaan2023self,shinn2023reflexion}. Based on this insight, we formalize \emph{Visual Typesetting Optimization} (VTO) as the task of transforming a compilable \LaTeX{} paper into a visually polished, page-budget-compliant PDF through iterative visual verification and source-level revision, and introduce a five-category defect taxonomy covering space utilization, float placement, typographic consistency, overflow, and cross-template migration. We position VTO as a critical missing stage between structural formatting and final publication.

We present PaperFit, a vision-in-the-loop agent that closes the sense--act--verify loop for typesetting optimization (Figure~\ref{fig1}c). It addresses the three challenges above through three design components: \emph{multi-source evidence integration} fuses source, log, PDF, and page-image signals into structured defect records, resolving incomplete observability; a \emph{constrained repair policy} explicitly defines permitted operations, forbidden pseudo-fixes, and protected content, taming the unconstrained repair space; and \emph{checklist-gated multi-round validation} recompiles, re-renders, and re-inspects the full document after every edit, catching cascading effects before they propagate.

To benchmark VTO, we construct PaperFit-Bench with 10 venue templates, 200 papers, and 13 defect types at three difficulty levels, and design six baselines that incrementally add capabilities from rule-only to multi-round visual repair. PaperFit achieves perfect compilation and rendering success, the highest visual quality and page-budget compliance, and substantially outperforms all baselines. The most informative comparison is against a naive multi-round visual agent sharing the same page images but lacking structured diagnosis, constrained repair, and gated validation: PaperFit surpasses it by a large margin in both visual quality and page-budget satisfaction, confirming that visual feedback is necessary but not sufficient. These results establish VTO as a critical missing stage in the document automation pipeline and highlight the decisive role of structured visual closed-loop control in producing publication-ready documents.

\section{Related Work}

\subsection{Document Layout Analysis and Automated Formatting}
Recent research in document automation primarily emphasizes structural formatting.
Early foundational work in sequence modeling~\cite{hochreiter_long_1997} and automatic evaluation~\cite{papineni_bleu_2001} established the building blocks for later document understanding systems. VTLayout~\cite{li_vtlayout_2021} represents a significant milestone by improving content block recognition through the integration of deep and shallow visual features with textual information. This integrated approach is further demonstrated by the LayoutLM series~\cite{xu_layoutlm_2019,xu_layoutlm_2020}, DocFormer~\cite{appalaraju_docformer_2021}, and the OCR-free DONUT~\cite{kim2022donut}. More recent efforts have extended document layout analysis to handle complex perturbations~\cite{chen_rodla_2024}, generate diverse large-scale layouts~\cite{noauthor_omnilayout_nodate, kang_omnidoclayout_2025}, and enable global-to-local adaptive perception~\cite{noauthor_doclayout-yolo_nodate}. These models excel at extracting structure from document images, but their output is a recognized layout or reconstructed markup rather than a visually optimized source file. A parallel line of work focuses on generating compilable \LaTeX{} documents from scratch. LLM-driven generators such as Rxiv-Maker~\cite{saraiva_rxiv-maker_2025} produce complete paper frameworks from natural descriptions, cross-lingual formatting systems~\cite{yadav_automated_2014} preserve layout across languages, and agentic writing tools~\cite{lu2024ai, weng2024cycleresearcher} can draft entire manuscripts including \LaTeX{} source. Recent systems such as FlexDoc~\cite{noauthor_flexdoc_nodate} further address document adaptation and compilation efficiency. However, all of these systems treat successful compilation as the terminal goal.

\subsection{Vision-Language Models for Visual Code Editing}
VLMs have significantly improved the mapping of visual signals to code, particularly in extracting structured representations from documents. Nougat \cite{blecher_nougat_2023} demonstrates this advancement by using a Swin Transformer to convert academic PDFs into markup language, thereby bridging the gap between human- and machine-readable formats. The process of converting images to LaTeX is further supported by benchmarks such as Im2Latex-100K \cite{kanervisto_im2latex-100k_2016} and advanced visual reasoning models like $A^2R^2$ \cite{noauthor_2r2_nodate}. Additional tools, including Math2LaTeX \cite{math2latex2025} and Vision-RWKV \cite{duan_vision-rwkv_2024}, have expanded the capabilities for mathematical and structural recognition. Nevertheless, a key limitation persists: most models treat LaTeX as a static translation target. LATTE \cite{jiang_latte_2025} introduced an iterative refinement framework for tables and formulae using visual feedback. Other studies have explored high-fidelity conversion through reinforcement learning for complex table images \cite{ling_table2latex-rl_2025,jayanth_monotone_2015}.

\subsection{Iterative Self-Refinement and Agentic Frameworks}

The development of multi-agent systems has enabled autonomous document optimization through collaborative pipelines. For example, PaperTalker \cite{noauthor_papertalker_nodate} employs a coordinated suite of agents for content parsing, slide generation, and virtual avatar rendering to convert papers into presentation videos. Similar agentic frameworks include Paper2Poster \cite{pang_paper2poster_2025}, which automates academic poster synthesis, and AutoFigure-Edit \cite{lin2026autofigureeditgeneratingeditablescientific}, which generates editable scientific illustrations. LaTeXAgent \cite{eatingchew_eric0801latexagent_2026} provides stateful editing capabilities. Recent studies also examine structured translation via multi-agent coordination \cite{zhu2025latextrans} and domain-specific review feedback \cite{lu_agent_2025}. A persistent challenge is establishing a reliable evaluation-optimization loop. Seeing is Improving (VFLM)~\cite{guo_seeing_2026, guo_visual_2025} uses visual rewards to guide iterative text layout refinement, directly addressing readability issues that are invisible at the code level.
ReLook~\cite{li_relook_2025} applies vision-grounded reinforcement learning to web code generation, and SimpleDoc~\cite{jain_simpledoc_2025} integrates visual verification into multi-modal document understanding.
DocReward~\cite{liu_docreward_2025} proposes learned reward models that score rendered document quality, providing an automated proxy for human visual judgment.

\section{The PaperFit-Benchmark}
\label{sec:bench}
\subsection{Overview}

We introduce PaperFit-Bench, a benchmark for evaluating automated LaTeX layout repair. Unlike existing benchmarks that assess compilation success or content correctness, PaperFit-Bench operationalizes evaluation as visual layout restoration from systematically perturbed sources. Each instance pairs a perturbed LaTeX source with its original compilable version as ground truth, enabling deterministic evaluation across five defect categories (Class A--E) and three difficulty tiers. The benchmark comprises 200 instances spanning 10 venues and both single- and double-column formats.

\subsection{Dataset Construction}
\label{Dataset construction}

\noindent\textbf{Data Collection.}
LaTeX source code of published papers was retrieved from arXiv, covering multiple subfields of artificial intelligence including nature language processing, computer vision, and reinforcement learning. This diversity mitigates evaluation bias toward any single typesetting style. As shown in Table~\ref{tab:benchmark_statistics}, the resulting corpus spans 10 venue templates covering both single-column formats and double-column formats, with page limits ranging from 7 to 14. Each sample contains an average of 6.3 figures and 5.3 tables, providing substantial floating-element density that exercises the full range of layout repair capabilities. This venue diversity ensures that evaluation is not biased toward any single layout style or page constraint.

\input{Tables/Benchmark_statistics}

\noindent\textbf{Preprocessing.}
A standardized compilation test is applied in a controlled build environment; samples that fail compilation or depend on private macro packages are excluded. Appendix sections are uniformly removed. A dual quality-control mechanism combining manual verification ensures that each sample contains at least three figures and at least two tables.

\begin{figure}[ht]
  \centering
  \includegraphics[width=1.0\linewidth]{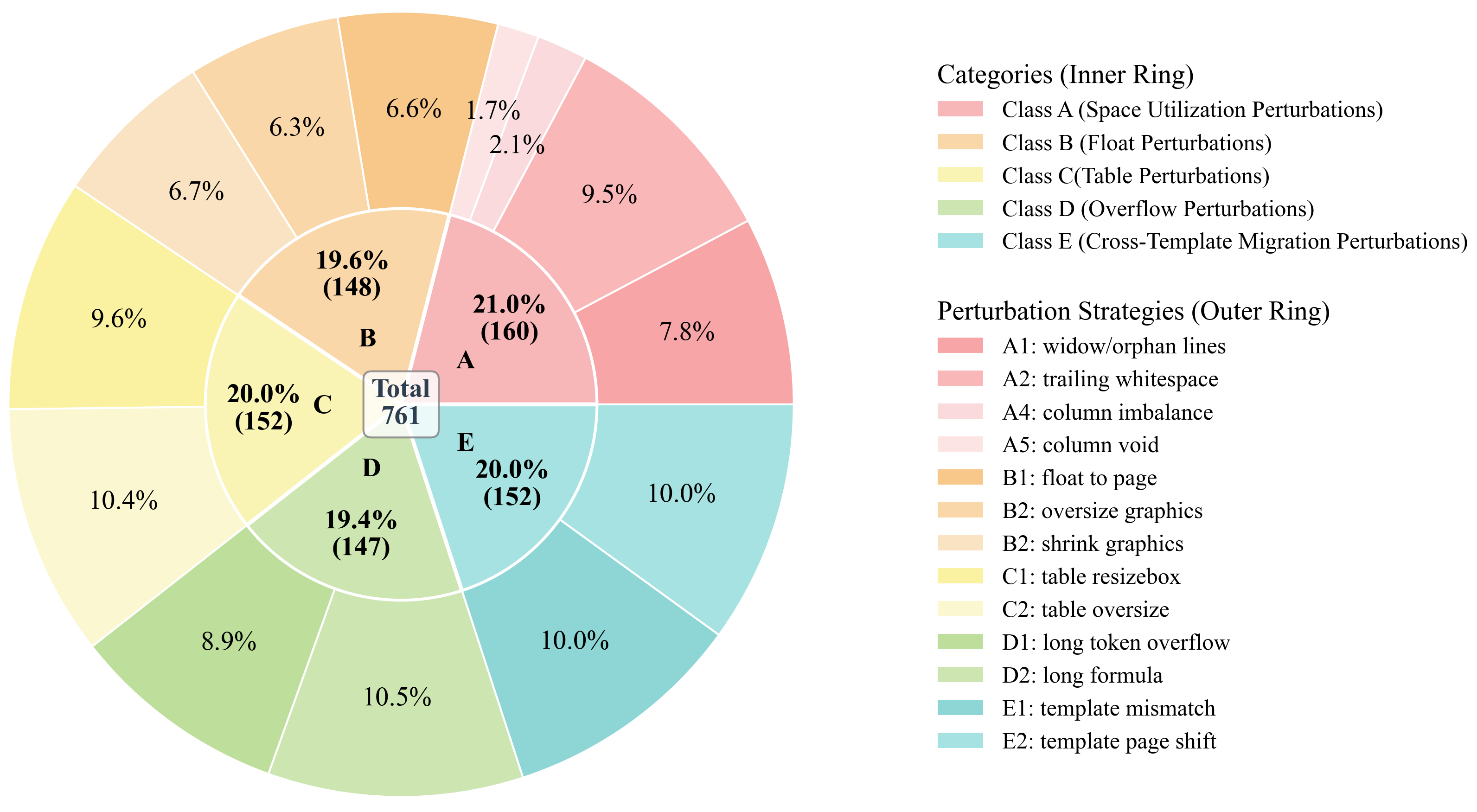}
  \caption{Perturbation distribution and category composition. The inner ring shows proportions of five perturbation categories (Class A--E); the outer ring shows specific perturbation.}
  \label{fig:Perturbation_strategy_double_pie}
\end{figure}

\noindent\textbf{Perturbation Design and Difficulty Tiers.}
We adopt thirteen perturbation strategies organized into five categories aligned with our VTO defect taxonomy (Figure~\ref{fig:Perturbation_strategy_double_pie}): space utilization (Class~A), float placement (Class~B), table width (Class~C), overflow (Class~D), and cross-template migration (Class~E). 

A key design principle of PaperFit-Bench is that it prioritizes realism over simplicity.
PaperFit-Bench is a mixed-disturbance benchmark rather than a collection of one-defect toy examples. Each case is generated from an academic paper project and is associated with a case metadata record and a disturbance manifest. The benchmark contains three difficulty buckets: easy, medium, and hard. These buckets should be interpreted as empirical difficulty groups, not as deterministic recipes. A hard case, for example, may combine template-transfer pressure, table overflow, and page-budget drift, while an easy case may still contain a nontrivial local table or float issue.

These five active disturbance families cover the main visual typesetting optimization failure modes considered in this work.
Space-utilization disturbances create widows, orphans, trailing whitespace, column imbalance, or intra-column voids. Float disturbances move figures or tables away from their natural reading position, shrink graphics, or enlarge graphics beyond the available width. Table disturbances create underutilized or overwide tables. Overflow disturbances introduce long unbreakable tokens or single-line equations that exceed the line width. Template-transfer disturbances create width mismatches or page-budget shifts after changing the surrounding template constraints. A complete listing of perturbation strategies, including their implementation details, validation status, and adoption frequencies, is provided in Table~\ref{tab:perturbation_strategy}.

\input{Tables/Perturbation_strategy}

Beyond defining the perturbation types themselves, our benchmark construction methodology includes an important additional layer of documentation.
The benchmark construction records both the intended perturbation and its concrete source-level realization. This is important because the same high-level defect can appear in different LaTeX forms. For example, an overwide figure may arise from an explicit width larger than \verb|\linewidth|, while a page-budget shift may arise from template transfer together with a text-height change. The evaluation therefore treats the manifest as the source of disturbance intent, and the compile/render outputs as evidence of the actual realized failure.

Each instance is assigned a difficulty tier by the number of co-occurring perturbations: Easy (1--2), Medium (3--4), and Hard (5--8), distributed in a 3:4:3 ratio (Table~\ref{tab:perturbation_difficulty}). Cross-template perturbations (E1, E2) become increasingly prominent in harder instances.

\input{Tables/Perturbation_difficulty}

\noindent\textbf{Assembly and Finalization.}
Perturbed sources are assembled into complete problem instances and undergo final quality verification to ensure compilation succeeds and visual perturbations are realized. The final benchmark contains 200 instances.

Having completed the description of our benchmark construction pipeline, we now compare PaperFit-Bench against representative existing document processing benchmarks to highlight its unique characteristics.

\input{Tables/Benchmark_comparison}

As summarized in Table~\ref{tab:benchmark_comparison}, PaperFit-Bench fills an important gap in the literature. It is the only benchmark that simultaneously supports systematic perturbation injection, visual evaluation based on rendered page outputs, multi-modal evidence integration, and iterative full-document repair workflows—all essential features for evaluating modern AI-powered LaTeX layout optimization agents.

\section{Method}
\label{sec:method}

\subsection{Preliminaries}

Let $x$ denote a compilable \LaTeX{} project, $\tau$ the target template, and $b$ an optional page budget.  Executing the compile-render pipeline produces log evidence $\ell$, a PDF $P$ (upon successful compilation), rendered page images $I$, and a page count $p$.  \emph{Visual Typesetting Optimization} (VTO) seeks a revised source $x^{*}$ that minimizes residual visual defects under hard constraints: 
\begin{equation}
\label{eq:objective}
x^{*} = \arg\min_{x'}\;
  \underbrace{\textstyle\sum_{d \in \mathcal{D}(x')}
    w_{c(d)}\, s(d)}_{\text{visual defect score}}
  \;+\;
  \lambda_e\,\Delta(x, x')
\end{equation}
\vspace{-2pt}
\begin{align}
\text{s.t.}\quad
  & \textsc{Compile}(x', \tau) = \text{success},
    \label{eq:c1}\\
  & \textsc{Render}(x', \tau) = \text{success},
    \label{eq:c2}\\
  & \textsc{Content}(x') \supseteq \textsc{Content}(x),
    \label{eq:c3}\\
  & |\textsc{Pages}(x', \tau)| = b
    \quad\text{(when $b$ is specified)},
    \label{eq:c4}
\end{align}
where $\mathcal{D}(x')$ is the set of visual defects detected in the rendered pages of $x'$ under template $\tau$, each characterized by its category $c(d)$ and severity $s(d)$; $w_{c(d)}$ weights defect categories according to the VTO taxonomy; $\Delta(x, x')$ measures source-level edit distance to encourage minimal, auditable changes; and $\lambda_e$ balances edit conservatism against visual improvement. 

The hard constraints enforce that $x'$ compiles and renders under template $\tau$ (Eqs.~\ref{eq:c1}--\ref{eq:c2}), preserves all scientific content including figures, tables, captions, labels, citations, and bibliography entries (Eq.~\ref{eq:c3}), and meets the page budget when specified (Eq.~\ref{eq:c4}).  Constraints are prioritized in strict order: content preservation $>$ compilation/rendering $>$ page budget $>$ visual quality $>$ edit minimality. Because the objective is observable only after compiling and rendering and because even minor source edits can trigger non-local layout cascades, VTO cannot be solved by single-pass generation. We formulate it as an iterative, evidence-driven search with visual verification after every edit.

\subsection{Sense: Multi-Source Evidence Integration}
\label{sec:method_diagnosis}

No single evidence source reliably captures all typesetting defects. A table may compile without warnings, use a standard \texttt{tabular} environment, and land on the correct page—yet overflow the column boundary. Only the page-image layer reveals this defect; only the source layer can localize the repair target. PaperFit therefore fuses four complementary evidence layers:

\noindent\textbf{Source-layer signals (\texttt{.tex}).}
The source layer provides document structure, template configuration, macro definitions, float environments, table structure, and counts of protected objects such as figures, tables, captions, labels, citations, and bibliography commands. This layer identifies editable regions, safeguards key scientific objects, and reveals structural mismatches resulting from template migration.

\noindent\textbf{Log-layer signals (\texttt{.log}).}
Compilation logs offer deterministic execution evidence, including compile failures, undefined control sequences, unresolved references, missing citations, overfull or underfull warnings, and template-compatibility errors. When the input fails to compile or render, this layer serves as the primary evidence for restoring an executable state.

\noindent\textbf{PDF-layer signals (\texttt{.pdf}).}
The compiled PDF provides document-level outcomes, including final page count, page order, and float landing behavior. This layer helps determine whether the page budget is met and whether floats have drifted far from their first citation.

\noindent\textbf{Page-image-layer signals.}
Rendered pages reveal two-dimensional visual defects that source code or logs cannot reliably detect, such as sparse final pages, double-column \emph{column-void} artifacts, float stacking, oversized tables, local whitespace, cross-page imbalance, and visual inconsistency.

The diagnosis stage converts the collected evidence into structured defect records.
\begin{equation}
d=(c,o,r,e),
\end{equation}
where $c \in \{\mathrm{A}, \mathrm{B}, \mathrm{C}, \mathrm{D}, \mathrm{E}\}$ is the defect category, $o$ is the location (page and spatial region), $r \in \{\text{blocking}, \text{degrading}, \text{cosmetic}\}$ is the severity, and $e$ is the supporting evidence. These records form the interface between diagnosis and repair: every subsequent edit is traceable to explicit multi-source evidence, and the severity field determines repair priority in the next stage.

\begin{figure}[ht]
\centering
\includegraphics[width=\linewidth]{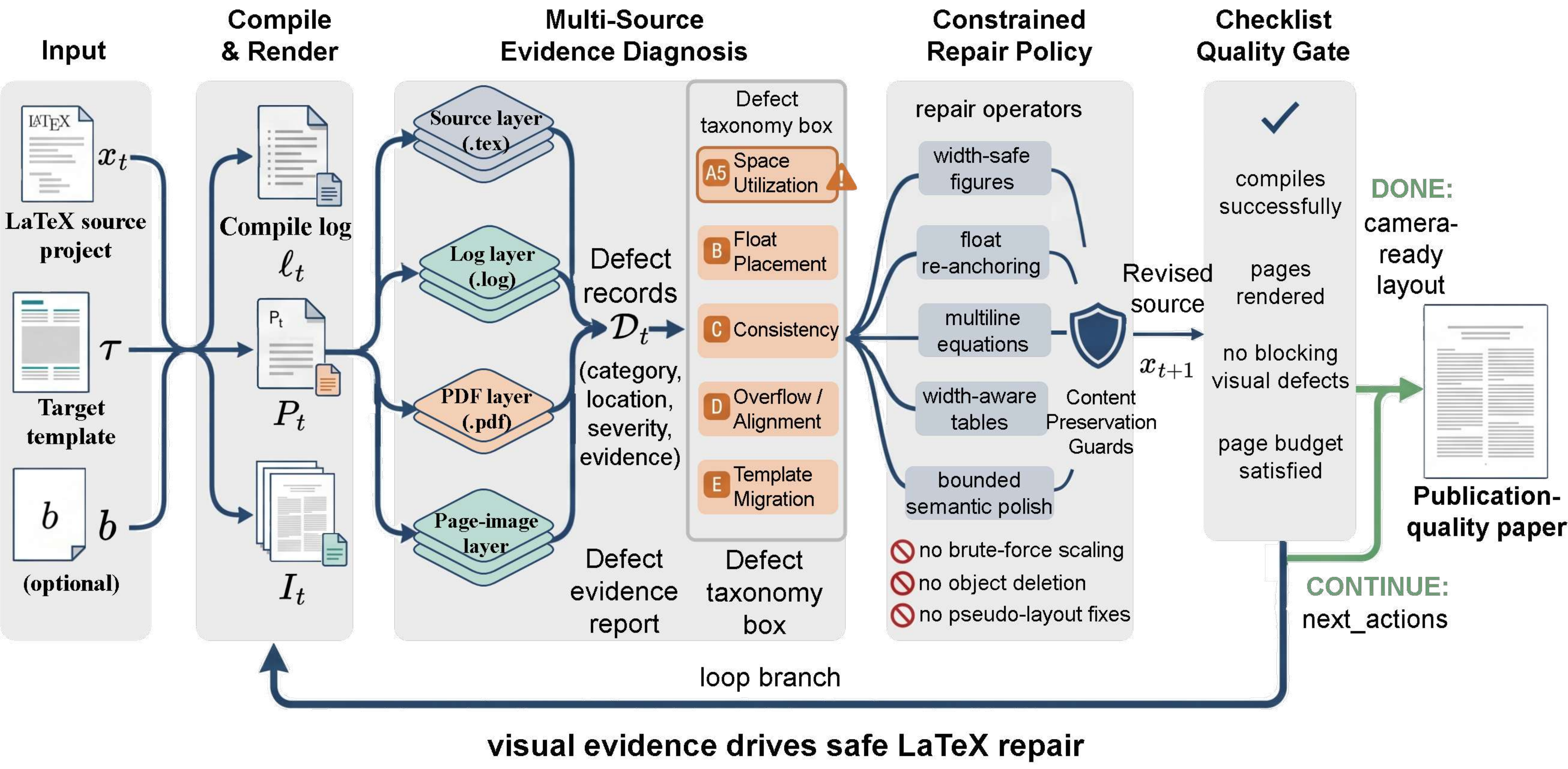}
\vspace{-2mm}
\caption{Overview of the PaperFit pipeline. PaperFit diagnoses layout defects from source, log, PDF, and page-image evidence, applies repairs under a repair preference profile, and validates outputs through a checklist-gated multi-round loop.}
\vspace{-2mm}
\label{fig:pipeline}
\end{figure}

\vspace{-1mm}
\subsection{Act: Constrained Repair Policy}
\vspace{-1mm}
\label{sec:method_repair}

Given the defect set $\mathcal{D}_t$, PaperFit must select repair actions from an enormous space, most of which are pseudo-fixes: technically compilable but typographically harmful. We control this space through a \emph{repair preference profile} $\pi$ that encodes action tiers, defect-category-specific strategies, forbidden operations, and protected content.

\vspace{-2mm}
\subsubsection{Repair Action Tiers}

\label{sec:action_tiers}
We categorize all \LaTeX{} repair actions into three tiers based on their side-effect risk:
\begin{itemize}[leftmargin=*, nosep]
    \item \textbf{Layout-native} (preferred): float re-anchoring (\texttt{[htbp]} parameter adjustment), equation splitting into multiline forms (\texttt{align}, \texttt{multline}), table restructuring with width-aware environments (\texttt{tabularx}, \texttt{table*}), and figure width normalization to template-safe values. These operations address the root cause of the defect without side effects.
    \item \textbf{Spacing-manipulative} (restricted): local \texttt{\textbackslash vspace} adjustment, \texttt{\textbackslash setlength} modification, and column-break hints are permitted only with explicit local justification and must pass re-verification.
    \item \textbf{Pseudo-fix} (forbidden as primary repair): \texttt{\textbackslash resizebox} on tables, \texttt{\textbackslash newpage}/\texttt{\textbackslash pagebreak} for budget control, \texttt{\textbackslash scalebox} for graphics, and content deletion. These commands may temporarily mask a defect but distort typography, violate template norms, or shift defects to other pages.
\end{itemize}

\vspace{-2mm}
\subsubsection{Defect-Aware Repair Selection}

\label{sec:repair_selection}
The repair profile $\pi$ specifies a priority ordering across defect categories and preferred strategies:
\begin{itemize}[leftmargin=*, nosep]
    \item \textbf{Compile errors} (highest): restore compilation via log-guided source repair.
    \item \textbf{Overflow (D)}: split long equations; break unbreakable tokens.
    \item \textbf{Float placement (B)}: re-anchor floats near first citation; normalize figure widths.
    \item \textbf{Table consistency (C)}: replace \texttt{\textbackslash resizebox} with \texttt{tabularx}; restructure overwide tables.
    \item \textbf{Space utilization (A)}: adjust float positions and parameters to eliminate widows/orphans and whitespace.
    \item \textbf{Cross-template (E)}: reconcile width/height mismatches from template migration.
\end{itemize}
\noindent At each round, the system selects the highest-priority unresolved defect from $\mathcal{D}_t$ and applies the top-ranked layout-native strategy for that category. If layout-native options are exhausted, spacing-manipulative actions may be attempted under the restricted policy.

\vspace{-2mm}
\subsubsection{Content Preservation and Semantic Polish Fallback}

\label{sec:content_guards}
Before applying any repair, the system snapshots the count and location of all protected objects (figures, tables, captions, labels, citations, and bibliography entries). After repair, it verifies that no protected object has been deleted, displaced across section boundaries, or had its caption altered. Violations trigger automatic rollback to the pre-repair state.

When layout-native repairs have been exhausted but minor page-budget gaps, widows/orphans, or sparse final pages persist, PaperFit permits bounded \emph{semantic polishing}: minimal wording adjustments (e.g., tightening a verbose sentence, replacing a long phrase with a concise equivalent) that do not alter claims, results, numbers, citations, or factual meaning. This fallback is invoked only after layout-native options fail and remains subject to the content preservation guards. It serves as a last-resort mechanism rather than a primary repair strategy.

\subsection{Verify: Checklist Quality Control}
\label{sec:method_gate}

A single post-repair compilation check cannot ensure global layout because \LaTeX{} edits are highly non-local: a small change in float width can cascade into page-break rearrangements across the entire document. PaperFit recompiles, re-renders, and re-inspects the \emph{complete} document after every edit:
\begin{equation}
\mathcal{S}_t=(x_t,\ell_t,P_t,I_t,\mathcal{D}_t,\mathcal{H}_t,a_t),
\end{equation}
where $x_t$ is the current source, $\ell_t$ is compile-log evidence, $P_t$ is the PDF, $I_t$ is the rendered page set, $\mathcal{D}_t$ is the structured defect report, $\mathcal{H}_t$ is hard-constraint signals, and $a_t$ stores next actions.

Each round follows six steps: (1) compile and collect logs; (2) parse deterministic signals (errors, references, overfull boxes); (3) render all pages; (4) build structured defect records from multi-source evidence; (5) apply constrained repairs per defect category and repair preference profile; (6) recompile/rerender and let the gatekeeper decide. 

The gatekeeper outputs one of three decisions: \textsc{DONE} (all constraints pass, no blocking residual defects), \textsc{CONTINUE} (safe but issues remain), or \textsc{BLOCKED} (repair is unsafe or infeasible). The \textsc{DONE} checklist requires successful compilation, rendering, page-level visual inspection, absence of blocking defects, page-budget satisfaction, and preservation of protected content.

\section{Experiment}
\label{sec:experiment}

\subsection{Experimental Setting}

We evaluate on PaperFit-Bench (Section~\ref{Dataset construction}). Each method receives the same LaTeX project and target page budget; outputs are compiled, rendered, and scored with both programmatic checks and VLM-based visual evaluation.


\subsubsection{Baselines.} 
We compare six baselines with PaperFit, spanning three feedback paradigms. \emph{Rule-based:} Perturbed (unmodified input) and RuleLog (deterministic rule/log repair). \emph{Text-only:} TextST (single-turn source edit) and TextMR (multi-round source-plus-log edit). \emph{Visual:} VisualST (single-turn source-plus-image edit) and VisualMR (multi-round visual agent with fixed rounds). 

These baselines are systematically constructed to isolate the incremental value of each core capability in the visual typesetting optimization pipeline. They differ only in the evidence sources they can access, the number of repair iterations allowed, and whether they incorporate PaperFit's structured repair machinery:

\paragraph{Perturbed: perturbed input.}
Perturbed is the unmodified disturbed project. It measures the raw difficulty of the benchmark after perturbation and provides the visual reference for VLM pairwise comparison whenever the perturbed input can be rendered.

\paragraph{RuleLog: deterministic rule/log repair.}
RuleLog applies deterministic repair rules driven by source and compile-log signals. It does not use model-based visual feedback. Its role is to test how far simple execution and log repair can go without page-level visual evidence.

\paragraph{TextST: single-turn text-only model repair.}
TextST sends the LaTeX source to a model in a single repair turn. It does not inspect rendered page images. This baseline tests whether source-only reasoning can repair layout defects without observing the final pages.

\paragraph{TextMR: multi-round text/log repair.}
TextMR extends TextST with multiple text/log feedback rounds. It can react to compilation errors and logs across rounds, but it still does not use page images as visual evidence.

\paragraph{VisualST: single-turn visual repair.}
VisualST receives the LaTeX source and rendered page images, then performs one visual repair turn. If compilation or rendering fails, the failure is accounted for by the same evaluation protocol rather than being removed from the denominator. VisualST isolates the value and limitation of adding page images without closed-loop visual iteration.

\paragraph{VisualMR: naive multi-round visual agent baseline.}
VisualMR is a fixed-round visual agent baseline. It can inspect source, logs, and page images over a small fixed number of rounds and can directly repair compile, render, and layout issues. It does not use PaperFit's defect taxonomy, structured diagnosis records, constrained repair policy, repair-plan artifacts, rollback-aware gatekeeper, or PaperFit runtime. This makes VisualMR the closest baseline for testing whether multi-round visual feedback alone is sufficient.

\paragraph{PaperFit.}
PaperFit uses the same basic input project and target page budget, but adds structured multi-source diagnosis, a repair preference profile, and checklist-gated validation. The distinction between VisualMR and PaperFit is therefore not whether a model sees page images, but whether the multi-round process is organized around explicit defects, constrained repairs, and acceptance gates.


\subsubsection{Evaluation protocol.}
We evaluate all methods using a dual-metric framework that combines programmatic correctness checks and human-aligned visual assessment, ensuring outputs are both technically valid and publication-ready. We report four primary binary metrics: compile success, render success, \emph{Page hit} (exact page-budget match), and \emph{Win rate} (fraction of cases judged visually better than the Perturbed baseline).

For quantitative composite evaluation, we use two complementary scores:
- \emph{Program}: A 0--5 composite of non-visual execution reliability and content fidelity
- \emph{VLM}: A gated 0--5 visual quality score based on rendered page assessment

We report both scores because Visual Typesetting Optimization (VTO) requires outputs to simultaneously satisfy hard technical constraints and subjective visual quality standards. A method that produces visually appealing pages but fails to preserve content or meet page budgets is not acceptable for publication, just as a technically correct but visually defective document fails to meet the core goal of typesetting optimization.

\paragraph{Program Score.} 
The Program score summarizes non-visual execution and fidelity signals on a 0--5 scale, computed as the average of five equally weighted dimensions, each normalized to $[0,1]$:
\[
\mathrm{Program}=5\cdot\frac{1}{5}\sum_{k=1}^{5}s_k.
\]
The five dimensions are:
\begin{itemize}[leftmargin=1.5em, nosep]
    \item \texttt{compile\_reliability}: Whether the candidate compiles and renders into usable pages
    \item \texttt{content\_integrity}: Whether protected scientific objects (figures, tables, captions, labels, citations, bibliography entries) are fully preserved
    \item \texttt{reference\_quality}: Whether all references resolve correctly and no severe log errors remain
    \item \texttt{page\_precision}: Whether the output satisfies the target page budget, with a penalty for excessive source rewriting
    \item \texttt{content\_embedding\_similarity}: Semantic similarity between the original and final LaTeX sources
\end{itemize}

Program is intentionally not a visual score. A method can receive a high Program score by producing a compilable, faithful, page-controlled document even if its final layout still contains visible whitespace or float-quality issues. Conversely, an output that looks acceptable in a rendered screenshot will be penalized by Program if it loses protected content, violates page budget, or leaves unresolved references.

\paragraph{VLM Visual Score.} 
The visual evaluation uses rendered page images and produces a gated 0--5 score. It operates in two modes:
1. Pairwise comparison mode: When the Perturbed baseline renders successfully, the evaluator compares the perturbed input and candidate output side-by-side
2. Render-rescue mode: When the Perturbed baseline cannot be rendered, a renderable candidate receives credit for recovering from a non-renderable state

The raw VLM score combines three weighted components:
\[
\mathrm{VLM}_{\mathrm{raw}}=
0.35\,S_{\mathrm{abs}}+
0.40\,S_{\mathrm{repair}}+
0.25\,S_{\mathrm{final}}.
\]
Here:
- $S_{\mathrm{abs}}$ measures absolute repair-oriented quality, including defect resolution, constraint alignment, visual quality, new-defect avoidance, and publication readiness
- $S_{\mathrm{repair}}$ measures pairwise repair quality relative to Perturbed (when renderable) or render-rescue quality (when Perturbed is not renderable)
- $S_{\mathrm{final}}$ measures final-paper aesthetics, including professionalism, space utilization, float placement, typographic consistency, and visual balance

The final reported VLM score applies strict constraint gates to the raw score:
- Non-renderable candidates are capped at the minimum score
- Compile-dirty but renderable outputs are penalized and capped
- Page-budget failure, unresolved references, or major newly introduced visual defects also trigger score capping

The \emph{Win rate} reported in the main results is the fraction of cases in which a method is judged better than the Perturbed baseline under this full visual evaluation protocol.

\subsection{Main Quantitative Results}

\input{Tables/Exp_main_results}

\textbf{Neither text/log feedback nor single-turn visual feedback is sufficient.} RuleLog, TextST, TextMR, and VisualST represent progressively richer feedback signals, from compile logs to multi-round text to rendered page images. Yet none exceeds a VLM score of 2.19 or a Win rate of 0.43 (Table~\ref{tab:exp_main_results}). Text/log methods cannot judge two-dimensional layout failures such as excessive white space or float cascades, while single-turn visual editing often fails to handle non-local cascades.

\textbf{Naive multi-round visual repair improves usability but remains weak on page control.} VisualMR reaches 0.975 on both compile and render success, confirming that multi-round visual and log feedback can remove most execution failures. 

However, its Page hit is only 0.549 and its Win rate is 0.650. Without explicit planning, constrained repair, and gatekeeper validation, multi-round visual editing still struggles to satisfy page budgets and avoid newly introduced visual defects.

\textbf{PaperFit gives the best trade-off between visual quality and constraint satisfaction.} PaperFit achieves perfect compile and render success (1.000), the best VLM score (3.391), Win rate (0.895), and Page hit (0.805), with a Program score of 4.579 essentially tied with VisualMR. 
 
All methods maintain high content embedding similarity ($>$0.97), confirming that these gains come from layout-structure repair rather than semantic drift.

\subsection{Capability Boundary Comparison}
\label{sec:external_capability}

Rather than running additional external systems as direct experimental baselines—which would require substantial engineering adaptation to our specific task—we use recent papers and widely adopted open-source projects as external capability anchors. This choice avoids conflating method capability with engineering adaptation: existing external systems target related but different problems. Some systems specialize in parsing PDFs or document structure, some reconstruct local LaTeX objects from images, and some edit code repositories through command-line feedback. None of these system families directly targets full-paper visual typesetting repair for an existing LaTeX project.

\input{Tables/External_capability_boundary}

In Table~\ref{tab:external_capability_boundary}, \emph{multi-source input} means that a system can process at least one task-relevant input modality, such as PDFs, page images, local object images, code repositories, or textual instructions. DP systems extract text, equations, tables, and layout structure from PDF or document inputs, but their objective is document understanding or PDF-to-markup conversion rather than source-level repair. LR systems recover local LaTeX objects from formula or table images, but object-level image-to-LaTeX reconstruction is not equivalent to full-paper layout repair. CA systems can edit code repositories and iterate with command-line feedback, making them the closest external capability class to PaperFit; however, their feedback loop is usually organized around software task success or test passing, not visual diagnosis over page images rendered from compiled PDFs.

VisualMR instantiates the general-purpose visual coding-agent capability class in the controlled setting. It can inspect files, run commands, compile and render the project, view page images, and edit LaTeX over fixed rounds. However, VisualMR is denied PaperFit's VTO taxonomy, structured repair plans, constrained repair policy, runtime state management, and checklist-gated validation. VisualMR is therefore marked as partial for full-paper visual diagnosis and layout repair, and as absent for page-budget, template, and gatekeeper constraints.

The capability matrix shows that external systems cover different local segments of the PaperFit capability chain, but no external family simultaneously covers multi-source input, LaTeX/code editing, execution feedback, page-image-based full-paper diagnosis, float/table/page-level repair, and page-budget/template/gatekeeper constraints. PaperFit's contribution is not any single input parser, code editor, or local LaTeX recognizer; it is the integration of these capabilities into a full-paper visual typesetting optimization loop for existing LaTeX projects.

\subsection{Model Backend Comparison}
\label{sec:llm_comparison}

\begin{figure*}[ht]
    \centering
    \includegraphics[width=\textwidth]{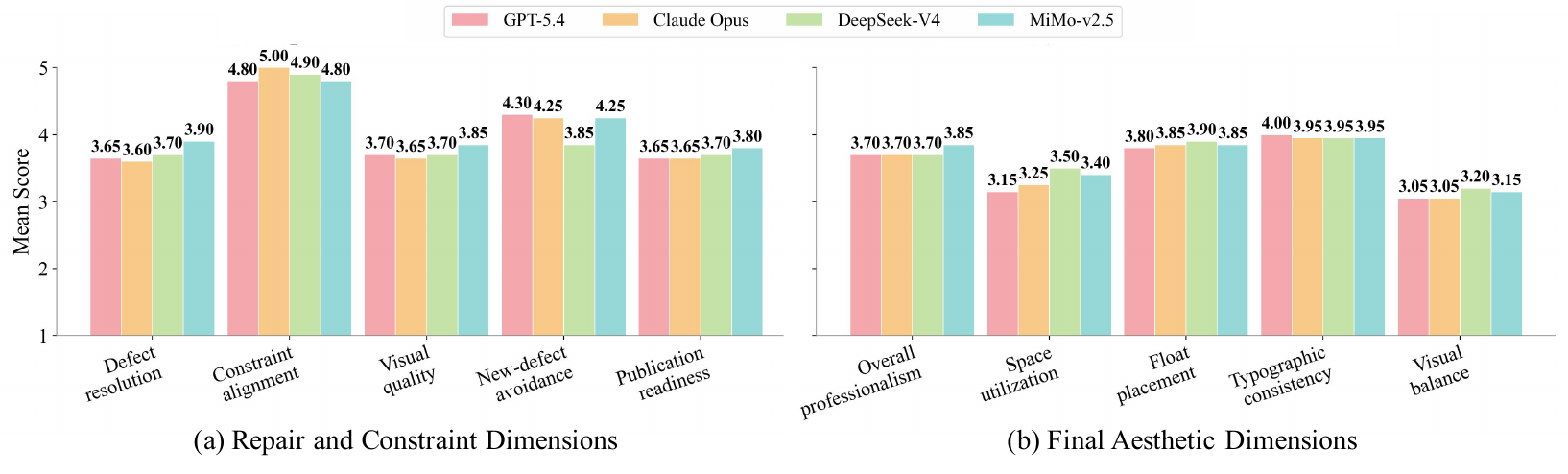}
    \vspace{-4mm}
    \caption{Fine-grained VLM scores for the LLM backend comparison. Panel~(a) reports repair and constraint dimensions, and panel~(b) reports final aesthetic dimensions. 
    }
    \vspace{-2mm}
    \label{fig:model_comparison_vlm_dimension}
\end{figure*}

\begin{figure*}[ht]
    \centering
    \includegraphics[width=0.98\textwidth]{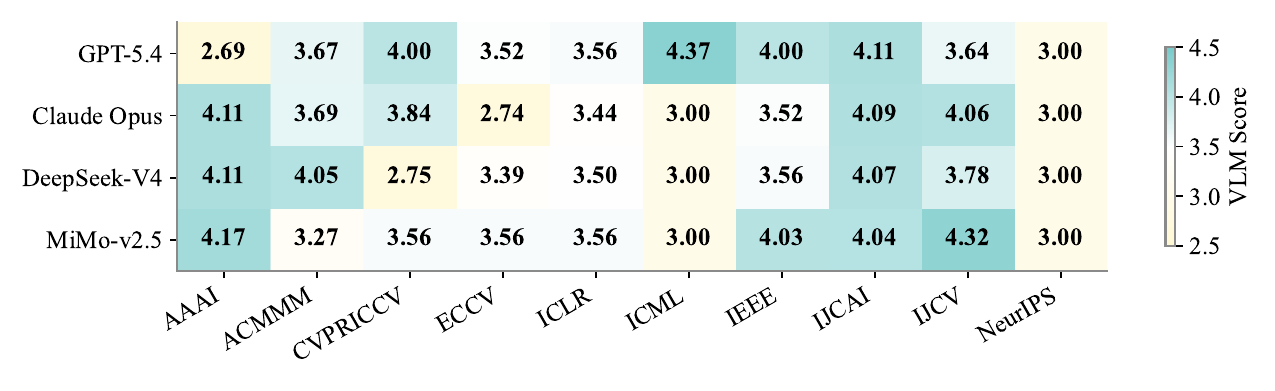}
    \vspace{-3mm}
    \caption{Venue-level VLM score distribution for the LLM backend comparison. 
    }
    \label{fig:model_comparison_venue_heatmap}
\end{figure*}

To isolate the role of the language-model backend, we run the same PaperFit workflow with four diverse LLMs on 20 representative cases. Table~\ref{tab:model_comparison} and Figures~\ref{fig:model_comparison_vlm_dimension}--\ref{fig:model_comparison_venue_heatmap} show three consistent patterns.

\input{Tables/Model_comparison}

\textbf{Aggregate performance is stable across backends.} All backends obtain high VLM scores (3.52--3.66), strong win rates (90--100\%), and near-perfect compile/render reliability. The overall VLM spread is only 0.14 points, far smaller than the 0.59-point gap between PaperFit and VisualMR in Table~\ref{tab:exp_main_results}, suggesting that the main improvement is from PaperFit rather than a particular model.

\textbf{Backend differences reflect repair style rather than a single dominant model.} Figure~\ref{fig:model_comparison_vlm_dimension}(a) shows that MiMo-v2.5 has the strongest repair-oriented profile, leading in defect resolution (3.90), visual quality (3.85), and publication readiness (3.80). GPT-5.4 instead leads in new-defect avoidance (4.30) and remains competitive on constraint alignment, which explains why its gated visual score slightly exceeds MiMo despite a lower raw visual score.

\textbf{The residual bottleneck is visual balance, not execution reliability.} Figure~\ref{fig:model_comparison_vlm_dimension}(b) shows that DeepSeek-V4 leads in space utilization (3.50), float placement (3.90), and visual balance (3.20), while MiMo-v2.5 has the highest overall professionalism (3.85). However, across all backends, typographic consistency is consistently high, whereas space utilization and visual balance remain the weakest dimensions. Venue-level results in Figure~\ref{fig:model_comparison_venue_heatmap} further show that no backend dominates uniformly across templates. 

\textbf{Difficulty-split results.} Table~\ref{tab:model_comparison_difficulty} reports the difficulty-split VLM scores for the four LLM backends. The VLM spread remains $\leq$0.14 within each difficulty tier, and no single backend dominates across all three levels---GPT-5.4 leads on easy and medium cases while DeepSeek-V4 Pro achieves the highest score on hard cases. This cross-over pattern confirms that the ranking reflects stochastic variation rather than a systematic backend advantage.

\begin{table}[ht]
\centering
\caption{Difficulty-split VLM scores for the LLM comparison (20 cases). All four backends remain effective across difficulty levels, with score spread $\leq 0.14$ on each split.}
\label{tab:model_comparison_difficulty}
\setlength{\tabcolsep}{4pt}
\small
\begin{tabular*}{\textwidth}{@{\extracolsep{\fill}}lccc@{}}
\toprule
\textbf{LLM Backend} & \textbf{Easy ($n$=6)} & \textbf{Medium ($n$=8)} & \textbf{Hard ($n$=6)} \\
\midrule
GPT-5.4~\cite{openai_gpt54_2026}            & \textbf{3.821} & \textbf{3.792} & 3.310 \\
Claude Opus 4.6~\cite{anthropic_claude_opus_46_2026}    & 3.598 & 3.442 & 3.638 \\
DeepSeek-V4 Pro~\cite{deepseek_v4_pro_2026}    & 3.163 & 3.509 & \textbf{3.893} \\
MiMo-v2.5-pro~\cite{xiaomi_mimo_v25_pro_2026}     & 3.648 & 3.778 & 3.486 \\
\bottomrule
\end{tabular*}
\end{table}

\subsection{Human--VLM Evaluation Correlation}
\label{sec:human_vlm_correlation}

To assess alignment between human judgments and automated scores, the Spearman correlation coefficient ($r$) is computed between VLM scores and average human ratings across all methods. As shown in 
\begin{wrapfigure}[13]{r}{0.5\textwidth}
    \centering
    \vspace{-2mm}
    \includegraphics[width=\linewidth]{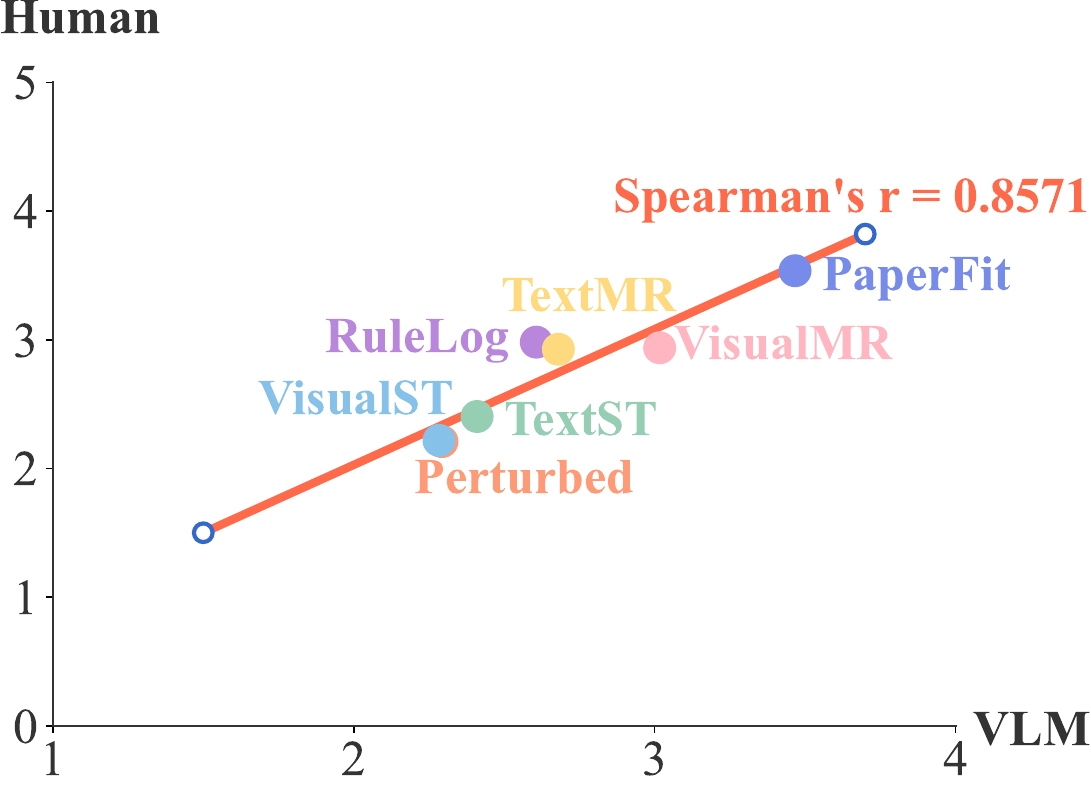}
    \caption{Human/VLM evaluation correlation.}
    \label{fig:spearman_correlation}
\end{wrapfigure}
Figure~\ref{fig:spearman_correlation}, the overall correlation is exceptionally high ($r = 0.8571$), confirming that the automated metric closely reflects human-perceived quality and faithfully captures model performance trends in the typesetting repair domain.

\subsection{Qualitative Case Study}
\label{sec:case_study}

\begin{figure}[ht]
\centering
\includegraphics[width=\textwidth]{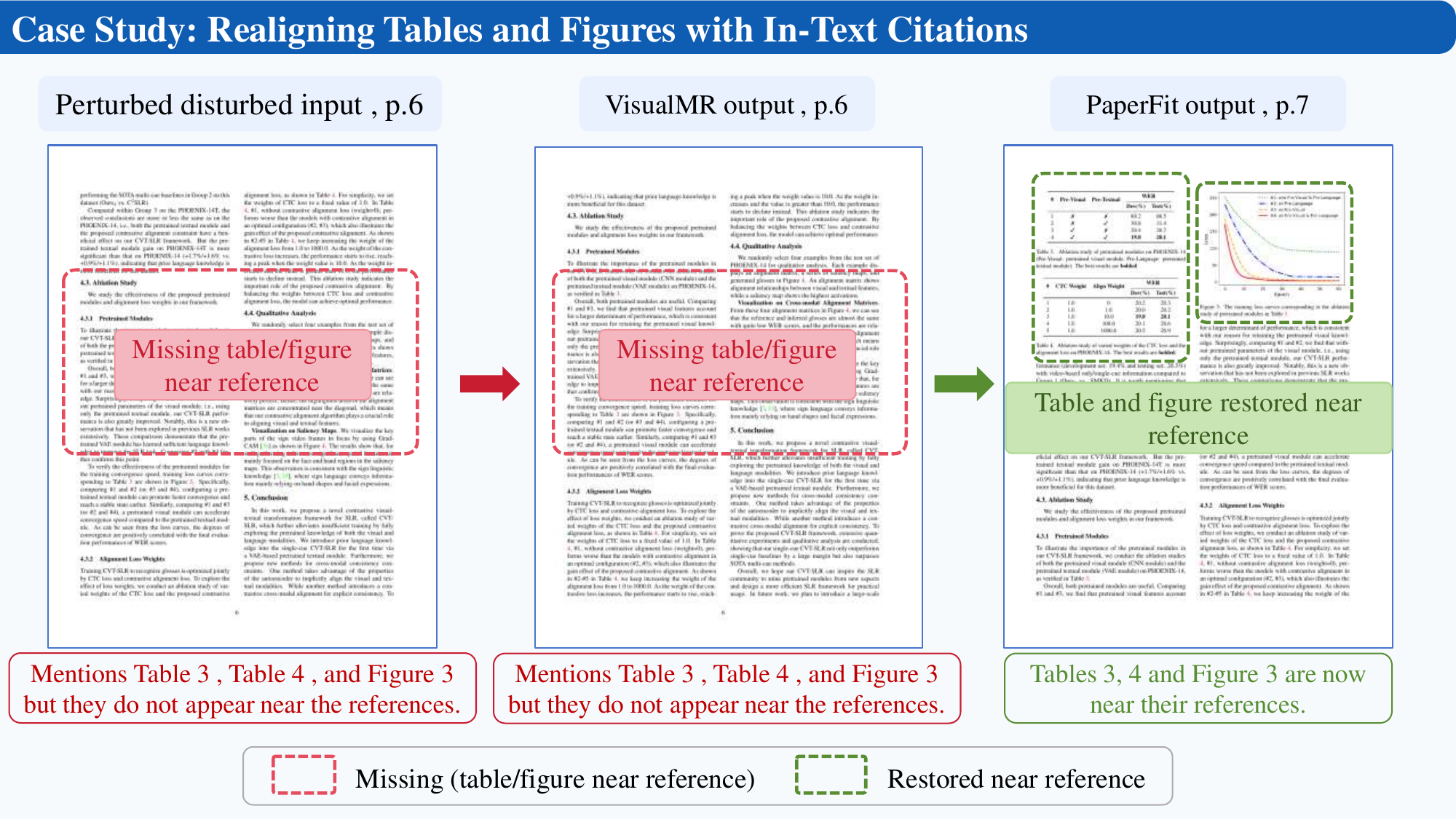}
\caption{Case Study: Realigning Tables and Figures with In-Text Citations.}
\vspace{-1mm}
\label{fig:case_study1}
\end{figure}

\begin{figure}[ht]
    \centering
    \vspace{-2mm}
    \includegraphics[width=\textwidth]{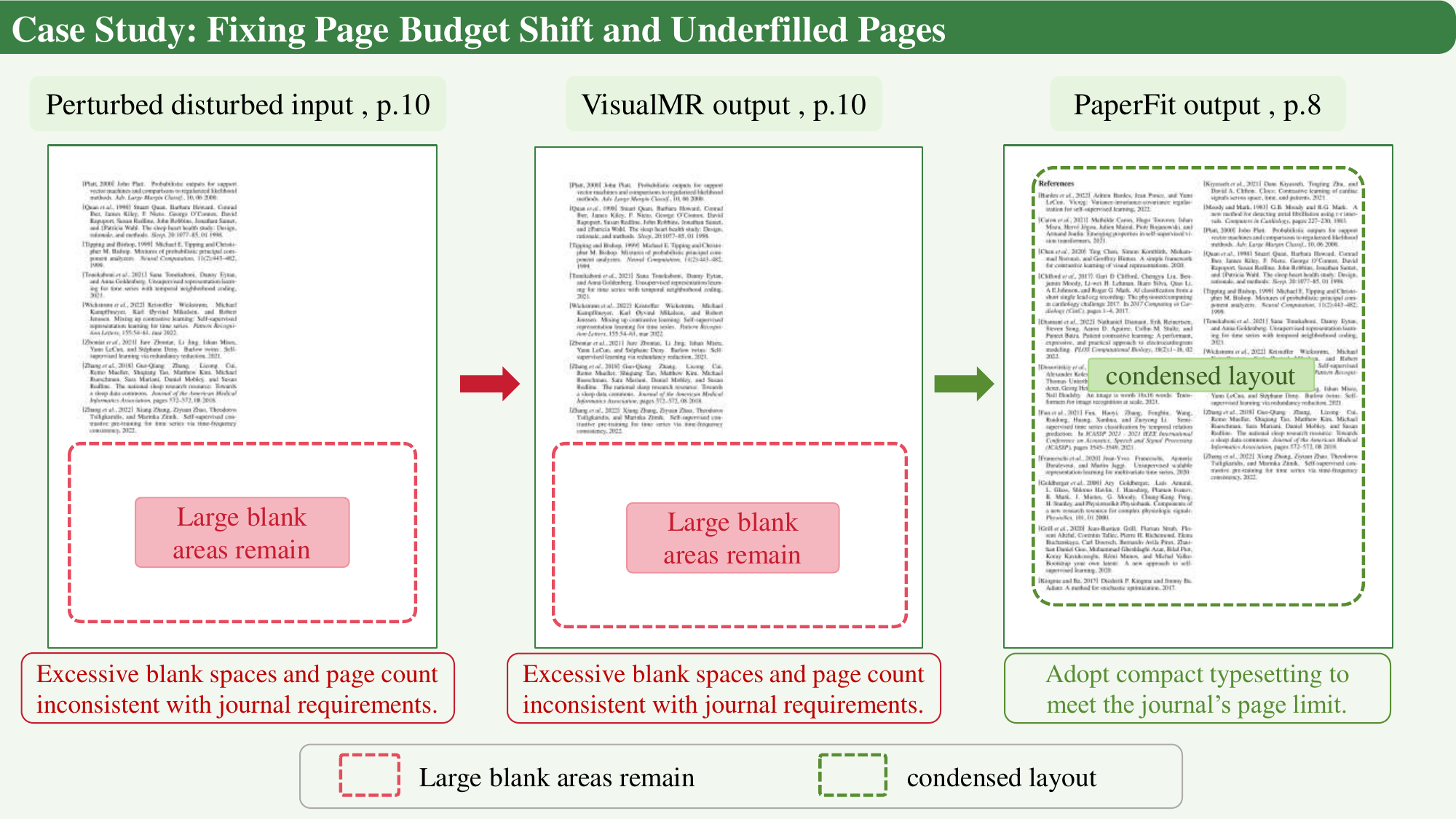}
    \caption{Case Study: Fixing Page Budget Shift and Underfilled Pages.}
    \vspace{-4mm}
    \label{fig:case_study2}
\end{figure}

\begin{figure}[ht]
    \centering
    \vspace{-2mm}
    \includegraphics[width=\textwidth]{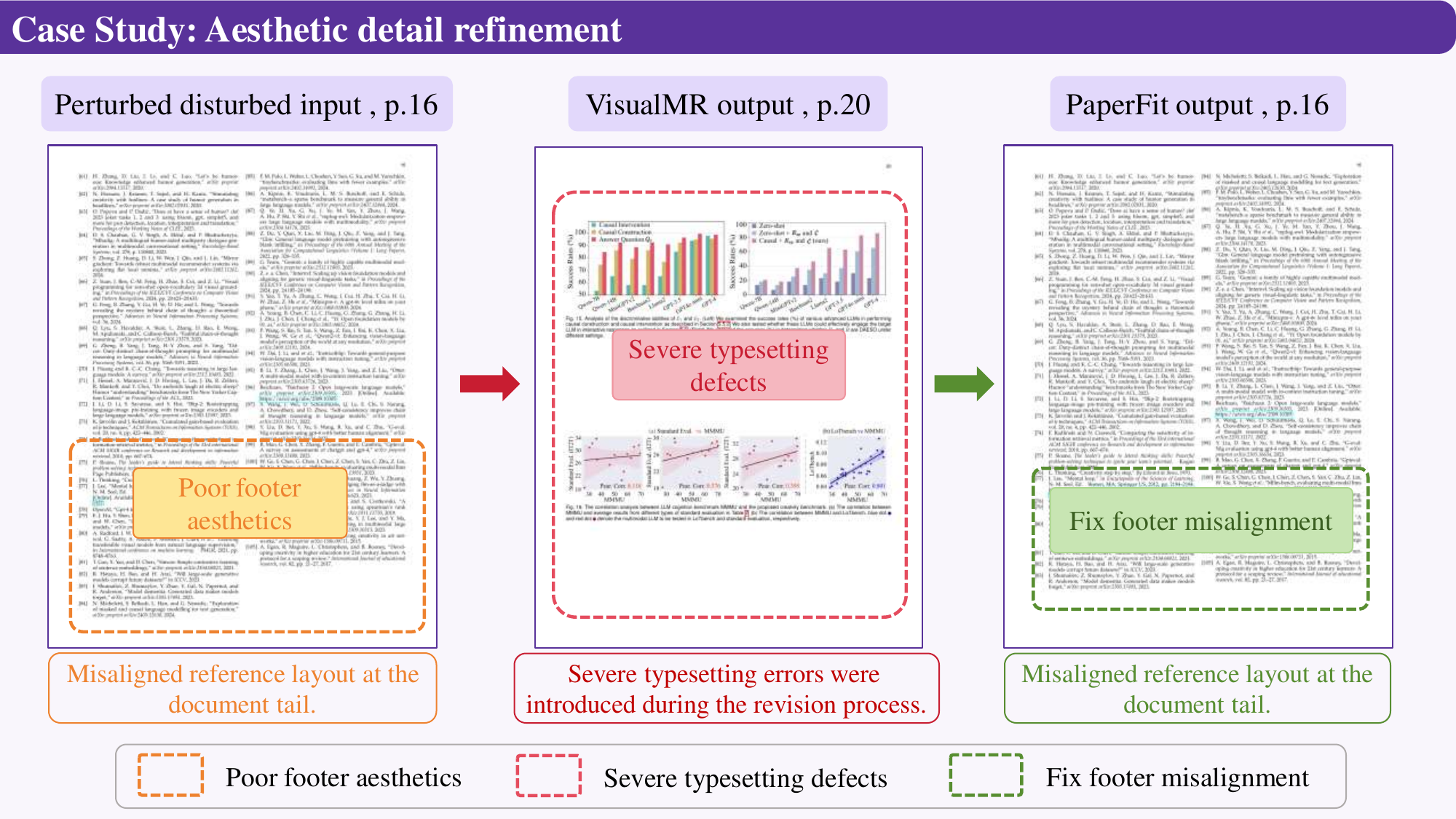}
    \caption{Case Study: Aesthetic Detail Refinement.}
    \vspace{-4mm}
    \label{fig:case_study3}
\end{figure}

\begin{figure}[ht]
    \centering
    \vspace{-2mm}
    \includegraphics[width=\textwidth]{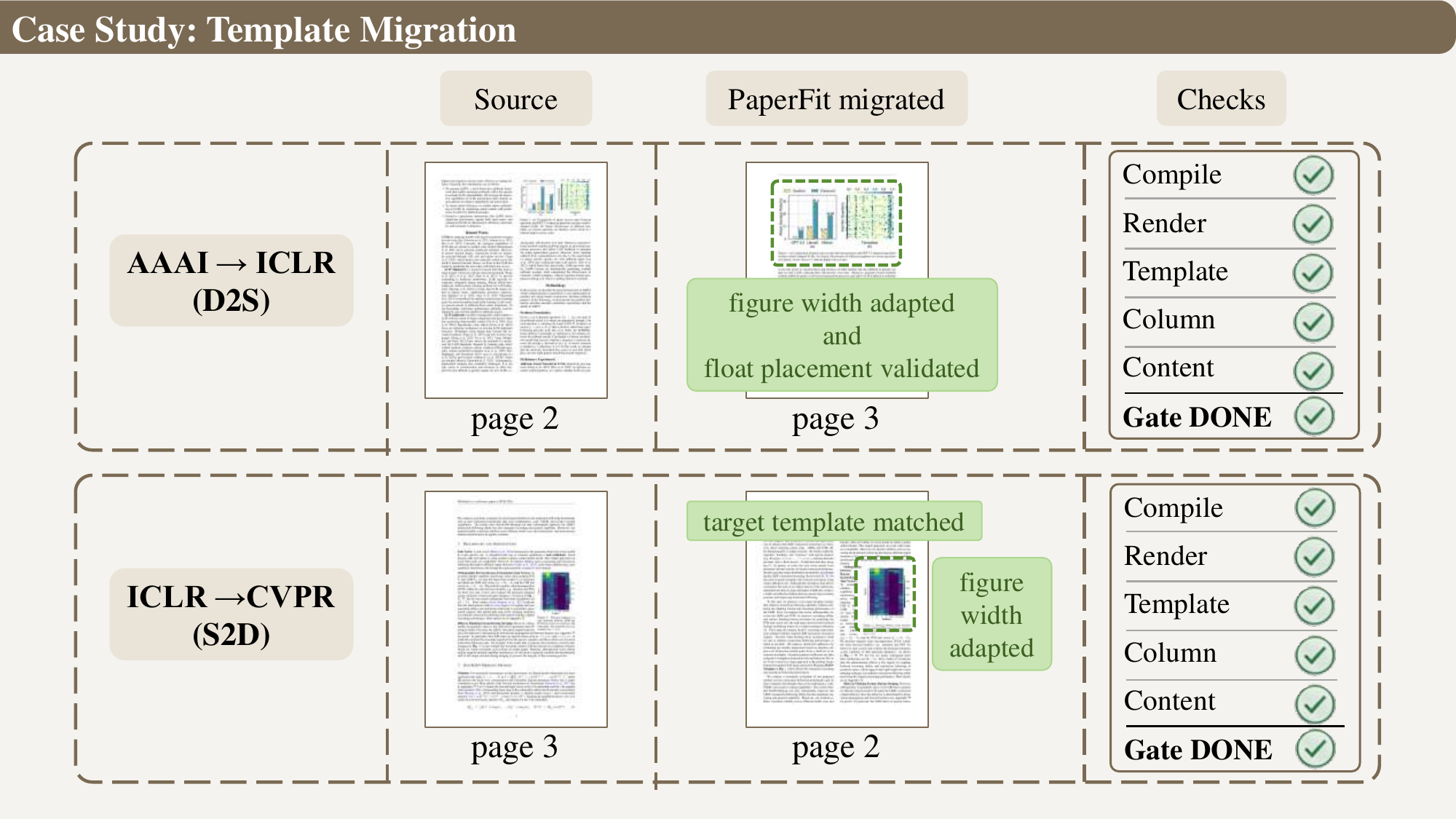}
    \caption{Case Study: Template Migration.}
    \vspace{-4mm}
    \label{fig:case_study4}
\end{figure}

Figure~\ref{fig:case_study1}--~\ref{fig:case_study4} presents the qualitative cases spanning distinct VTO modes.

\textbf{Case Study: Realigning Tables and Figures with In-Text Citations.} As shown in Figure~\ref{fig:case_study1} On the CVPR/ICCV case (target 10 pages), the disturbed input displaces tables and figures away from their semantic anchors. Both Perturbed and VisualMR render the ablation-study page but show a large region that mentions Table~3, Table~4, and Figure~3 without the corresponding visual objects nearby. PaperFit restores Tables~3--4 and Figure~3 near their references and satisfies the 10-page budget, whereas VisualMR produces 13 pages.

\textbf{Case Study: Fixing Page Budget Shift and Underfilled Pages.} As shown in Figure~\ref{fig:case_study2},  On the IJCAI case (target 8 pages), template migration creates excessive blank spaces and a page-count mismatch. VisualMR compiles and renders but leaves large blank areas on the final references page, stopping at 10 pages. PaperFit adopts compact typesetting to condense the layout and meets the 8-page limit while preserving the reference section.

\textbf{Case Study: Aesthetic Detail Refinement.} As shown in Figure~\ref{fig:case_study3}, On the IEEE case (target 16 pages), the disturbed input exhibits poor footer aesthetics with misaligned reference layout at the document tail. VisualMR restores compilation but introduces severe typesetting defects and expands the document to 20 pages under a 16-page target. PaperFit fixes the footer misalignment, restores a compact reference layout, and returns to 16 pages.

\textbf{Case Study: Template Migration.} As shown in Figure~\ref{fig:case_study4}, on two mainstream academic layout conversion cases (AAAI → ICLR for double-to-single column, and ICLR → CVPR for single-to-double column), direct template migration causes severe layout mismatches including figure width overflow and disordered float placement, failing to meet the target venue's formatting requirements. PaperFit automatically adapts figure dimensions to fit the target layout constraints, validates and optimizes float placement, and precisely matches the target template specifications. All core validation checks (compilation, rendering, template matching, column alignment, and content integrity) are fully passed, achieving end-to-end compliant template migration without manual intervention.

Across all case studies, VisualMR produces renderable output but fails to resolve the underlying layout defects and misses the page constraint. It lacks persistent defect records and acceptance gates, so it stops at the first renderable result. PaperFit instead diagnoses each failure through its structured taxonomy, applies constrained repairs, and validates outputs through a checklist gate. This qualitative evidence supports the quantitative trend in Section~\ref{sec:experiment}: visual feedback alone is useful, but reliable full-paper VTO requires an organized closed loop with explicit defect records, repair constraints, and acceptance gates.

\begin{figure}[ht]
    \centering
    \includegraphics[width=\textwidth]{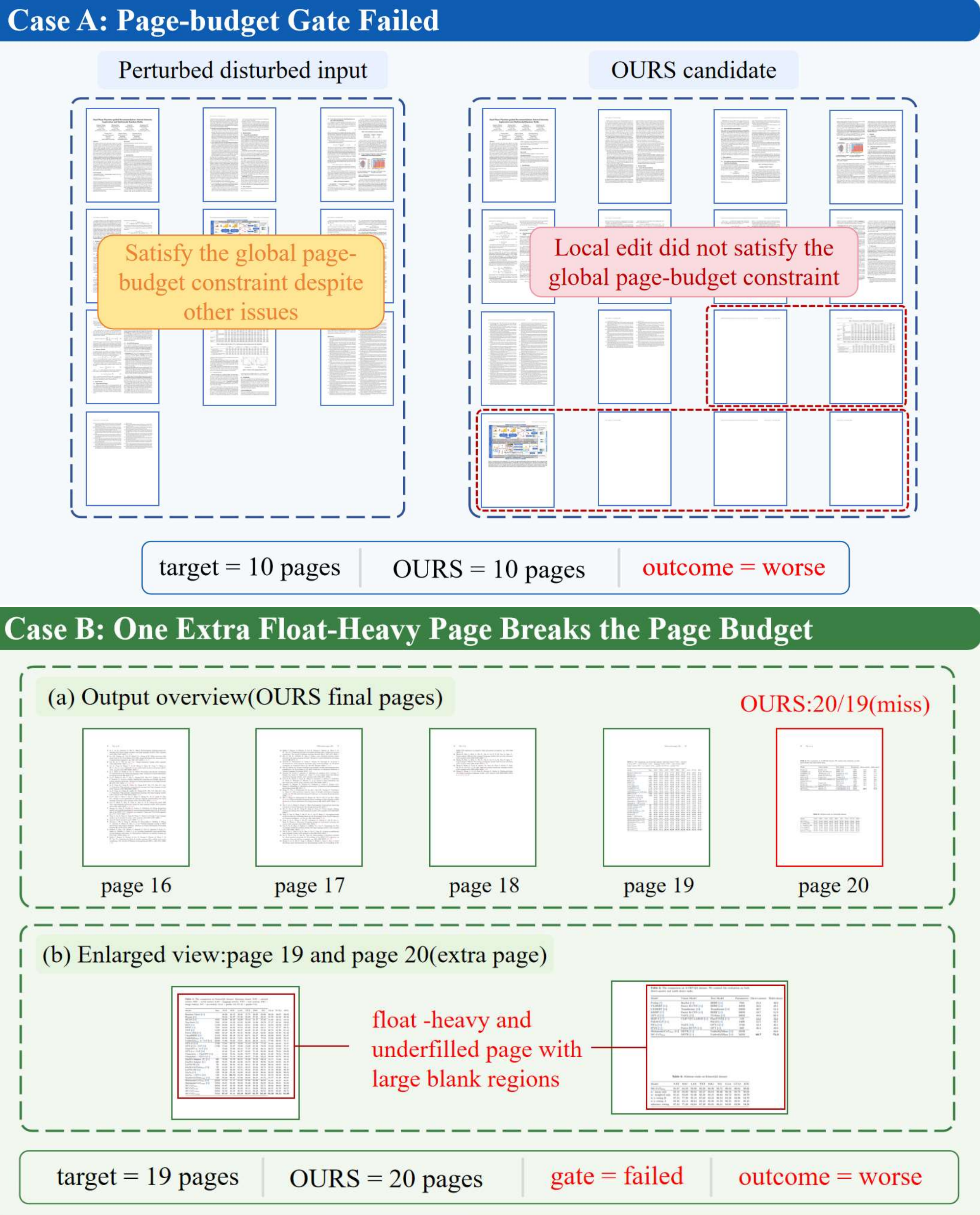}
    \caption{Error analysis: page-budget violations. Case A: Page-budget gate failed; target 10 pages, OURS produces 16 sparse pages. Case B: One extra float-heavy page; target 19 pages, OURS produces 20 with an underfilled final page.}
    \label{fig:error_case1}
\end{figure}

\subsection{Error Analysis}
\label{sec:error_analysis}

To understand the remaining failure modes of the PaperFit, we examine four representative error cases grouped into two figures. Each figure contains two failure examples, comparing the perturbed input with the OURS candidate output.

\subsubsection{Global Page-Budget Violations}

Figure~\ref{fig:error_case1} shows two cases in which the agent violates the global page-budget constraint.

\textbf{Case A: Page-budget Gate Failed.} An ACM Multimedia paper with a target of 10~pages produces a 16-page output. The agent's iterative repairs create several sparse trailing pages, indicating effective local edits but insufficient global page-budget control.

\textbf{Case B: One Extra Float-Heavy Page Breaks the Page Budget.} An ECCV paper with a target of 19~pages produces a 20-page output. The final page contains only a single large figure with substantial whitespace, showing that even a one-page deviation constitutes a hard failure when the added page is largely empty.

\subsubsection{Residual Visual Defects and Invalid Output}

\begin{figure}[ht]
    \centering
    \includegraphics[width=\textwidth]{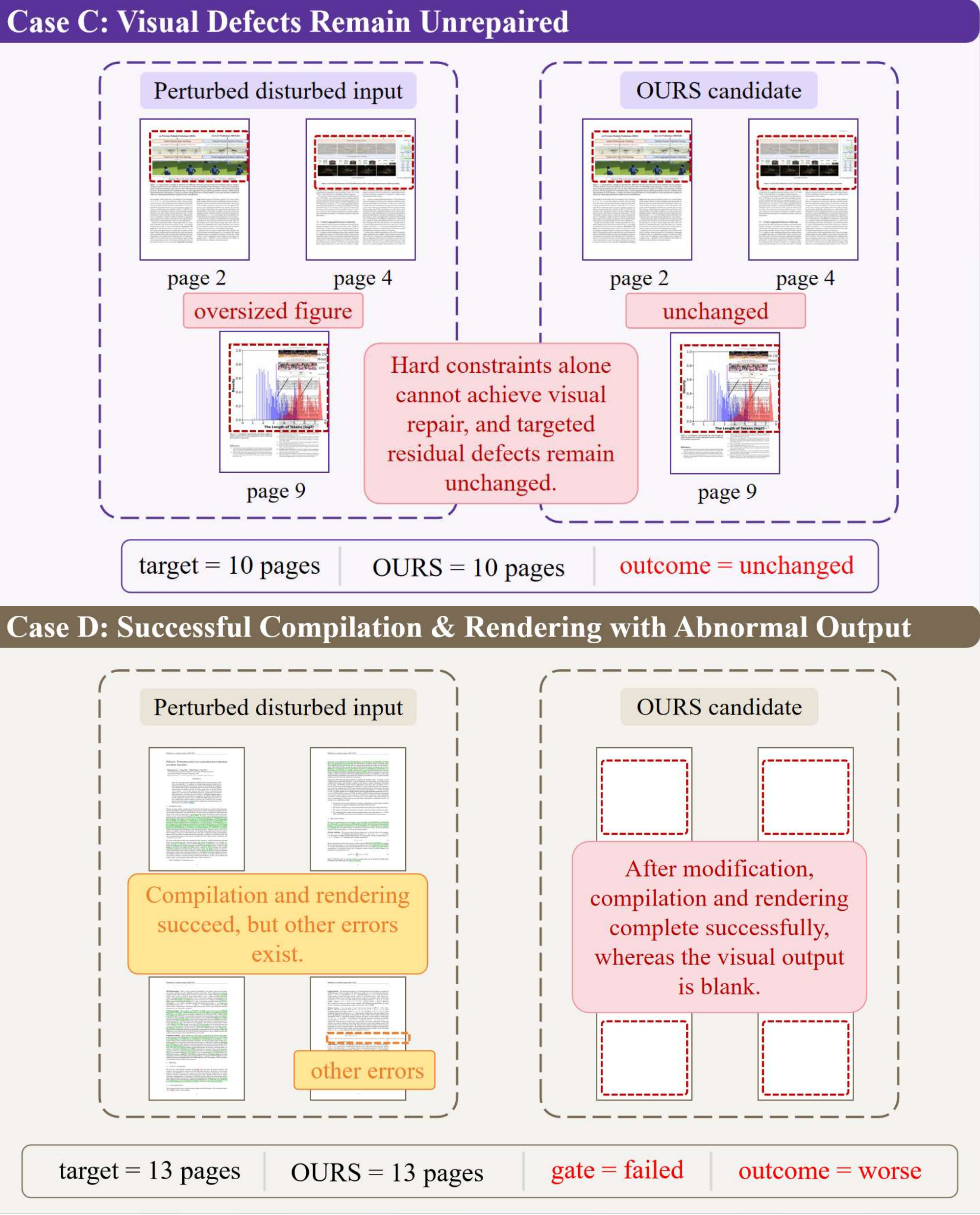}
    \caption{Error analysis: visual defects and invalid output. Case C: Visual defects remain unrepaired; compiles and meets page budget but the target figure defect persists. Case D: Successful compilation and rendering with abnormal output; correct page count but produces visually invalid grayed pages.}
    \label{fig:error_case2}
\end{figure}

Figure~\ref{fig:error_case2} shows two cases in which compilation and page-count metadata pass, but the visual output remains defective.

\textbf{Case C: Visual Defects Remain Unrepaired.} An ACM Multimedia paper meets both compilation and page-budget targets (10/10), yet the oversized and cropped figure remains essentially unrepaired. Satisfying hard constraints alone does not guarantee that the intended visual repair has been achieved.

\textbf{Case D: Successful Compilation \& Rendering with Abnormal Output.} An ICLR paper compiles successfully with the correct page count (13/13), but the rendered pages are grayed and visually invalid. LaTeX-level compilation success is insufficient as a sole quality indicator.

\section{Conclusion}
\label{sec:conclusion}
This paper identifies Visual Typesetting Optimization as a missing stage in document automation and introduces PaperFit, a vision-in-the-loop agent that bridges the gap between compilable and publication-ready LaTeX through multi-source evidence integration, constrained repair policies, and checklist-gated validation. On PaperFit-Bench (200 papers, 10 templates, 13 defects), PaperFit achieves perfect compile success, the highest VLM score, and an 80.5\% page-budget hit rate.

\clearpage
\newpage
\bibliographystyle{plainnat}
\setcitestyle{numbers}
\bibliography{ref}

\clearpage
\newpage
\beginappendix

\section{Benchmark Papers}
All papers comprising the PaperFit-Bench are cataloged in Table \ref{tab:benchmark1}--\ref{tab:benchmark8}, where complete details pertaining to the benchmark corpus are available.

\input{Tables/Benchmark_table}

\section{Prompt Records}
\label{sec:appendix_prompts}

For reproducibility, all model-facing baselines are associated with fixed prompt templates and saved run artifacts. RuleLog is not prompt-based: it is a deterministic rule/log baseline and therefore has no LLM prompt. TextST, TextMR, and VisualST use versioned text and vision prompt templates in the baseline adapter implementation. VisualMR writes the actual per-case agent prompt to \texttt{reports/agent\_prompt.txt}. PaperFit's agent-backed runs write the actual per-case prompt to \texttt{reports/claude\_prompt.txt}, together with raw responses and usage reports.

\begin{table}[h]
\centering
\caption{Prompt and artifact record for prompt-based methods.}
\label{tab:appendix_prompt_record}
\footnotesize
\begin{tabular*}{\textwidth}{@{\extracolsep{\fill}}llll@{}}
\toprule
\textbf{Method} & \textbf{Prompt role} & \textbf{Template source} & \textbf{Saved artifacts} \\
\midrule
RuleLog & Rule/log only & None & Rule reports \\
TextST & Source-only edit & Adapter prompts & Response, usage, boundary \\
TextMR & Source + log edit & Adapter prompts & Responses, usage, boundaries \\
VisualST & Source + image edit & Adapter prompts & Response, usage, boundary \\
VisualMR & Fixed-round visual agent & Agent prompt builder & Prompt, response, usage \\
PaperFit & Structured repair agent & PaperFit agent prompt & Prompt, response, usage \\
\bottomrule
\end{tabular*}
\end{table}

The prompt templates also define the input boundary for each baseline. TextST is restricted to source-only repair, TextMR adds compile-log feedback, VisualST adds rendered page images but only one visual edit turn, and VisualMR uses a fixed-round agent instruction that explicitly forbids PaperFit runtime, PaperFit skills, structured repair plans, taxonomy documents, and gatekeeper artifacts. PaperFit's prompt, in contrast, includes the VTO taxonomy, forbidden operations, repair priority, and quality-gate workflow used by the proposed system. These prompt records make the baseline boundaries auditable and prevent hidden prompt differences from being treated as implementation details.

\paragraph{Prompt templates.}
The following boxes show the core prompt templates used by the prompt-based methods. Case-specific fields such as the main TeX filename, target page count, maximum rounds, source window, compile-log excerpt, and rendered page images are filled at runtime.

\begin{promptbox}[TextST Source-Only Repair Prompt]
\small
\textbf{Role.} You are a LaTeX editing model for academic paper layout repair. You only revise the provided TeX source directly.

\medskip
\textbf{Input evidence.} Main \texttt{.tex} source only. No compile log and no rendered page images.

\medskip
\textbf{Hard constraints.}
\begin{itemize}[leftmargin=1.2em,itemsep=1pt]
    \item Preserve academic meaning, figures, tables, captions, labels, citations, and bibliography entries.
    \item Do not rewrite the whole paper; prefer one small local edit.
    \item Keep unchanged lines byte-identical whenever possible.
    \item Prefer local edits around floats, widths, line breaks, and spacing-safe LaTeX parameters.
    \item If a safe local fix is not possible, return \texttt{NO\_CHANGE}.
\end{itemize}

\medskip
\textbf{Output form.} Return either \texttt{NO\_CHANGE} or a revised TeX block, followed by a boundary-report JSON block describing the edit target and input scope.
\end{promptbox}
\captionof{figure}{Prompt template used for \textbf{TextST source-only LaTeX repair}. The method receives only the main TeX source, applies a small local source edit when safe, and records a boundary report for traceability.}
\label{prompt:b2_source_only}

\begin{promptbox}[TextMR Source-and-Log Repair Prompt]
\small
\textbf{Role.} You are a text/log LaTeX repair model. You revise the TeX source directly using source-level evidence and compile-log feedback.

\medskip
\textbf{Input evidence.} Main \texttt{.tex} source plus compile-log excerpt when available. No rendered page images.

\medskip
\textbf{Repair loop.}
\begin{itemize}[leftmargin=1.2em,itemsep=1pt]
    \item Inspect compile errors, undefined controls, overfull warnings, template mismatch hints, and local source signals.
    \item Propose a small source edit that addresses the most plausible local cause.
    \item Keep the edit within the selected source window when a local target is provided.
    \item Stop with \texttt{NO\_CHANGE} if the requested repair requires visual judgment or broad rewriting.
\end{itemize}

\medskip
\textbf{Output form.} Return a revised TeX block or local replacement window, together with a boundary-report JSON block using input scope \texttt{source\_plus\_log}.
\end{promptbox}
\captionof{figure}{Prompt template used for \textbf{TextMR source-and-log LaTeX repair}. The method augments source-only editing with compile-log feedback while still excluding rendered page images.}
\label{prompt:b3_source_log}

\begin{promptbox}[VisualST Single-Turn Visual Repair Prompt]
\small
\textbf{Role.} You are a vision-language model for academic paper layout repair. You may read TeX source, compile logs, and rendered page images, but you may not use structured planning artifacts, repair plans, gatekeeper logic, or a custom execution pipeline.

\medskip
\textbf{Input evidence.} Source pack and rendered page images from the pre-repair output. One visual repair turn only.

\medskip
\textbf{Hard constraints.}
\begin{itemize}[leftmargin=1.2em,itemsep=1pt]
    \item Preserve academic meaning and all protected paper objects.
    \item Do not add explanatory prose outside the required output form.
    \item Do not rewrite the whole paper.
    \item Prefer at most one local edit in one file, targeting visible layout defects such as overwide figures, underused table width, float placement, or local spacing.
    \item Return \texttt{NO\_CHANGE} if a safe local fix cannot be identified from the images.
\end{itemize}

\medskip
\textbf{Output form.} Return either \texttt{NO\_CHANGE}, a revised main TeX source, or one revised included TeX file, followed by the boundary-report JSON block.
\end{promptbox}
\captionof{figure}{Prompt template used for \textbf{VisualST single-turn visual repair}. The method receives rendered page images and source context, then performs one constrained visual edit without a structured repair workflow.}
\label{prompt:b4_single_turn_visual}

\begin{promptbox}[VisualMR Fixed-Round Visual Agent Prompt]
\small
\textbf{Role.} You are an autonomous LaTeX paper layout repair agent running in unattended benchmark mode.

\medskip
\textbf{Scope.} Work only inside the current case directory. You may inspect and edit \texttt{.tex}, \texttt{.bib}, \texttt{.cls}, \texttt{.sty}, image metadata, and LaTeX build artifacts. You may run ordinary LaTeX and local PDF/image tools.

\medskip
\textbf{Forbidden PaperFit resources.} Do not use PaperFit runtime, PaperFit CLI commands, PaperFit skills, PaperFit scripts, PaperFit repair plans, PaperFit gatekeeper outputs, PaperFit taxonomy documents, or any PaperFit structured artifacts.

\medskip
\textbf{Fixed-round loop.}
\begin{enumerate}[leftmargin=1.4em,itemsep=1pt]
    \item Compile with \texttt{latexmk} or an equivalent local LaTeX command.
    \item If compilation fails, inspect the log and repair the root LaTeX error.
    \item If a PDF exists, inspect or render pages using ordinary local tools.
    \item Edit source files minimally, then recompile and reinspect.
    \item Stop when the paper compiles/renders cleanly and no obvious layout issue remains, or after the maximum number of repair rounds.
\end{enumerate}

\medskip
\textbf{Output form.} Return compact JSON with success, rounds used, compile/render status, summary, and remaining issues.
\end{promptbox}
\captionof{figure}{Prompt template used for \textbf{VisualMR fixed-round visual agent repair}. The method can iterate over source, logs, and rendered pages for a fixed round budget while explicitly excluding PaperFit structured artifacts.}
\label{prompt:b5_fixed_round_agent}

\begin{promptbox}[PaperFit (OURS) Structured Repair Agent Prompt]
\small
\textbf{Role.} You are running in unattended benchmark mode for academic paper layout repair using the PaperFit closed-loop workflow.

\medskip
\textbf{Execution rules.}
\begin{itemize}[leftmargin=1.2em,itemsep=1pt]
    \item Work only inside the current case directory.
    \item Preserve all academic meaning and content.
    \item Compile and render after every edit before proceeding.
    \item Use the target page budget when provided.
\end{itemize}

\medskip
\textbf{Injected VTO knowledge.} Diagnose and repair defects using the five-category VTO taxonomy: space utilization, float placement, typographic consistency, overflow, and cross-template migration.

\medskip
\textbf{Repair policy.} Fix compile errors first, then overflow, float placement, space utilization, typographic consistency, and cross-template issues. Avoid pseudo-layout fixes such as brute-force spacing, forced page breaks, table scaling, object deletion, or content rewriting.

\medskip
\textbf{Quality gate.} After each repair cycle, verify compilation, rendering, page-level visual inspection, residual defect status, page-budget satisfaction, and protected content preservation. Continue until all gate conditions pass or the maximum round budget is reached.

\medskip
\textbf{Output form.} Return JSON with success, rounds used, summary, remaining defects, and notes.
\end{promptbox}
\captionof{figure}{Prompt template used for \textbf{PaperFit structured repair}. The method injects the VTO taxonomy, repair priority, forbidden operations, and checklist quality gate used by the proposed closed-loop system.}
\label{prompt:paperfit_structured_agent}

\section{Reproducibility Notes}
\label{sec:appendix_reproducibility}

For each method and case, the evaluation records the generated source, compile logs, rendered pages when available, programmatic metric outputs, and VLM reports. Aggregated tables are computed from case-level reports rather than from hand-entered summary values. Missing, non-compilable, or non-renderable outputs are not silently dropped; they are handled by the same failure accounting rules used for all methods. All LaTeX outputs are compiled with the local TeX toolchain and rendered into page images before visual evaluation. Baseline outputs and aggregate metrics are stored under the baseline result directory, while the paper tables are generated from the completed method-level summaries. The release package will include the benchmark metadata, disturbance manifests, baseline implementations, evaluation scripts, and aggregation scripts needed to reproduce the programmatic metrics and VLM-based visual scoring.

\section{Limitations}

PaperFit's visual inspection depends on a VLM evaluator; subtle or ambiguous layout issues---such as microtypographic defects or font-level kerning errors---may still be missed by current vision models. On hard cases with 5--8 co-occurring perturbations, the page-budget hit rate drops to approximately 70\%, indicating that highly complex multi-defect scenarios remain challenging. 

The system is currently limited to LaTeX projects and has been evaluated only on English-language academic papers; coverage of other document languages is left to future work. Finally, multi-round recompilation and re-rendering incurs higher computational cost than single-pass methods, and reducing this overhead while preserving repair quality is an important practical direction.

\end{document}

%% file: Tables/Benchmark_statistics.tex
\begin{table*}[ht]
\centering
\small
\vspace{-2mm}
\caption{Benchmark papers statistics by conference.}
\label{tab:benchmark_statistics}
\begin{tabular*}{\textwidth}{@{\extracolsep{\fill}} lccccc @{}}
\toprule
\multicolumn{1}{c}{\textbf{Conference}} & \textbf{\# Papers} & \textbf{Columns} & \textbf{Page Limit} & \textbf{Avg. Figures} & \textbf{Avg. Tables} \\
\midrule
AAAI & 20 & 2 & 7 & 5.60 & 5.30 \\
ACM MM & 20 & 2 & 8 & 7.25 & 6.40 \\
CVPR/ICCV & 20 & 1 & 8 & 5.25 & 4.70 \\
ECCV & 20 & 1 & 14 & 5.65 & 4.35 \\
ICLR & 20 & 1 & 9 & 6.30 & 4.85 \\
ICML & 20 & 2 & 8 & 5.55 & 5.05 \\
IEEE Trans & 20 & 2 & 12 & 4.60 & 3.85 \\
IJCAI & 20 & 2 & 7 & 5.80 & 4.70 \\
IJCV & 20 & 2 & 10+ & 10.75 & 9.20 \\
NeurIPS & 20 & 1 & 9 & 5.95 & 4.40 \\
\bottomrule
\end{tabular*}
\end{table*}

%% file: Tables/Perturbation_strategy.tex
\begin{table*}[htb]
\centering
\small
\setlength{\tabcolsep}{5pt}
\caption{Summary of perturbation strategies. ``Validated'' column: $\checkmark$ = post-compilation semantic verification confirms the defect manifests (e.g., overfull log, page shift, or layout inspection), with results recorded in \texttt{disturbance\_manifest.json}; $\times$ = standard compilation check only, without defect-level verification.}
\label{tab:perturbation_strategy}
\begin{tabular*}{\textwidth}{@{\extracolsep{\fill}} l c c p{4cm} c c @{}}
\toprule
\textbf{Perturbation ID} & \textbf{Category} & \textbf{Defect} & \multicolumn{1}{c}{\textbf{Implementation}} & \textbf{Validated} & \textbf{Frequency} \\
\midrule
A1\_widow\_orphan        & A & A1 & Force widow/orphan lines via truncated short paragraphs & $\checkmark$ & 59 \\
A2\_trailing\_whitespace  & A & A2 & Inject trailing whitespace before bibliography & $\times$ & 72 \\
A4\_column\_imbalance     & A & A4 & Inject vertical gaps on double-column final pages & $\checkmark$ & 16 \\
A5\_column\_void          & A & A5 & Insert vertical voids within body text columns & $\checkmark$ & 13 \\
B1\_float\_to\_page       & B & B1 & Push selected floats to dedicated float-only pages & $\times$ & 50 \\
B2\_oversize\_graphics    & B & B2 & Enlarge graphics beyond available column width & $\times$ & 48 \\
B2\_shrink\_graphics      & B & B2 & Shrink graphics to noticeably undersized width & $\times$ & 51 \\
C1\_table\_resizebox      & C & C1 & Wrap tables in \textbackslash resizebox to undersize & $\times$ & 73 \\
C2\_table\_oversize       & C & C2 & Widen tables beyond available column width & $\times$ & 79 \\
D1\_long\_token\_overflow & D & D1 & Append unbreakable tokens to trigger line overflow & $\times$ & 68 \\
D2\_long\_formula         & D & D2 & Insert ultra-wide formulas to trigger display overflow & $\checkmark$ & 80 \\
E1\_template\_mismatch    & E & E1 & Cross-template migration with unreasonable image widths & $\checkmark$ & 76 \\
E2\_template\_page\_shift & E & E2 & Cross-template migration with reduced \textbackslash textheight & $\checkmark$ & 76 \\
\bottomrule
\end{tabular*}
\end{table*}

%% file: Tables/Perturbation_difficulty.tex
\begin{table*}[ht]
\centering
\small
\setlength{\tabcolsep}{10pt}
\caption{Distribution of perturbation difficulty levels and most frequently used perturbation types.}
\label{tab:perturbation_difficulty}
\begin{tabular*}{\textwidth}{@{\extracolsep{\fill}} l c l @{}}
\toprule
\textbf{Difficulty} & \multicolumn{1}{c}{\textbf{Samples}} & \multicolumn{1}{c}{\textbf{Common Perturbation Types}} \\
\midrule
Easy   & 60  & \parbox[c]{9cm}{\raggedright A2 (trailing whitespace), C1 (table resizebox), C2 (table oversize), D1 (long token overflow), D2 (long formula), E1 (template mismatch), E2 (template page shift)} \\
Medium & 80  & \parbox[c]{9cm}{\raggedright A2 (trailing whitespace), C1 (table resizebox), C2 (table oversize), D2 (long formula), E2 (template page shift)} \\
Hard   & 60  & \parbox[c]{9cm}{\raggedright A2 (trailing whitespace), C2 (table oversize), D1 (long token overflow), D2 (long formula), E1 (template mismatch), E2 (template page shift)} \\
\bottomrule
\end{tabular*}
\end{table*}

%% file: Tables/Benchmark_comparison.tex
\begin{table*}[t]
\centering
\caption{Comparison with representative benchmarks. PaperFit-Bench is the only benchmark that combines systematic perturbation injection, visual evaluation from rendered pages, multi-modal evidence chains, and full-document iterative repair.}
\label{tab:benchmark_comparison}
\setlength{\tabcolsep}{5pt}
\small
\begin{tabular*}{\textwidth}{@{\extracolsep{\fill}}lllcccc@{}}
\toprule
\textbf{Benchmark} & \textbf{Task} & \textbf{Perturbation} & \textbf{Visual Eval} & \textbf{Multi-Modal} & \textbf{Iterative} \\
\midrule
Im2Latex-100K & Formula reconstruction & -- & \xmark & \xmark & \xmark \\
TeXpert & LaTeX code generation & -- & \xmark & \xmark & \xmark \\
RoDLA & Layout robustness & Limited & Partial & \cmark & \xmark \\
DocReward & Quality assessment & -- & \xmark & \xmark & \xmark \\
LATTE & Element-level refinement & -- & \xmark & \cmark & \cmark \\
\midrule
\textbf{PaperFit-Bench} & \textbf{Visual typesetting repair} & \textbf{13 strategies} & \textbf{\cmark} & \textbf{\cmark} & \textbf{\cmark} \\
\bottomrule
\end{tabular*}
\end{table*}

%% file: Tables/Exp_main_results.tex
\begin{table*}[ht]
\centering
\vspace{-2mm}
\caption{Main results. VLM is the visual evaluation score; Program is the 0--5 composite programmatic score. All other quantities are rates in $[0,1]$.}
\vspace{-2mm}
\label{tab:exp_main_results}
\setlength{\tabcolsep}{3.5pt}
\small
\begin{tabular*}{\textwidth}{@{\extracolsep{\fill}}ccccccc@{}}
\toprule
\textbf{Method} & \textbf{Compile} $\uparrow$ & \textbf{Render} $\uparrow$ & \textbf{VLM} $\uparrow$ & \textbf{Win} $\uparrow$ & \textbf{Program} $\uparrow$ & \textbf{Page hit} $\uparrow$ \\
\midrule
Perturbed & 0.5800 & 0.8200 & 1.8275 & 0.0000 & 3.6344 & 0.3750 \\
RuleLog & 0.5200 & 0.7600 & 2.1838 & 0.3800 & 3.3401 & 0.4444 \\
TextST & 0.5850 & 0.5850 & 1.8522 & 0.2800 & 2.5738 & 0.4530 \\
TextMR & 0.6100 & 0.6100 & 2.1601 & 0.4250 & 2.7433 & 0.6230 \\
VisualST & 0.6250 & 0.6250 & 1.8741 & 0.2950 & 2.7681 & 0.4560 \\
\hline
VisualMR & 0.9750 & 0.9750 & 2.8006 & 0.6500 & 4.5789 & 0.5487 \\
PaperFit & \textbf{1.0000} & \textbf{1.0000} & \textbf{3.3907} & \textbf{0.8950} & \textbf{4.5790} & \textbf{0.8050} \\
\bottomrule
\vspace{-2mm}
\end{tabular*}
\end{table*}

%% file: Tables/External_capability_boundary.tex
\begin{table*}[h]
\centering
\caption{External capability boundary matrix. System families: \emph{DocParser} denotes PDF/document parsers, including MinerU, Marker, and Nougat~\cite{wang_mineru_2024,datalab_marker_2024,blecher_nougat_2023}; \emph{LocalRecon} denotes local LaTeX reconstruction systems, including LATTE, Table2LaTeX-RL, and LaTeX-OCR~\cite{jiang_latte_2025,ling_table2latex-rl_2025,blecher_latex-ocr_2022}; \emph{CodeAgent} denotes general coding agents, including OpenHands, Aider, and SWE-agent~\cite{wang_openhands_2024,gauthier_aider_2023,yang_swe-agent_2024}; \emph{B5-VisualAgent} is our general-purpose visual coding-agent baseline; \emph{PaperFit} is our full system. Capability abbreviations: MSI = multi-source input; Edit = LaTeX/code generation or editing; Loop = execution feedback loop; PVD = full-paper visual diagnosis from rendered page images; Layout = float/table/page-level layout repair; Gate = page-budget, template, and checklist-gated validation. \checkmark indicates full coverage, \xmark indicates no coverage, and $\triangle$ indicates partial coverage.}
\label{tab:external_capability_boundary}
\setlength{\tabcolsep}{4.2pt}
\small
\begin{tabular*}{\textwidth}{@{\extracolsep{\fill}}lcccccc@{}}
\toprule
\textbf{System family} & \textbf{MSI} & \textbf{Edit} & \textbf{Loop} & \textbf{PVD} & \textbf{Layout} & \textbf{Gate} \\
\midrule
DocParser~\cite{wang_mineru_2024,datalab_marker_2024,blecher_nougat_2023} & \checkmark & \xmark & \xmark & \xmark & \xmark & \xmark \\
LocalRecon~\cite{jiang_latte_2025,ling_table2latex-rl_2025,blecher_latex-ocr_2022} & \checkmark & \checkmark & $\triangle$ & \xmark & \xmark & \xmark \\
CodeAgent~\cite{wang_openhands_2024,gauthier_aider_2023,yang_swe-agent_2024} & \checkmark & \checkmark & \checkmark & \xmark & \xmark & \xmark \\
B5-VisualAgent & \checkmark & \checkmark & \checkmark & $\triangle$ & $\triangle$ & \xmark \\
PaperFit & \checkmark & \checkmark & \checkmark & \checkmark & \checkmark & \checkmark \\
\bottomrule
\end{tabular*}
\end{table*}

%% file: Tables/Model_comparison.tex
\begin{table*}[ht]
\centering
\vspace{-2mm}
\caption{Model comparison on 20 representative cases (6 easy, 8 medium, 6 hard). 
}
\label{tab:model_comparison}
\setlength{\tabcolsep}{4pt}
\small
\begin{tabular*}{\textwidth}{@{\extracolsep{\fill}}lccccc@{}}
\toprule
\textbf{LLM Backend} & \textbf{Compile} $\uparrow$ & \textbf{Render} $\uparrow$ & \textbf{VLM} $\uparrow$ & \textbf{Win} $\uparrow$ & \textbf{Page hit} $\uparrow$ \\
\midrule
GPT-5.4 \cite{openai_gpt54_2026}            & 100.00\% & 100.00\% & 3.656 & 95.00\% & 95.00\% \\
Claude Opus 4.6 \cite{anthropic_claude_opus_46_2026}    & 100.00\% & 100.00\% & 3.548 & 90.00\% & 100.00\% \\
DeepSeek-V4 Pro \cite{deepseek_v4_pro_2026}    & 95.00\% & 100.00\% & 3.521 & 95.00\% & 100.00\% \\
MiMo-v2.5-pro \cite{xiaomi_mimo_v25_pro_2026}     & 100.00\% & 100.00\% & 3.652 & \textbf{100.00\%} & 95.00\% \\
\bottomrule
\end{tabular*}
\end{table*}


%% file: Tables/Benchmark_table.tex
\begin{table*}[ht]
\centering
\caption{Benchmark papers organized by conference.}
\label{tab:benchmark1}
\resizebox{\textwidth}{!}{
\begin{tabular}{p{11cm}p{3cm}}
\toprule
\multicolumn{1}{c}{\textbf{Paper}} & \textbf{Conference} \\
\midrule
\fontsize{9}{11}\selectfont
Cross-gate mlp with protein complex invariant embedding is a one-shot antibody designer \cite{tan2024cross} & AAAI 2024 \\
Psc-cpi: Multi-scale protein sequence-structure contrasting for efficient and generalizable compound-protein interaction prediction \cite{wu2024psc} & AAAI 2024 \\
Foldtoken: Learning protein language via vector quantization and beyond \cite{gao2025foldtoken} & AAAI 2025 \\
Isr-dpo: Aligning large multimodal models for videos by iterative self-retrospective dpo \cite{ahn2025isr} & AAAI 2025 \\
Gim: A million-scale benchmark for generative image manipulation detection and localization \cite{chen2025gim} & AAAI 2025 \\
Enhancing adversarial transferability with adversarial weight tuning \cite{chen2025enhancing} & AAAI 2025 \\
Bias unveiled: Investigating social bias in LLM-generated code \cite{ling2025bias} & AAAI 2025 \\
Multi-view pedestrian occupancy prediction with a novel synthetic dataset \cite{aung2025multi} & AAAI 2025 \\
GLCF: A Global-Local Multimodal Coherence Analysis Framework for Talking Face Generation Detection \cite{chen2025glcf} & AAAI 2025 \\
JailPO: A Novel Black-box Jailbreak Framework via Preference Optimization against Aligned LLMs \cite{li2025jailpo} & AAAI 2025 \\
Neural continuous-time supermartingale certificates \cite{neustroev2025neural} & AAAI 2025 \\
Real-time calibration model for low-cost sensor in fine-grained time series \cite{ahn2025real} & AAAI 2025 \\
Relation-aware equivariant graph networks for epitope-unknown antibody design and specificity optimization \cite{wu2025relation} & AAAI 2025 \\
Causal-inspired multitask learning for video-based human pose estimation \cite{chen2025causal} & AAAI 2025 \\
dyab: Flow matching for flexible antibody design with alphafold-driven pre-binding antigen \cite{tan2025dyab} & AAAI 2025 \\
Unveiling the Landscape of Clinical Depression Assessment: From Behavioral Signatures to Psychiatric Reasoning \cite{chen2026unveiling} & AAAI 2026 \\
PITE: Multi-Prototype Alignment for Individual Treatment Effect Estimation \cite{cao2026pite} & AAAI 2026 \\
FreDN: Spectral Disentanglement for Time Series Forecasting via Learnable Frequency Decomposition \cite{an2026fredn} & AAAI 2026 \\
Spiking Heterogeneous Graph Attention Networks \cite{cao2026spiking} & AAAI 2026 \\
\bottomrule
\end{tabular}
}
\end{table*}
\clearpage
\begin{table*}[ht]
\centering
\caption{Benchmark papers organized by conference (cont.).}
\label{tab:benchmark2}
\resizebox{\textwidth}{!}{
\begin{tabular}{p{11cm}p{3cm}}
\toprule
\multicolumn{1}{c}{\textbf{Paper}} & \textbf{Conference} \\
\midrule
\fontsize{9}{11}\selectfont
NL2CA: Auto-formalizing Cognitive Decision-Making from Natural Language Using an Unsupervised CriticNL2LTL Framework \cite{deng2026nl2ca} & AAAI 2026 \\
\midrule
Co-learning: Learning from noisy labels with self-supervision \cite{tan2021co} & ACM MM 2021 \\
Convert: Contrastive graph clustering with reliable augmentation \cite{yang2023convert} & ACM MM 2023 \\
Chain-of-Cooking: Cooking Process Visualization via Bidirectional Chain-of-Thought Guidance \cite{xu2025chain} & ACM MM 2025 \\
Capturing more: Learning multi-domain representations for robust online handwriting verification \cite{zhang2025capturing} & ACM MM 2025 \\
Gather and trace: Rethinking video textvqa from an instance-oriented perspective \cite{zhang2025gather} & ACM MM 2025 \\
Audio Does Matter: Importance-Aware Multi-Granularity Fusion for Video Moment Retrieval \cite{lin2025audio} & ACM MM 2025 \\
Universally Unfiltered and Unseen: Input-Agnostic Multimodal Jailbreaks against Text-to-Image Model Safeguards \cite{yan2025universally} & ACM MM 2025 \\
AD-AVSR: Asymmetric Dual-stream Enhancement for Robust Audio-Visual Speech Recognition \cite{xue2025ad} & ACM MM 2025 \\
Advancing 3D Scene Understanding with MV-ScanQA Multi-View Reasoning Evaluation and TripAlign Pre-training Dataset \cite{mo2025advancing} & ACM MM 2025 \\
DIME-Net: A Dual-Illumination Adaptive Enhancement Network Based on Retinex and Mixture-of-Experts \cite{wang2025dime} & ACM MM 2025 \\
AnchorSync: Global Consistency Optimization for Long Video Editing \cite{liu2025anchorsync} & ACM MM 2025 \\
Dual-Phase Playtime-guided Recommendation: Interest Intensity Exploration and Multimodal Random Walks \cite{zhang2025dual} & ACM MM 2025 \\
Investigating Domain Gaps for Indoor 3D Object Detection \cite{zhao2025investigating} & ACM MM 2025 \\
Drawing2CAD: Sequence-to-Sequence Learning for CAD Generation from Vector Drawings \cite{qin2025drawing2cad} & ACM MM 2025 \\
Robust multimodal sentiment analysis of image-text pairs by distribution-based feature recovery and fusion \cite{wu2024robust} & ACM MM 2024 \\
Hud: Hierarchical uncertainty-aware disambiguation network for composed video retrieval \cite{chen2025hud} & ACM MM 2025 \\
Omnigen: Unified multimodal sensor generation for autonomous driving \cite{tang2025omnigen} & ACM MM 2025 \\
Magicfight: Personalized martial arts combat video generation \cite{huang2024magicfight} & ACM MM 2024 \\
RealHD: A High-Quality Dataset for Robust Detection of State-of-the-Art AI-Generated Images \cite{yu2025realhd} & ACM MM 2025 \\
Focustrack: One-stage focus-and-suppress framework for 3d point cloud object tracking \cite{zhou2025focustrack} & ACM MM 2025 \\
\midrule
Hyperspherical consistency regularization \cite{tan2022hyperspherical} & CVPR 2022 \\
Simvp: Simpler yet better video prediction \cite{gao2022simvp} & CVPR 2022 \\
Temporal attention unit: Towards efficient spatiotemporal predictive learning \cite{tan2023temporal} & CVPR 2023 \\
Cvt-slr: Contrastive visual-textual transformation for sign language recognition with variational alignment \cite{zheng2023cvt} & CVPR 2023 \\
General point model with autoencoding and autoregressive \cite{li2023general} & CVPR 2023 \\
\bottomrule
\end{tabular}
}
\end{table*}
\clearpage
\begin{table*}[ht]
\centering
\caption{Benchmark papers organized by conference (cont.).}
\label{tab:benchmark3}
\resizebox{\textwidth}{!}{
\begin{tabular}{p{11cm}p{3cm}}
\toprule
\multicolumn{1}{c}{\textbf{Paper}} & \textbf{Conference} \\
\midrule
\fontsize{9}{11}\selectfont
Mlip: Enhancing medical visual representation with divergence encoder and knowledge-guided contrastive learning \cite{li2024mlip} & CVPR 2024 \\
Self-ensembling gaussian splatting for few-shot novel view synthesis \cite{zhao2025self} & ICCV\ \ 2025 \\
From words to structured visuals: A benchmark and framework for text-to-diagram generation and editing \cite{wei2025words} & CVPR 2025 \\
Context-aware multimodal pretraining \cite{roth2025context} & CVPR 2025 \\
Trajectorycrafter: Redirecting camera trajectory for monocular videos via diffusion models \cite{yu2025trajectorycrafter} & ICCV\ \ 2025 \\
Towards a unified copernicus foundation model for earth vision \cite{wang2025towards} & ICCV\ \ 2025 \\
Sparseflex: High-resolution and arbitrary-topology 3d shape modeling \cite{he2025sparseflex} & ICCV\ \ 2025 \\
Mergevq: A unified framework for visual generation and representation with disentangled token merging and quantization \cite{li2025mergevq} & CVPR 2025 \\
Back on track: Bundle adjustment for dynamic scene reconstruction \cite{chen2025back} & ICCV\ \ 2025 \\
Rayzer: A self-supervised large view synthesis model \cite{jiang2025rayzer} & ICCV\ \ 2025 \\
Token activation map to visually explain multimodal llms \cite{li2025token} & ICCV\ \ 2025 \\
LOTS of Fashion! multi-conditioning for image generation via sketch-text pairing \cite{girella2025lots} & ICCV\ \ 2025 \\
Geoint-r1: Formalizing multimodal geometric reasoning with dynamic auxiliary constructions \cite{wei2025geoint} & CVPR 2026 \\
LaRender: Training-Free Occlusion Control in Image Generation via Latent Rendering \cite{zhan2025larender} & ICCV\ \ 2025 \\
Forecasting continuous non-conservative dynamical systems in so (3) \cite{bastian2025forecasting} & ICCV\ \ 2025 \\
\midrule
Boosting the power of small multimodal reasoning models to match larger models with self-consistency training \cite{tan2024boosting} & ECCV 2024 \\
Weakly supervised 3d object detection via multi-level visual guidance \cite{huang2024weakly} & ECCV 2024 \\
Git: Towards generalist vision transformer through universal language interface \cite{wang2024git} & ECCV 2024 \\
Towards neuro-symbolic video understanding \cite{choi2024towards} & ECCV 2024 \\
Learning to generate conditional tri-plane for 3d-aware expression controllable portrait animation \cite{ki2024learning} & ECCV 2024 \\
Semgrasp: Semantic grasp generation via language aligned discretization \cite{li2024semgrasp} & ECCV 2024 \\
Brave: Broadening the visual encoding of vision-language models \cite{kar2024brave} & ECCV 2024 \\
Omniview-tuning: Boosting viewpoint invariance of vision-language pre-training models \cite{ruan2024omniview} & ECCV 2024 \\
SparseSSP: 3D subcellular structure prediction from sparse-view transmitted light images \cite{zheng2024sparsessp} & ECCV 2024 \\
HiT-SR: Hierarchical transformer for efficient image super-resolution \cite{zhang2024hit} & ECCV 2024 \\
4D contrastive superflows are dense 3D representation learners \cite{xu20244d} & ECCV 2024 \\
\bottomrule
\end{tabular}
}
\end{table*}
\clearpage
\begin{table*}[ht]
\centering
\caption{Benchmark papers organized by conference (cont.).}
\label{tab:benchmark4}
\resizebox{\textwidth}{!}{
\begin{tabular}{p{11cm}p{3cm}}
\toprule
\multicolumn{1}{c}{\textbf{Paper}} & \textbf{Conference} \\
\midrule
\fontsize{9}{11}\selectfont
Ittakestwo: Leveraging peer representations for semi-supervised lidar semantic segmentation \cite{liu2024ittakestwo} & ECCV 2024 \\
Beat-it: Beat-synchronized multi-condition 3d dance generation \cite{huang2024beat} & ECCV 2024 \\
Nl2contact: Natural language guided 3d hand-object contact modeling with diffusion model \cite{zhang2024nl2contact} & ECCV 2024 \\
Adversarial robustification via text-to-image diffusion models \cite{choi2024adversarial} & ECCV 2024 \\
Integer-valued training and spike-driven inference spiking neural network for high-performance and energy-efficient object detection \cite{luo2024integer} & ECCV 2024 \\
Risurconv: Rotation invariant surface attention-augmented convolutions for 3d point cloud classification and segmentation \cite{zhang2024risurconv} & ECCV 2024 \\
Making large language models better planners with reasoning-decision alignment \cite{huang2024making} & ECCV 2024 \\
Towards model-agnostic dataset condensation by heterogeneous models \cite{moon2024towards} & ECCV 2024 \\
Zero-shot detection of ai-generated images \cite{cozzolino2024zero} & ECCV 2024 \\
\midrule
Pifold: Toward effective and efficient protein inverse folding \cite{gao2022pifold} & ICLR 2022 \\
Moganet: Multi-order gated aggregation network \cite{li2022moganet} & ICLR 2022 \\
RDesign: Hierarchical data-efficient representation learning for tertiary structure-based RNA design \cite{tan2023rdesign} & ICLR 2023 \\
Knowledge-design: Pushing the limit of protein design via knowledge refinement \cite{gao2023knowledge} & ICLR 2023 \\
Semireward: A general reward model for semi-supervised learning \cite{li2023semireward} & ICLR 2023 \\
Decoupling weighing and selecting for integrating multiple graph pre-training tasks \cite{fan2024decoupling} & ICLR 2024 \\
CBGBench: Fill in the blank of protein-molecule complex binding graph \cite{lin2024cbgbench} & ICLR 2024 \\
Metoken: Uniform micro-environment token boosts post-translational modification prediction \cite{tan2024metoken} & ICLR 2025 \\
Half-order Fine-Tuning for Diffusion Model: A Recursive Likelihood Ratio Optimizer \cite{ren2025half} & ICLR 2025 \\
Generative human geometry distribution \cite{tang2025generative} & ICLR 2025 \\
Redteamcua: Realistic adversarial testing of computer-use agents in hybrid web-os environments \cite{liao2025redteamcua} & ICLR 2025 \\
MedAgentGym: A Scalable Agentic Training Environment for Code-Centric Reasoning in Biomedical Data Science \cite{xu2025medagentgym} & ICLR 2025 \\
TabStruct: Measuring Structural Fidelity of Tabular Data \cite{jiang2025tabstruct} & ICLR 2026 \\
Why DPO is a Misspecified Estimator and How to Fix It \cite{gopalan2025dpo} & ICLR 2025 \\
Temporal Sparse Autoencoders: Leveraging the Sequential Nature of Language for Interpretability \cite{bhalla2025temporal} & ICLR 2025 \\
Compositional Diffusion with Guided search for Long-Horizon Planning \cite{mishra2025compositional} & ICLR 2025 \\
Extending Sequence Length is Not All You Need: Effective Integration of Multimodal Signals for Gene Expression Prediction \cite{yang2026extending} & ICLR 2026 \\
RAIN-Merging: A Gradient-Free Method to Enhance Instruction Following in Large Reasoning Models with Preserved Thinking Format \cite{huang2026rain} & ICLR 2026 \\
\bottomrule
\end{tabular}
}
\end{table*}
\clearpage
\begin{table*}[ht]
\centering
\caption{Benchmark papers organized by conference (cont.).}
\label{tab:benchmark5}
\resizebox{\textwidth}{!}{
\begin{tabular}{p{11cm}p{3cm}}
\toprule
\multicolumn{1}{c}{\textbf{Paper}} & \textbf{Conference} \\
\midrule
\fontsize{9}{11}\selectfont
Through the Lens of Contrast: Self-Improving Visual Reasoning in VLMs \cite{pan2026through} & ICLR 2026 \\
Modality-free Graph In-context Alignment \cite{zhuo2026modality} & ICLR 2026 \\
\midrule
Deciphering RNA secondary structure prediction: A probabilistic k-rook matching perspective \cite{tan2022deciphering} & ICML 2024 \\
A graph is worth $ k $ words: Euclideanizing graph using pure transformer \cite{gao2024graph} & ICML 2024 \\
Re-Dock: towards flexible and realistic molecular docking with diffusion bridge \cite{huang2024re} & ICML 2024 \\
Vqdna: Unleashing the power of vector quantization for multi-species genomic sequence modeling \cite{li2024vqdna} & ICML 2024 \\
R\'enyi Neural Processes \cite{wang2024r} & ICML 2024 \\
A unified framework for entropy search and expected improvement in Bayesian optimization \cite{cheng2025unified} & ICML 2025 \\
Sundial: A family of highly capable time series foundation models \cite{liu2025sundial} & ICML 2025 \\
Lora-one: One-step full gradient could suffice for fine-tuning large language models, provably and efficiently \cite{zhang2025lora} & ICML 2025 \\
Stair: Improving safety alignment with introspective reasoning \cite{zhang2025stair} & ICML 2025 \\
Videojam: Joint appearance-motion representations for enhanced motion generation in video models \cite{chefer2025videojam} & ICML 2025 \\
Conceptattention: Diffusion transformers learn highly interpretable features \cite{helbling2025conceptattention} & ICML 2025 \\
Videorope: What makes for good video rotary position embedding? \cite{wei2025videorope} & ICML 2025 \\
Itbench: Evaluating ai agents across diverse real-world it automation tasks \cite{jha2025itbench} & ICML 2025 \\
Train for the worst, plan for the best: Understanding token ordering in masked diffusions \cite{kim2025train} & ICML 2025 \\
Embodiedbench: Comprehensive benchmarking multi-modal large language models for vision-driven embodied agents \cite{yang2025embodiedbench} & ICML 2025 \\
Near optimal decision trees in a SPLIT second \cite{babbar2025near} & ICML 2025 \\
Training a generally curious agent \cite{tajwar2025training} & ICML 2025 \\
Retrieval-augmented perception: High-resolution image perception meets visual rag \cite{wang2025retrieval} & ICML 2025 \\
$\text{MGD}^3$: Mode-Guided Dataset Distillation using Diffusion Models \cite{chan2025mgd} & ICML 2025 \\
Learning with expected signatures: Theory and applications \cite{lucchese2025learning} & ICML 2025 \\
\midrule
Self-supervised learning on graphs: Contrastive, generative, or predictive \cite{wu2021self} & IEEE TPAMI 2021 \\
A survey on generative diffusion models \cite{cao2024survey} & IEEE TKDE\ \ 2024 \\
Revisiting the temporal modeling in spatio-temporal predictive learning under a unified view \cite{tan2023revisiting} & IEEE TPAMI 2023 \\
DSGNN: A dual-view supergrid-aware graph neural network for regional air quality estimation \cite{zhang2024dsgnn} & IEEE TKDE\ \ 2024 \\
Micro-macro spatial-temporal graph-based encoder-decoder for map-constrained trajectory recovery \cite{wei2024micro} & IEEE TKDE\ \ 2024 \\
\bottomrule
\end{tabular}
}
\end{table*}
\clearpage
\begin{table*}[ht]
\centering
\caption{Benchmark papers organized by conference (cont.).}
\label{tab:benchmark6}
\resizebox{\textwidth}{!}{
\begin{tabular}{p{11cm}p{3cm}}
\toprule
\multicolumn{1}{c}{\textbf{Paper}} & \textbf{Conference} \\
\midrule
\fontsize{9}{11}\selectfont
Robgc: Towards robust graph condensation \cite{gao2025robgc} & IEEE TKDE\ \ 2025 \\
Temporal-Aware Spiking Transformer Hashing Based on 3D-DWT \cite{mei2025temporal} & IEEE TKDE\ \ 2025 \\
Enhanced multi-scale cross-attention for person image generation \cite{tang2025enhanced} & IEEE TPAMI 2025 \\
A causality-aware paradigm for evaluating creativity of multimodal large language models \cite{huang2025causality} & IEEE TPAMI 2025 \\
Revisiting gradient-based uncertainty for monocular depth estimation \cite{hornauer2025revisiting} & IEEE TPAMI 2025 \\
Recent advances in discrete speech tokens: A review \cite{guo2025recent} & IEEE TPAMI 2025 \\
Document-level tabular numerical cross-checking: A coarse-to-fine approach \cite{pang2025document} & IEEE TPAMI 2025 \\
AEFS: Adaptive Early Feature Selection for Deep Recommender Systems \cite{hu2025aefs} & IEEE TPAMI 2025 \\
High-utility Sequential Rule Mining Utilizing Segmentation Guided by Confidence \cite{zhang2026high} & IEEE TKDE\ \ 2026 \\
Training-free Graph-based Imputation of Missing Modalities in Multimodal Recommendation \cite{malitesta2026training} & IEEE TPAMI 2026 \\
Spatio-temporal Decoupled Knowledge Compensator for Few-Shot Action Recognition \cite{qu2026spatio} & IEEE TPAMI 2026 \\
PPC-MT: Parallel Point Cloud Completion with Mamba-Transformer Hybrid Architecture \cite{li2026ppc} & IEEE TPAMI 2026 \\
Benchmarking Semantic Segmentation Models via Appearance and Geometry Attribute Editing \cite{yin2026benchmarking} & IEEE TPAMI 2026 \\
Deep Tabular Representation Corrector \cite{ye2025deep} & IEEE TPAMI 2025 \\
STABLE: Efficient Hybrid Nearest Neighbor Search via Magnitude-Uniformity and Cardinality-Robustness \cite{yang2026stable} & IEEE TKDE\ \ 2026 \\
\midrule
Eve: Efficient zero-shot text-based video editing with depth map guidance and temporal consistency constraints \cite{chen2023eve} & IJCAI 2024 \\
Factchd: Benchmarking fact-conflicting hallucination detection \cite{chen2023factchd} & IJCAI 2024 \\
Imperio: Language-guided backdoor attacks for arbitrary model control \cite{chow2024imperio} & IJCAI 2024 \\
Bring metric functions into diffusion models \cite{an2024bring} & IJCAI 2024 \\
Bridging generative and discriminative models for unified visual perception with diffusion priors \cite{dong2024bridging} & IJCAI 2024 \\
Llmem: Estimating gpu memory usage for fine-tuning pre-trained llms \cite{kim2024llmem} & IJCAI 2024 \\
Group-aware coordination graph for multi-agent reinforcement learning \cite{duan2024group} & IJCAI 2024 \\
PEACH: Pretrained-embedding Explanation Across Contextual and Hierarchical Structure \cite{cao2024peach} & IJCAI 2024 \\
Sentence-level or token-level? a comprehensive study on knowledge distillation \cite{wei2024sentence} & IJCAI 2024 \\
MAS-SAM: Segment any marine animal with aggregated features \cite{yan2024mas} & IJCAI 2024 \\
Meta in-context learning makes large language models better zero and few-shot relation extractors \cite{li2024meta} & IJCAI 2024 \\
\bottomrule
\end{tabular}
}
\end{table*}
\clearpage
\begin{table*}[ht]
\centering
\caption{Benchmark papers organized by conference (cont.).}
\label{tab:benchmark7}
\resizebox{\textwidth}{!}{
\begin{tabular}{p{11cm}p{3cm}}
\toprule
\multicolumn{1}{c}{\textbf{Paper}} & \textbf{Conference} \\
\midrule
\fontsize{9}{11}\selectfont
MGCBS: an optimal and efficient algorithm for solving multi-goal multi-agent path finding problem \cite{tang2024mgcbs} & IJCAI 2024 \\
Fast one-stage unsupervised domain adaptive person search \cite{cui2024fast} & IJCAI 2024 \\
Interpretable network visualizations: A human-in-the-loop approach for post-hoc explainability of cnn-based image classification \cite{bianchi2024interpretable} & IJCAI 2024 \\
Boosting single positive multi-label classification with generalized robust loss \cite{chen2024boosting} & IJCAI 2024 \\
Prompt learning for generalized vehicle routing \cite{liu2024prompt} & IJCAI 2024 \\
Contrastive learning is not optimal for quasiperiodic time series \cite{atienza2024contrastive} & IJCAI 2024 \\
Marco: a memory-augmented reinforcement framework for combinatorial optimization \cite{garmendia2024marco} & IJCAI 2024 \\
Sketchagent: Generating structured diagrams from hand-drawn sketches \cite{tan2025sketchagent} & IJCAI 2025 \\
Multimodal regression for enzyme turnover rates prediction \cite{hu2025multimodal} & IJCAI 2025 \\
\midrule
EdgeSAM: Prompt-in-the-loop distillation for SAM \cite{zhou2025edgesam} & IJCV 2025 \\
Indoor obstacle discovery on reflective ground via monocular camera \cite{xue2024indoor} & IJCV 2024 \\
Cyclic refiner: Object-aware temporal representation learning for multi-view 3d detection and tracking \cite{guo2024cyclic} & IJCV 2024 \\
From easy to hard: Learning curricular shape-aware features for robust panoptic scene graph generation \cite{shi2025easy} & IJCV 2025 \\
AgMTR: Agent mining transformer for few-shot segmentation in remote sensing \cite{bi2025agmtr} & IJCV 2025 \\
One-shot generative domain adaptation in 3d gans \cite{li2025one} & IJCV 2025 \\
Calibrated cache model for few-shot vision-language model adaptation \cite{ding2024calibrated} & IJCV 2024 \\
Globally correlation-aware hard negative generation \cite{peng2025globally} & IJCV 2025 \\
Multimodal alignment and fusion: A survey \cite{li2024multimodal} & IJCV 2024 \\
Rethinking generalizability and discriminability of self-supervised learning from evolutionary game theory perspective \cite{li2025rethinking} & IJCV 2025 \\
Inspiring the next generation of segment anything models: Comprehensively evaluate sam and sam 2 with diverse prompts towards context-dependent concepts under different scenes \cite{zhao2024inspiring} & IJCV 2024 \\
Is Contrastive Distillation Enough for Learning Comprehensive 3D Representations? \cite{zhang2024contrastive} & IJCV 2024 \\
PointSea: point cloud completion via self-structure augmentation \cite{zhu2025pointsea} & IJCV 2025 \\
Creatively Upscaling Images with Global-Regional Priors \cite{qian2025creatively} & IJCV 2025 \\
Vision Generalist Model: A Survey: Z. Wang et al. \cite{wang2025vision} & IJCV 2025 \\
Feature Hallucination for Self-supervised Action Recognition: L. Wang, P. Koniusz \cite{wang2025feature} & IJCV 2025 \\
Unsupervised Robust Domain Adaptation: Paradigm, Theory and Algorithm: F. Huang etal \cite{huang2026unsupervised} & IJCV 2026 \\
Free Lunch to Meet the Gap: Intermediate Domain Reconstruction for Cross-Domain Few-Shot Learning \cite{zhang2025free} & IJCV 2025 \\
\bottomrule
\end{tabular}
}
\end{table*}
\clearpage
\begin{table*}[ht]
\centering
\caption{Benchmark papers organized by conference (cont.).}
\label{tab:benchmark8}
\resizebox{\textwidth}{!}{
\begin{tabular}{p{11cm}p{3cm}}
\toprule
\multicolumn{1}{c}{\textbf{Paper}} & \textbf{Conference} \\
\midrule
\fontsize{9}{11}\selectfont
Object-Scene-Camera Decomposition and Recomposition for Data Efficient Monocular 3D Object Detection \cite{kuang2026object} & IJCV 2026 \\
Neural Discrimination-Prompted Transformers for Efficient UHD Image Restoration and Enhancement: C. Wang et al. \cite{wang2026neural} & IJCV 2026 \\
\midrule
Harnessing hard mixed samples with decoupled regularizer \cite{liu2023harnessing} & NeurIPS 2023 \\
Openstl: A comprehensive benchmark of spatio-temporal predictive learning \cite{tan2023openstl} & NeurIPS 2023 \\
Uniif: Unified molecule inverse folding \cite{gao2024uniif} & NeurIPS 2024 \\
Large language diffusion models \cite{nie2025large} & NeurIPS 2025 \\
Pan-lut: Efficient pan-sharpening via learnable look-up tables \cite{cai2025pan} & NeurIPS 2025 \\
Mean flows for one-step generative modeling \cite{geng2025mean} & NeurIPS 2025 \\
Pass@ k policy optimization: Solving harder reinforcement learning problems \cite{walder2025pass} & NeurIPS 2025 \\
Web-shepherd: Advancing prms for reinforcing web agents \cite{chae2025web} & NeurIPS 2025 \\
Ai-researcher: Autonomous scientific innovation \cite{tang2025ai} & NeurIPS 2025 \\
Qoq-med: Building multimodal clinical foundation models with domain-aware grpo training \cite{dai2025qoq} & NeurIPS 2025 \\
Learning long range dependencies through time reversal symmetry breaking \cite{pourcel2025learning} & NeurIPS 2025 \\
Playerone: Egocentric world simulator \cite{tu2025playerone} & NeurIPS 2025 \\
Ambient diffusion omni: Training good models with bad data \cite{daras2025ambient} & NeurIPS 2025 \\
Trident: Tri-modal molecular representation learning with taxonomic annotations and local correspondence \cite{jiang2025trident} & NeurIPS 2025 \\
Checklists are better than reward models for aligning language models \cite{viswanathan2025checklists} & NeurIPS 2025 \\
DeltaFlow: An efficient multi-frame scene flow estimation method \cite{zhang2025deltaflow} & NeurIPS 2025 \\
ImageNet-trained CNNs are not biased towards texture: Revisiting feature reliance through controlled suppression \cite{burgert2025imagenet} & NeurIPS 2025 \\
Abstain Mask Retain Core: Time Series Prediction by Adaptive Masking Loss with Representation Consistency \cite{liang2025abstain} & NeurIPS 2025 \\
Infinitystar: Unified spacetime autoregressive modeling for visual generation \cite{liu2025infinitystar} & NeurIPS 2025 \\
Inner Speech as Behavior Guides: Steerable Imitation of Diverse Behaviors for Human-AI coordination \cite{trivedi2026inner} & NeurIPS 2026 \\
\bottomrule
\end{tabular}
}
\end{table*}

%% file: ref.bib
@String(arxiv = {arxiv})

@String(OpenReview = {OpenReview})

@String(NIPS = {NeurIPS})

@String(CVPR = {CVPR})

@String(ICCV = {ICCV})

@String(ECCV = {ECCV})

@String(ACL = {ACL})

@String(AAAI = {AAAI})

@String(ACMMM = {ACM MM})

@String(IJCV = {IJCV})

@String(AI = {AI})

@String(TKDE = {IEEE TKDE})

@String(TPAMI = {IEEE TPAMI})

@String(IEEE = {IEEE Access})

@misc{blecher_nougat_2023,
	title = {Nougat: {Neural} {Optical} {Understanding} for {Academic} {Documents}},
	shorttitle = {Nougat},
	url = {https://arxiv.org/abs/2308.13418v1},
	language = {en},
	urldate = {2026-04-11},
	journal = {arXiv.org},
	author = {Blecher, Lukas and Cucurull, Guillem and Scialom, Thomas and Stojnic, Robert},
	month = aug,
	year = {2023}
}

@misc{guo_seeing_2026,
	title = {Seeing is {Improving}: {Visual} {Feedback} for {Iterative} {Text} {Layout} {Refinement}},
	shorttitle = {Seeing is {Improving}},
	url = {https://arxiv.org/abs/2603.22187v1},
	language = {en},
	urldate = {2026-04-11},
	journal = {arXiv.org},
	author = {Guo, Junrong and Fang, Shancheng and Qu, Yadong and Xie, Hongtao},
	month = mar,
	year = {2026}
}

@misc{pang_paper2poster_2025,
	title = {{Paper2Poster}: {Towards} {Multimodal} {Poster} {Automation} from {Scientific} {Papers}},
	shorttitle = {{Paper2Poster}},
	url = {http://arxiv.org/abs/2505.21497},
	doi = {10.48550/arXiv.2505.21497},
	urldate = {2026-04-11},
	publisher = {arXiv},
	author = {Pang, Wei and Lin, Kevin Qinghong and Jian, Xiangru and He, Xi and Torr, Philip},
	month = oct,
	year = {2025},
	note = {arXiv:2505.21497 [cs]}
}

@misc{jiang_latte_2025,
	title = {{LATTE}: {Improving} {Latex} {Recognition} for {Tables} and {Formulae} with {Iterative} {Refinement}},
	shorttitle = {{LATTE}},
	url = {http://arxiv.org/abs/2409.14201},
	doi = {10.48550/arXiv.2409.14201},
	urldate = {2026-04-11},
	publisher = {arXiv},
	author = {Jiang, Nan and Liang, Shanchao and Wang, Chengxiao and Wang, Jiannan and Tan, Lin},
	month = feb,
	year = {2025},
	note = {arXiv:2409.14201 [cs]}
}

@misc{noauthor_flexdoc_nodate,
	title = {{FlexDoc}: {Flexible} {Document} {Adaptation} through {Optimizing} both {Content} and {Layout}},
	url = {https://arxiv.org/html/2410.15504v1},
	urldate = {2026-04-11}
}

@misc{li_vtlayout_2021,
	title = {{VTLayout}: {Fusion} of {Visual} and {Text} {Features} for {Document} {Layout} {Analysis}},
	shorttitle = {{VTLayout}},
	url = {http://arxiv.org/abs/2108.13297},
	doi = {10.48550/arXiv.2108.13297},
	urldate = {2026-04-11},
	publisher = {arXiv},
	author = {Li, Shoubin and Ma, Xuyan and Pan, Shuaiqun and Hu, Jun Rect and Shi, Lin and Wang, Qing},
	month = aug,
	year = {2021},
	note = {arXiv:2108.13297 [cs]}
}

@misc{li_relook_2025,
	title = {{ReLook}: {Vision}-{Grounded} {RL} with a {Multimodal} {LLM} {Critic} for {Agentic} {Web} {Coding}},
	shorttitle = {{ReLook}},
	url = {http://arxiv.org/abs/2510.11498},
	doi = {10.48550/arXiv.2510.11498},
	urldate = {2026-04-11},
	publisher = {arXiv},
	author = {Li, Yuhang and Zhang, Chenchen and Lv, Ruilin and Liu, Ao and Deng, Ken and Zhang, Yuanxing and Liu, Jiaheng and Zhou, Wiggin and Zhou, Bo},
	month = oct,
	year = {2025},
	note = {arXiv:2510.11498 [cs]}
}

@inproceedings{xu_layoutlm_2020,
	title = {{LayoutLM}: {Pre}-training of {Text} and {Layout} for {Document} {Image} {Understanding}},
	shorttitle = {{LayoutLM}},
	url = {http://arxiv.org/abs/1912.13318},
	doi = {10.1145/3394486.3403172},
	urldate = {2026-04-11},
	booktitle = {Proceedings of the 26th {ACM} {SIGKDD} {International} {Conference} on {Knowledge} {Discovery} \& {Data} {Mining}},
	author = {Xu, Yiheng and Li, Minghao and Cui, Lei and Huang, Shaohan and Wei, Furu and Zhou, Ming},
	month = aug,
	year = {2020},
	note = {arXiv:1912.13318 [cs]},
	pages = {1192--1200}
}

@misc{appalaraju_docformer_2021,
	title = {{DocFormer}: {End}-to-{End} {Transformer} for {Document} {Understanding}},
	shorttitle = {{DocFormer}},
	url = {http://arxiv.org/abs/2106.11539},
	doi = {10.48550/arXiv.2106.11539},
	urldate = {2026-04-11},
	publisher = {arXiv},
	author = {Appalaraju, Srikar and Jasani, Bhavan and Kota, Bhargava Urala and Xie, Yusheng and Manmatha, R.},
	month = sep,
	year = {2021},
	note = {arXiv:2106.11539 [cs]}
}

@misc{chen_rodla_2024,
	title = {{RoDLA}: {Benchmarking} the {Robustness} of {Document} {Layout} {Analysis} {Models}},
	copyright = {arXiv.org perpetual, non-exclusive license},
	shorttitle = {{RoDLA}},
	url = {https://arxiv.org/abs/2403.14442},
	doi = {10.48550/ARXIV.2403.14442},
	urldate = {2026-04-11},
	publisher = {arXiv},
	author = {Chen, Yufan and Zhang, Jiaming and Peng, Kunyu and Zheng, Junwei and Liu, Ruiping and Torr, Philip and Stiefelhagen, Rainer},
	year = {2024},
	note = {Version Number: 1}
}

@misc{saraiva_rxiv-maker_2025,
	title = {Rxiv-{Maker}: an automated template engine for streamlined scientific publications},
	copyright = {Creative Commons Attribution 4.0 International},
	shorttitle = {Rxiv-{Maker}},
	url = {https://arxiv.org/abs/2508.00836},
	doi = {10.48550/ARXIV.2508.00836},
	urldate = {2026-04-11},
	publisher = {arXiv},
	author = {Saraiva, Bruno M. and Brito, António D. and Jaquemet, Guillaume and Henriques, Ricardo},
	year = {2025},
	note = {Version Number: 5}
}

@misc{noauthor_2r2_nodate,
	title = {\${A}{\textasciicircum}{2R}{\textasciicircum}2\$: {Advancing} {Img2LaTeX} {Conversion} via {Visual} {Reasoning} with {Attention}-{Guided} {Refinement} {\textbar} {OpenReview}},
	url = {https://openreview.net/forum?id=UcrNxBtXWM},
	urldate = {2026-04-11}
}

@misc{ling_table2latex-rl_2025,
	title = {{Table2LaTeX}-{RL}: {High}-{Fidelity} {LaTeX} {Code} {Generation} from {Table} {Images} via {Reinforced} {Multimodal} {Language} {Models}},
	shorttitle = {{Table2LaTeX}-{RL}},
	url = {https://arxiv.org/abs/2509.17589v1},
	language = {en},
	urldate = {2026-04-11},
	journal = {arXiv.org},
	author = {Ling, Jun and Qi, Yao and Huang, Tao and Zhou, Shibo and Huang, Yanqin and Yang, Jiang and Song, Ziqi and Zhou, Ying and Yang, Yang and Shen, Heng Tao and Wang, Peng},
	month = sep,
	year = {2025}
}

@misc{noauthor_papertalker_nodate,
	title = {{PaperTalker} {Multi}-{Agent} {Framework}},
	url = {https://www.emergentmind.com/topics/papertalker-multi-agent-framework},
	urldate = {2026-04-11}
}

@misc{eatingchew_eric0801latexagent_2026,
	title = {Eric0801/{LaTeXAgent}},
	url = {https://github.com/Eric0801/LaTeXAgent},
	urldate = {2026-04-11},
	author = {EatingChew},
	month = feb,
	year = {2026},
	note = {original-date: 2026-01-25T03:50:39Z}
}

@inproceedings{lu_agent_2025,
	title = {Agent {Reviewers}: {Domain}-specific {Multimodal} {Agents} with {Shared} {Memory} for {Paper} {Review}},
	shorttitle = {Agent {Reviewers}},
	url = {https://openreview.net/forum?id=s7HUJamWqX},
	language = {en},
	urldate = {2026-04-11},
	author = {Lu, Kai and Xu, Shixiong and Li, Jinqiu and Ding, Kun and Meng, Gaofeng},
	month = jun,
	year = {2025}
}

@article{guo_visual_2025,
	title = {Visual {Feedback} for {Self}-{Improving} {Text} {Layout} with {MLLM} via {Reinforcement} {Learning}},
	url = {https://openreview.net/forum?id=wUYRMxrULV},
	language = {en},
	urldate = {2026-04-11},
	author = {Guo, Junrong and Fang, Shancheng and Qu, Yadong and Wang, Xiaorui and Xie, Hongtao},
	month = oct,
	year = {2025}
}

@misc{jain_simpledoc_2025,
	title = {{SimpleDoc}: {Multi}-{Modal} {Document} {Understanding} with {Dual}-{Cue} {Page} {Retrieval} and {Iterative} {Refinement}},
	copyright = {Creative Commons Attribution 4.0 International},
	shorttitle = {{SimpleDoc}},
	url = {https://arxiv.org/abs/2506.14035},
	doi = {10.48550/ARXIV.2506.14035},
	urldate = {2026-04-11},
	publisher = {arXiv},
	author = {Jain, Chelsi and Wu, Yiran and Zeng, Yifan and Liu, Jiale and Dai, S hengyu and Shao, Zhenwen and Wu, Qingyun and Wang, Huazheng},
	year = {2025},
	note = {Version Number: 1}
}

@misc{noauthor_omnilayout_nodate,
	title = {{OmniLayout}: {Enabling} {Coarse}-to-{Fine} {Learning} with {LLMs} for {Universal} {Document} {Layout} {Generation}},
	url = {https://arxiv.org/html/2510.26213v1},
	urldate = {2026-04-11}
}

@article{xu_layoutlm_2019,
	title = {{LayoutLM}: {Pre}-training of {Text} and {Layout} for {Document} {Image} {Understanding}},
	copyright = {arXiv.org perpetual, non-exclusive license},
	shorttitle = {{LayoutLM}},
	url = {https://arxiv.org/abs/1912.13318},
	doi = {10.48550/ARXIV.1912.13318},
	urldate = {2026-04-11},
	author = {Xu, Yiheng and Li, Minghao and Cui, Lei and Huang, Shaohan and Wei, Furu and Zhou, Ming},
	year = {2019},
	note = {Publisher: arXiv Version Number: 5}
}

@misc{noauthor_doclayout-yolo_nodate,
	title = {{DocLayout}-{YOLO}: {Enhancing} {Document} {Layout} {Analysis} through {Diverse} {Synthetic} {Data} and {Global}-to-{Local} {Adaptive} {Perception}},
	url = {https://arxiv.org/html/2410.12628v1},
	urldate = {2026-04-11}
}

@misc{kang_omnidoclayout_2025,
	title = {{OmniDocLayout}: {Towards} {Diverse} {Document} {Layout} {Generation} via {Coarse}-to-{Fine} {LLM} {Learning}},
	copyright = {arXiv.org perpetual, non-exclusive license},
	shorttitle = {{OmniDocLayout}},
	url = {https://arxiv.org/abs/2510.26213},
	doi = {10.48550/ARXIV.2510.26213},
	urldate = {2026-04-11},
	publisher = {arXiv},
	author = {Kang, Hengrui and Gu, Zhuangcheng and Zhao, Zhiyuan and Wen, Zichen and Wang, Bin and Li, Weijia and He, Conghui},
	year = {2025},
	note = {Version Number: 2}
}

@inproceedings{yadav_automated_2014,
	address = {Nagpur India},
	title = {Automated layout preservation in cross language translation of document: an integrated approach and implementation},
	isbn = {978-1-60558-814-8},
	shorttitle = {Automated layout preservation in cross language translation of document},
	url = {https://dl.acm.org/doi/10.1145/2675744.2675750},
	doi = {10.1145/2675744.2675750},
	language = {en},
	urldate = {2026-04-11},
	booktitle = {Proceedings of the 7th {ACM} {India} {Computing} {Conference}},
	publisher = {ACM},
	author = {Yadav, Vivek and Ramanathan, Chandrashekar},
	month = oct,
	year = {2014},
	pages = {1--8}
}

@misc{liu_docreward_2025,
	title = {{DocReward}: {A} {Document} {Reward} {Model} for {Structuring} and {Stylizing}},
	copyright = {arXiv.org perpetual, non-exclusive license},
	shorttitle = {{DocReward}},
	url = {https://arxiv.org/abs/2510.11391},
	doi = {10.48550/ARXIV.2510.11391},
	urldate = {2026-04-11},
	publisher = {arXiv},
	author = {Liu, Junpeng and Zhao, Yuzhong and Cao, Bowen and Ding, Jiayu and Jia, Yilin and Lv, Tengchao and Huang, Yupan and Huang, Shaohan and Yang, Nan and Dong, Li and Cui, Lei and Ge, Tao and Wang, Xun and Jiao, Huitian and Mao, Sun and Kartik, FNU and Chen, Si-Qing and Lam, Wai and Wei, Furu},
	year = {2025},
	note = {Version Number: 2}
}

@misc{duan_vision-rwkv_2024,
	title = {Vision-{RWKV}: {Efficient} and {Scalable} {Visual} {Perception} with {RWKV}-{Like} {Architectures}},
	shorttitle = {Vision-{RWKV}},
	url = {https://arxiv.org/abs/2403.02308v3},
	language = {en},
	urldate = {2026-04-11},
	journal = {arXiv.org},
	author = {Duan, Yuchen and Wang, Weiyun and Chen, Zhe and Zhu, Xizhou and Lu, Lewei and Lu, Tong and Qiao, Yu and Li, Hongsheng and Dai, Jifeng and Wang, Wenhai},
	month = mar,
	year = {2024}
}

@misc{kanervisto_im2latex-100k_2016,
	title = {{Im2Latex}-{100K} ,  {Arxiv}:1609.04938},
	copyright = {Creative Commons Zero - CC0 1.0, Open Access},
	shorttitle = {{Im2Latex}-{100K} ,  {Arxiv}},
	url = {https://zenodo.org/record/56198},
	doi = {10.5281/ZENODO.56198},
	urldate = {2026-04-11},
	publisher = {Zenodo},
	author = {Kanervisto, Anssi},
	month = jun,
	year = {2016}
}

@inproceedings{jayanth_monotone_2015,
	address = {Lisbon, Portugal},
	title = {Monotone {Submodularity} in {Opinion} {Summaries}},
	url = {http://aclweb.org/anthology/D15-1017},
	doi = {10.18653/v1/D15-1017},
	language = {en},
	urldate = {2026-04-11},
	booktitle = {Proceedings of the 2015 {Conference} on {Empirical} {Methods} in {Natural} {Language} {Processing}},
	publisher = {Association for Computational Linguistics},
	author = {Jayanth, Jayanth and Sundararaj, Jayaprakash and Bhattacharyya, Pushpak},
	year = {2015},
	pages = {169--178}
}

@article{hochreiter_long_1997,
	title = {Long {Short}-{Term} {Memory}},
	volume = {9},
	issn = {0899-7667, 1530-888X},
	url = {https://direct.mit.edu/neco/article/9/8/1735-1780/6109},
	doi = {10.1162/neco.1997.9.8.1735},
	language = {en},
	number = {8},
	urldate = {2026-04-11},
	journal = {Neural Computation},
	author = {Hochreiter, Sepp and Schmidhuber, Jürgen},
	month = nov,
	year = {1997},
	pages = {1735--1780}
}

@inproceedings{papineni_bleu_2001,
	address = {Philadelphia, Pennsylvania},
	title = {{BLEU}: a method for automatic evaluation of machine translation},
	shorttitle = {{BLEU}},
	url = {http://portal.acm.org/citation.cfm?doid=1073083.1073135},
	doi = {10.3115/1073083.1073135},
	language = {en},
	urldate = {2026-04-11},
	booktitle = {Proceedings of the 40th {Annual} {Meeting} on {Association} for {Computational} {Linguistics}  - {ACL} '02},
	publisher = {Association for Computational Linguistics},
	author = {Papineni, Kishore and Roukos, Salim and Ward, Todd and Zhu, Wei-Jing},
	year = {2001},
	pages = {311}
}

@article{zhu2025latextrans,
  title={LaTeXTrans: Structured LaTeX Translation with Multi-Agent Coordination},
  author={Zhu, Ziming and Wang, Chenglong and Xing, Shunjie and Huo, Yifu and Tian, Fengning and Du, Quan and Yang, Di and Zhang, Chunliang and Xiao, Tong and Zhu, Jingbo},
  journal=arxiv,
  year={2025}
}

@misc{math2latex2025,
  title={Math2LaTeX: Equation OCR to LaTeX with Qwen2-VL},
  author={Ovi, Sultanul and others},
  year={2025},
  howpublished={\url{https://github.com/sultanul-ovi/Math2LaTeX-Equation-OCR-to-LaTeX-with-Qwen2-VL}}
}

@inproceedings{kim2022donut,
  title     = {OCR-Free Document Understanding Transformer},
  author    = {Kim, Geewook and Hong, Teakgyu and Yim, Moonbin and Nam, JeongYeon and Park, Jinyoung and Yim, Jinyeong and Hwang, Wonseok and Yun, Sangdoo and Han, Dongyoon and Park, Seunghyun},
  booktitle = {European Conference on Computer Vision (ECCV)},
  year      = {2022}
}

@manual{pandoc,
  title = {Pandoc: a universal document converter},
  author = {MacFarlane, John},
  year = {2025},
  url = {https://pandoc.org/},
  note = {Version 3.x}
}

@misc{lin2026autofigureeditgeneratingeditablescientific,
      title={AutoFigure-Edit: Generating Editable Scientific Illustration}, 
      author={Zhen Lin and Qiujie Xie and Minjun Zhu and Shichen Li and Qiyao Sun and Enhao Gu and Yiran Ding and Ke Sun and Fang Guo and Panzhong Lu and Zhiyuan Ning and Yixuan Weng and Yue Zhang},
      year={2026},
      eprint={2603.06674},
      archivePrefix={arXiv},
      primaryClass={cs.CV},
      url={https://arxiv.org/abs/2603.06674}, 
}

@inproceedings{tan2024cross,
  title={Cross-gate mlp with protein complex invariant embedding is a one-shot antibody designer},
  author={Tan, Cheng and Gao, Zhangyang and Wu, Lirong and Xia, Jun and Zheng, Jiangbin and Yang, Xihong and Liu, Yue and Hu, Bozhen and Li, Stan Z},
  booktitle=AAAI,
  volume={38},
  pages={15222--15230},
  year={2024}
}

@inproceedings{wu2024psc,
  title={Psc-cpi: Multi-scale protein sequence-structure contrasting for efficient and generalizable compound-protein interaction prediction},
  author={Wu, Lirong and Huang, Yufei and Tan, Cheng and Gao, Zhangyang and Hu, Bozhen and Lin, Haitao and Liu, Zicheng and Li, Stan Z},
  booktitle=AAAI,
  volume={38},
  pages={310--319},
  year={2024}
}

@inproceedings{gao2025foldtoken,
  title={Foldtoken: Learning protein language via vector quantization and beyond},
  author={Gao, Zhangyang and Tan, Cheng and Wang, Jue and Huang, Yufei and Wu, Lirong and Li, Stan Z},
  booktitle=AAAI,
  volume={39},
  pages={219--227},
  year={2025}
}

@inproceedings{ahn2025isr,
  title={Isr-dpo: Aligning large multimodal models for videos by iterative self-retrospective dpo},
  author={Ahn, Daechul and Choi, Yura and Kim, San and Yu, Youngjae and Kang, Dongyeop and Choi, Jonghyun},
  booktitle=AAAI,
  volume={39},
  pages={1728--1736},
  year={2025}
}

@inproceedings{chen2025gim,
  title={Gim: A million-scale benchmark for generative image manipulation detection and localization},
  author={Chen, Yirui and Huang, Xudong and Zhang, Quan and Li, Wei and Zhu, Mingjian and Yan, Qiangyu and Li, Simiao and Chen, Hanting and Hu, Hailin and Yang, Jie and others},
  booktitle=AAAI,
  volume={39},
  pages={2311--2319},
  year={2025}
}

@inproceedings{chen2025enhancing,
  title={Enhancing adversarial transferability with adversarial weight tuning},
  author={Chen, Jiahao and Feng, Zhou and Zeng, Rui and Pu, Yuwen and Zhou, Chunyi and Jiang, Yi and Gan, Yuyou and Li, Jinbao and Ji, Shouling},
  booktitle=AAAI,
  volume={39},
  pages={2061--2069},
  year={2025}
}

@inproceedings{ling2025bias,
  title={Bias unveiled: Investigating social bias in LLM-generated code},
  author={Ling, Lin and Rabbi, Fazle and Wang, Song and Yang, Jinqiu},
  booktitle=AAAI,
  volume={39},
  pages={27491--27499},
  year={2025}
}

@inproceedings{aung2025multi,
  title={Multi-view pedestrian occupancy prediction with a novel synthetic dataset},
  author={Aung, Sithu and Sagong, Min-Cheol and Cho, Junghyun},
  booktitle=AAAI,
  volume={39},
  pages={1782--1790},
  year={2025}
}

@inproceedings{chen2025glcf,
  title={GLCF: A Global-Local Multimodal Coherence Analysis Framework for Talking Face Generation Detection},
  author={Chen, Xiaocan and Yin, Qilin and Liu, Jiarui and Lu, Wei and Luo, Xiangyang and Zhou, Jiantao},
  booktitle=AAAI,
  volume={39},
  pages={75--83},
  year={2025}
}

@inproceedings{li2025jailpo,
  title={JailPO: A Novel Black-box Jailbreak Framework via Preference Optimization against Aligned LLMs},
  author={Li, Hongyi and Ye, Jiawei and Wu, Jie and Yan, Tianjie and Wang, Chu and Li, Zhixin},
  booktitle=AAAI,
  volume={39},
  pages={27419--27427},
  year={2025}
}

@inproceedings{neustroev2025neural,
  title={Neural continuous-time supermartingale certificates},
  author={Neustroev, Grigory and Giacobbe, Mirco and Lukina, Anna},
  booktitle=AAAI,
  volume={39},
  pages={27538--27546},
  year={2025}
}

@inproceedings{ahn2025real,
  title={Real-time calibration model for low-cost sensor in fine-grained time series},
  author={Ahn, Seokho and Kim, Hyungjin and Shin, Sungbok and Seo, Young-Duk},
  booktitle=AAAI,
  volume={39},
  pages={3--11},
  year={2025}
}

@inproceedings{wu2025relation,
  title={Relation-aware equivariant graph networks for epitope-unknown antibody design and specificity optimization},
  author={Wu, Lirong and Lin, Haitao and Huang, Yufei and Gao, Zhangyang and Tan, Cheng and Liu, Yunfan and Wu, Tailin and Li, Stan Z},
  booktitle=AAAI,
  volume={39},
  pages={895--904},
  year={2025}
}

@inproceedings{chen2025causal,
  title={Causal-inspired multitask learning for video-based human pose estimation},
  author={Chen, Haipeng and Wu, Sifan and Wang, Zhigang and Yin, Yifang and Jiao, Yingying and Lyu, Yingda and Liu, Zhenguang},
  booktitle=AAAI,
  volume={39},
  pages={2052--2060},
  year={2025}
}

@inproceedings{tan2025dyab,
  title={dyab: Flow matching for flexible antibody design with alphafold-driven pre-binding antigen},
  author={Tan, Cheng and Zhang, Yijie and Gao, Zhangyang and Huang, Yufei and Lin, Haitao and Wu, Lirong and Wu, Fandi and Blanchette, Mathieu and Li, Stan Z},
  booktitle=AAAI,
  volume={39},
  pages={782--790},
  year={2025}
}

@inproceedings{chen2026unveiling,
  title={Unveiling the Landscape of Clinical Depression Assessment: From Behavioral Signatures to Psychiatric Reasoning},
  author={Chen, Zhuang and Bi, Guanqun and Zhang, Wen and Hu, Jiawei and Wang, Aoyun and Xiao, Xiyao and Feng, Kun and Huang, Minlie},
  booktitle=AAAI,
  volume={40},
  pages={1748--1756},
  year={2026}
}

@inproceedings{cao2026pite,
  title={PITE: Multi-Prototype Alignment for Individual Treatment Effect Estimation},
  author={Cao, Fuyuan and Zhang, Jiaxuan and Li, Xiaoli},
  booktitle=AAAI,
  volume={40},
  pages={19871--19879},
  year={2026}
}

@inproceedings{an2026fredn,
  title={FreDN: Spectral Disentanglement for Time Series Forecasting via Learnable Frequency Decomposition},
  author={An, Zhongde and You, Jinhong and Li, Jiyanglin and Tang, Yiming and Li, Wen and Du, Heming and Du, Shouguo},
  booktitle=AAAI,
  volume={40},
  pages={19623--19631},
  year={2026}
}

@inproceedings{deng2026nl2ca,
  title={NL2CA: Auto-formalizing Cognitive Decision-Making from Natural Language Using an Unsupervised CriticNL2LTL Framework},
  author={Deng, Zihao and Li, Yijia and Zhang, Renrui and Ye, Peijun},
  booktitle=AAAI,
  volume={40},
  pages={1766--1773},
  year={2026}
}

@inproceedings{cao2026spiking,
  title={Spiking Heterogeneous Graph Attention Networks},
  author={Cao, Buqing and Peng, Qian and Xie, Xiang and Chen, Liang and Shi, Min and Liu, Jianxun},
  booktitle=AAAI,
  volume={40},
  pages={19853--19861},
  year={2026}
}

@inproceedings{tan2021co,
  title={Co-learning: Learning from noisy labels with self-supervision},
  author={Tan, Cheng and Xia, Jun and Wu, Lirong and Li, Stan Z},
  booktitle=ACMMM,
  pages={1405--1413},
  year={2021}
}

@inproceedings{yang2023convert,
  title={Convert: Contrastive graph clustering with reliable augmentation},
  author={Yang, Xihong and Tan, Cheng and Liu, Yue and Liang, Ke and Wang, Siwei and Zhou, Sihang and Xia, Jun and Li, Stan Z and Liu, Xinwang and Zhu, En},
  booktitle=ACMMM,
  pages={319--327},
  year={2023}
}

@inproceedings{xu2025chain,
  title={Chain-of-Cooking: Cooking Process Visualization via Bidirectional Chain-of-Thought Guidance},
  author={Xu, Mengling and Tao, Ming and Bao, Bing-Kun},
  booktitle=ACMMM,
  pages={9287--9295},
  year={2025}
}

@inproceedings{zhang2025capturing,
  title={Capturing more: Learning multi-domain representations for robust online handwriting verification},
  author={Zhang, Peirong and Ding, Kai and Jin, Lianwen},
  booktitle=ACMMM,
  pages={1471--1479},
  year={2025}
}

@inproceedings{zhang2025gather,
  title={Gather and trace: Rethinking video textvqa from an instance-oriented perspective},
  author={Zhang, Yan and Zeng, Gangyan and Wu, Daiqing and Shen, Huawen and Li, Binbin and Zhou, Yu and Ma, Can and Bi, Xiaojun},
  booktitle=ACMMM,
  pages={876--885},
  year={2025}
}

@inproceedings{lin2025audio,
  title={Audio Does Matter: Importance-Aware Multi-Granularity Fusion for Video Moment Retrieval},
  author={Lin, Junan and Liu, Daizong and Chen, Xianke and Qu, Xiaoye and Yang, Xun and Zhu, Jixiang and Zhang, Sanyuan and Dong, Jianfeng},
  booktitle=ACMMM,
  pages={6027--6036},
  year={2025}
}

@inproceedings{yan2025universally,
  title={Universally Unfiltered and Unseen: Input-Agnostic Multimodal Jailbreaks against Text-to-Image Model Safeguards},
  author={Yan, Song and Wei, Hui and Fei, Jinlong and Yang, Guoliang and Zhao, Zhengyu and Wang, Zheng},
  booktitle=ACMMM,
  pages={11279--11287},
  year={2025}
}

@inproceedings{xue2025ad,
  title={AD-AVSR: Asymmetric Dual-stream Enhancement for Robust Audio-Visual Speech Recognition},
  author={Xue, Junxiao and Liu, Xiaozhen and Wu, Xuecheng and Yin, Xinyi and Huang, Danlei and Yu, Fei},
  booktitle=ACMMM,
  pages={3--11},
  year={2025}
}

@inproceedings{mo2025advancing,
  title={Advancing 3D Scene Understanding with MV-ScanQA Multi-View Reasoning Evaluation and TripAlign Pre-training Dataset},
  author={Mo, Wentao and Chen, Qingchao and Peng, Yuxin and Huang, Siyuan and Liu, Yang},
  booktitle=ACMMM,
  pages={12973--12980},
  year={2025}
}

@inproceedings{wang2025dime,
  title={DIME-Net: A Dual-Illumination Adaptive Enhancement Network Based on Retinex and Mixture-of-Experts},
  author={Wang, Ziang and Wang, Xiaoqin and Wang, Dingyi and Li, Qiang and Qiao, Shushan},
  booktitle=ACMMM,
  pages={8184--8193},
  year={2025}
}

@inproceedings{zhang2025dual,
  title={Dual-Phase Playtime-guided Recommendation: Interest Intensity Exploration and Multimodal Random Walks},
  author={Zhang, Jingmao and Zhao, Zhiting and Lin, Yunqi and Ma, Jianghong and Wei, Tianjun and Zhang, Haijun and Zhang, Xiaofeng},
  booktitle=ACMMM,
  pages={6232--6241},
  year={2025}
}

@inproceedings{liu2025anchorsync,
  title={AnchorSync: Global Consistency Optimization for Long Video Editing},
  author={Liu, Zichi and Wang, Yinggui and Wei, Tao and Ma, Chao},
  booktitle=ACMMM,
  pages={4494--4503},
  year={2025}
}

@inproceedings{zhao2025investigating,
  title={Investigating Domain Gaps for Indoor 3D Object Detection},
  author={Zhao, Zijing and Xu, Zhu and Chen, Qingchao and Peng, Yuxin and Liu, Yang},
  booktitle=ACMMM,
  pages={13198--13205},
  year={2025}
}

@inproceedings{qin2025drawing2cad,
  title={Drawing2CAD: Sequence-to-Sequence Learning for CAD Generation from Vector Drawings},
  author={Qin, Feiwei and Lu, Shichao and Hou, Junhao and Wang, Changmiao and Fang, Meie and Liu, Ligang},
  booktitle=ACMMM,
  pages={10573--10582},
  year={2025}
}

@inproceedings{wu2024robust,
  title={Robust multimodal sentiment analysis of image-text pairs by distribution-based feature recovery and fusion},
  author={Wu, Daiqing and Yang, Dongbao and Zhou, Yu and Ma, Can},
  booktitle=ACMMM,
  pages={5780--5789},
  year={2024}
}

@inproceedings{chen2025hud,
  title={Hud: Hierarchical uncertainty-aware disambiguation network for composed video retrieval},
  author={Chen, Zhiwei and Hu, Yupeng and Li, Zixu and Fu, Zhiheng and Wen, Haokun and Guan, Weili},
  booktitle=ACMMM,
  pages={6143--6152},
  year={2025}
}

@inproceedings{tang2025omnigen,
  title={Omnigen: Unified multimodal sensor generation for autonomous driving},
  author={Tang, Tao and Ma, Enhui and Zhou, Xia and Wang, Letian and Yan, Tianyi and Zhang, Xueyang and Zhan, Kun and Jia, Peng and Lang, Xianpeng and Bian, Jia-Wang and others},
  booktitle=ACMMM,
  pages={9365--9374},
  year={2025}
}

@inproceedings{huang2024magicfight,
  title={Magicfight: Personalized martial arts combat video generation},
  author={Huang, Jiancheng and Yan, Mingfu and Chen, Songyan and Huang, Yi and Chen, Shifeng},
  booktitle=ACMMM,
  pages={10833--10842},
  year={2024}
}

@inproceedings{yu2025realhd,
  title={RealHD: A High-Quality Dataset for Robust Detection of State-of-the-Art AI-Generated Images},
  author={Yu, Hanzhe and Ye, Yun and Rong, Jintao and Xuan, Qi and Ma, Chen},
  booktitle=ACMMM,
  pages={11394--11403},
  year={2025}
}

@inproceedings{zhou2025focustrack,
  title={Focustrack: One-stage focus-and-suppress framework for 3d point cloud object tracking},
  author={Zhou, Sifan and Nie, Jiahao and Zhao, Ziyu and Cao, Yichao and Lu, Xiaobo},
  booktitle=ACMMM,
  pages={7366--7375},
  year={2025}
}

@inproceedings{tan2022hyperspherical,
  title={Hyperspherical consistency regularization},
  author={Tan, Cheng and Gao, Zhangyang and Wu, Lirong and Li, Siyuan and Li, Stan Z},
  booktitle=CVPR,
  pages={7244--7255},
  year={2022}
}

@inproceedings{gao2022simvp,
  title={Simvp: Simpler yet better video prediction},
  author={Gao, Zhangyang and Tan, Cheng and Wu, Lirong and Li, Stan Z},
  booktitle=CVPR,
  pages={3170--3180},
  year={2022}
}

@inproceedings{tan2023temporal,
  title={Temporal attention unit: Towards efficient spatiotemporal predictive learning},
  author={Tan, Cheng and Gao, Zhangyang and Wu, Lirong and Xu, Yongjie and Xia, Jun and Li, Siyuan and Li, Stan Z},
  booktitle=CVPR,
  pages={18770--18782},
  year={2023}
}

@inproceedings{zheng2023cvt,
  title={Cvt-slr: Contrastive visual-textual transformation for sign language recognition with variational alignment},
  author={Zheng, Jiangbin and Wang, Yile and Tan, Cheng and Li, Siyuan and Wang, Ge and Xia, Jun and Chen, Yidong and Li, Stan Z},
  booktitle=CVPR,
  pages={23141--23150},
  year={2023}
}

@article{li2023general,
  title={General point model with autoencoding and autoregressive},
  author={Li, Zhe and Gao, Zhangyang and Tan, Cheng and Li, Stan Z and Yang, Laurence T},
  journal=arxiv,
  year={2023}
}

@inproceedings{li2024mlip,
  title={Mlip: Enhancing medical visual representation with divergence encoder and knowledge-guided contrastive learning},
  author={Li, Zhe and Yang, Laurence T and Ren, Bocheng and Nie, Xin and Gao, Zhangyang and Tan, Cheng and Li, Stan Z},
  booktitle=CVPR,
  pages={11704--11714},
  year={2024}
}

@inproceedings{zhao2025self,
  title={Self-ensembling gaussian splatting for few-shot novel view synthesis},
  author={Zhao, Chen and Wang, Xuan and Zhang, Tong and Javed, Saqib and Salzmann, Mathieu},
  booktitle=ICCV,
  pages={4940--4950},
  year={2025}
}

@inproceedings{wei2025words,
  title={From words to structured visuals: A benchmark and framework for text-to-diagram generation and editing},
  author={Wei, Jingxuan and Tan, Cheng and Chen, Qi and Wu, Gaowei and Li, Siyuan and Gao, Zhangyang and Sun, Linzhuang and Yu, Bihui and Guo, Ruifeng},
  booktitle=CVPR,
  pages={13315--13325},
  year={2025}
}

@inproceedings{roth2025context,
  title={Context-aware multimodal pretraining},
  author={Roth, Karsten and Akata, Zeynep and Damen, Dima and Balazevic, Ivana and H{\'e}naff, Olivier J},
  booktitle=CVPR,
  pages={4267--4279},
  year={2025}
}

@inproceedings{yu2025trajectorycrafter,
  title={Trajectorycrafter: Redirecting camera trajectory for monocular videos via diffusion models},
  author={Yu, Mark and Hu, Wenbo and Xing, Jinbo and Shan, Ying},
  booktitle=ICCV,
  pages={100--111},
  year={2025}
}

@inproceedings{wang2025towards,
  title={Towards a unified copernicus foundation model for earth vision},
  author={Wang, Yi and Xiong, Zhitong and Liu, Chenying and Stewart, Adam J and Dujardin, Thomas and Bountos, Nikolaos Ioannis and Zavras, Angelos and Gerken, Franziska and Papoutsis, Ioannis and Leal-Taix{\'e}, Laura and others},
  booktitle=ICCV,
  pages={9888--9899},
  year={2025}
}

@inproceedings{he2025sparseflex,
  title={Sparseflex: High-resolution and arbitrary-topology 3d shape modeling},
  author={He, Xianglong and Zou, Zi-Xin and Chen, Chia-Hao and Guo, Yuan-Chen and Liang, Ding and Yuan, Chun and Ouyang, Wanli and Cao, Yan-Pei and Li, Yangguang},
  booktitle=ICCV,
  pages={14822--14833},
  year={2025}
}

@inproceedings{li2025mergevq,
  title={Mergevq: A unified framework for visual generation and representation with disentangled token merging and quantization},
  author={Li, Siyuan and Zhang, Luyuan and Wang, Zedong and Tian, Juanxi and Tan, Cheng and Liu, Zicheng and Yu, Chang and Xie, Qingsong and Lu, Haonan and Wang, Haoqian and others},
  booktitle=CVPR,
  pages={19713--19723},
  year={2025}
}

@inproceedings{chen2025back,
  title={Back on track: Bundle adjustment for dynamic scene reconstruction},
  author={Chen, Weirong and Zhang, Ganlin and Wimbauer, Felix and Wang, Rui and Araslanov, Nikita and Vedaldi, Andrea and Cremers, Daniel},
  booktitle=ICCV,
  pages={4951--4960},
  year={2025}
}

@inproceedings{jiang2025rayzer,
  title={Rayzer: A self-supervised large view synthesis model},
  author={Jiang, Hanwen and Tan, Hao and Wang, Peng and Jin, Haian and Zhao, Yue and Bi, Sai and Zhang, Kai and Luan, Fujun and Sunkavalli, Kalyan and Huang, Qixing and others},
  booktitle=ICCV,
  pages={4918--4929},
  year={2025}
}

@inproceedings{li2025token,
  title={Token activation map to visually explain multimodal llms},
  author={Li, Yi and Wang, Hualiang and Ding, Xinpeng and Wang, Haonan and Li, Xiaomeng},
  booktitle=ICCV,
  pages={48--58},
  year={2025}
}

@inproceedings{girella2025lots,
  title={LOTS of Fashion! multi-conditioning for image generation via sketch-text pairing},
  author={Girella, Federico and Talon, Davide and Liu, Ziyue and Ruan, Zanxi and Wang, Yiming and Cristani, Marco},
  booktitle=ICCV,
  pages={19711--19720},
  year={2025}
}

@article{wei2025geoint,
  title={Geoint-r1: Formalizing multimodal geometric reasoning with dynamic auxiliary constructions},
  author={Wei, Jingxuan and Jia, Caijun and Chen, Qi and He, Honghao and Sun, Linzhuang and He, Conghui and Wu, Lijun and Yu, Bihui and Tan, Cheng},
  journal=arxiv,
  year={2025}
}

@inproceedings{zhan2025larender,
  title={LaRender: Training-Free Occlusion Control in Image Generation via Latent Rendering},
  author={Zhan, Xiaohang and Liu, Dingming},
  booktitle=ICCV,
  pages={19679--19688},
  year={2025}
}

@inproceedings{bastian2025forecasting,
  title={Forecasting continuous non-conservative dynamical systems in so (3)},
  author={Bastian, Lennart and Rashed, Mohammad and Navab, Nassir and Birdal, Tolga},
  booktitle=ICCV,
  pages={14845--14855},
  year={2025}
}

@inproceedings{tan2024boosting,
  title={Boosting the power of small multimodal reasoning models to match larger models with self-consistency training},
  author={Tan, Cheng and Wei, Jingxuan and Gao, Zhangyang and Sun, Linzhuang and Li, Siyuan and Guo, Ruifeng and Yu, Bihui and Li, Stan Z},
  booktitle=ECCV,
  pages={305--322},
  year={2024},
  organization={Springer}
}

@inproceedings{huang2024weakly,
  title={Weakly supervised 3d object detection via multi-level visual guidance},
  author={Huang, Kuan-Chih and Tsai, Yi-Hsuan and Yang, Ming-Hsuan},
  booktitle=ECCV,
  pages={175--191},
  year={2024},
  organization={Springer}
}

@inproceedings{wang2024git,
  title={Git: Towards generalist vision transformer through universal language interface},
  author={Wang, Haiyang and Tang, Hao and Jiang, Li and Shi, Shaoshuai and Naeem, Muhammad Ferjad and Li, Hongsheng and Schiele, Bernt and Wang, Liwei},
  booktitle=ECCV,
  pages={55--73},
  year={2024},
  organization={Springer}
}

@inproceedings{choi2024towards,
  title={Towards neuro-symbolic video understanding},
  author={Choi, Minkyu and Goel, Harsh and Omama, Mohammad and Yang, Yunhao and Shah, Sahil and Chinchali, Sandeep},
  booktitle=ECCV,
  pages={220--236},
  year={2024},
  organization={Springer}
}

@inproceedings{ki2024learning,
  title={Learning to generate conditional tri-plane for 3d-aware expression controllable portrait animation},
  author={Ki, Taekyung and Min, Dongchan and Chae, Gyeongsu},
  booktitle=ECCV,
  pages={476--493},
  year={2024},
  organization={Springer}
}

@inproceedings{li2024semgrasp,
  title={Semgrasp: Semantic grasp generation via language aligned discretization},
  author={Li, Kailin and Wang, Jingbo and Yang, Lixin and Lu, Cewu and Dai, Bo},
  booktitle=ECCV,
  pages={109--127},
  year={2024},
  organization={Springer}
}

@inproceedings{kar2024brave,
  title={Brave: Broadening the visual encoding of vision-language models},
  author={Kar, O{\u{g}}uzhan Fatih and Tonioni, Alessio and Poklukar, Petra and Kulshrestha, Achin and Zamir, Amir and Tombari, Federico},
  booktitle=ECCV,
  pages={113--132},
  year={2024},
  organization={Springer}
}

@inproceedings{ruan2024omniview,
  title={Omniview-tuning: Boosting viewpoint invariance of vision-language pre-training models},
  author={Ruan, Shouwei and Dong, Yinpeng and Liu, Hanqing and Huang, Yao and Su, Hang and Wei, Xingxing},
  booktitle=ECCV,
  pages={309--327},
  year={2024},
  organization={Springer}
}

@inproceedings{zheng2024sparsessp,
  title={SparseSSP: 3D subcellular structure prediction from sparse-view transmitted light images},
  author={Zheng, Jintu and Ding, Yi and Liu, Qizhe and Chen, Yuehui and Cao, Yi and Hu, Ying and Wang, Zenan},
  booktitle=ECCV,
  pages={267--283},
  year={2024},
  organization={Springer}
}

@inproceedings{zhang2024hit,
  title={HiT-SR: Hierarchical transformer for efficient image super-resolution},
  author={Zhang, Xiang and Zhang, Yulun and Yu, Fisher},
  booktitle=ECCV,
  pages={483--500},
  year={2024},
  organization={Springer}
}

@inproceedings{xu20244d,
  title={4D contrastive superflows are dense 3D representation learners},
  author={Xu, Xiang and Kong, Lingdong and Shuai, Hui and Zhang, Wenwei and Pan, Liang and Chen, Kai and Liu, Ziwei and Liu, Qingshan},
  booktitle=ECCV,
  pages={58--80},
  year={2024},
  organization={Springer}
}

@inproceedings{liu2024ittakestwo,
  title={Ittakestwo: Leveraging peer representations for semi-supervised lidar semantic segmentation},
  author={Liu, Yuyuan and Chen, Yuanhong and Wang, Hu and Belagiannis, Vasileios and Reid, Ian and Carneiro, Gustavo},
  booktitle=ECCV,
  pages={81--99},
  year={2024},
  organization={Springer}
}

@inproceedings{huang2024beat,
  title={Beat-it: Beat-synchronized multi-condition 3d dance generation},
  author={Huang, Zikai and Xu, Xuemiao and Xu, Cheng and Zhang, Huaidong and Zheng, Chenxi and Qin, Jing and He, Shengfeng},
  booktitle=ECCV,
  pages={273--290},
  year={2024},
  organization={Springer}
}

@inproceedings{zhang2024nl2contact,
  title={Nl2contact: Natural language guided 3d hand-object contact modeling with diffusion model},
  author={Zhang, Zhongqun and Wang, Hengfei and Yu, Ziwei and Cheng, Yihua and Yao, Angela and Chang, Hyung Jin},
  booktitle=ECCV,
  pages={284--300},
  year={2024},
  organization={Springer}
}

@inproceedings{choi2024adversarial,
  title={Adversarial robustification via text-to-image diffusion models},
  author={Choi, Daewon and Jeong, Jongheon and Jang, Huiwon and Shin, Jinwoo},
  booktitle=ECCV,
  pages={158--177},
  year={2024},
  organization={Springer}
}

@inproceedings{luo2024integer,
  title={Integer-valued training and spike-driven inference spiking neural network for high-performance and energy-efficient object detection},
  author={Luo, Xinhao and Yao, Man and Chou, Yuhong and Xu, Bo and Li, Guoqi},
  booktitle=ECCV,
  pages={253--272},
  year={2024},
  organization={Springer}
}

@inproceedings{zhang2024risurconv,
  title={Risurconv: Rotation invariant surface attention-augmented convolutions for 3d point cloud classification and segmentation},
  author={Zhang, Zhiyuan and Yang, Licheng and Xiang, Zhiyu},
  booktitle=ECCV,
  pages={93--109},
  year={2024},
  organization={Springer}
}

@inproceedings{huang2024making,
  title={Making large language models better planners with reasoning-decision alignment},
  author={Huang, Zhijian and Tang, Tao and Chen, Shaoxiang and Lin, Sihao and Jie, Zequn and Ma, Lin and Wang, Guangrun and Liang, Xiaodan},
  booktitle=ECCV,
  pages={73--90},
  year={2024},
  organization={Springer}
}

@inproceedings{moon2024towards,
  title={Towards model-agnostic dataset condensation by heterogeneous models},
  author={Moon, Jun-Yeong and Kim, Jung Uk and Park, Gyeong-Moon},
  booktitle=ECCV,
  pages={234--250},
  year={2024},
  organization={Springer}
}

@inproceedings{cozzolino2024zero,
  title={Zero-shot detection of ai-generated images},
  author={Cozzolino, Davide and Poggi, Giovanni and Nie{\ss}ner, Matthias and Verdoliva, Luisa},
  booktitle=ECCV,
  pages={54--72},
  year={2024},
  organization={Springer}
}

@article{gao2022pifold,
  title={Pifold: Toward effective and efficient protein inverse folding},
  author={Gao, Zhangyang and Tan, Cheng and Chac{\'o}n, Pablo and Li, Stan Z},
  journal=arxiv,
  year={2022}
}

@article{li2022moganet,
  title={Moganet: Multi-order gated aggregation network},
  author={Li, Siyuan and Wang, Zedong and Liu, Zicheng and Tan, Cheng and Lin, Haitao and Wu, Di and Chen, Zhiyuan and Zheng, Jiangbin and Li, Stan Z},
  journal=arxiv,
  year={2022}
}

@article{tan2023rdesign,
  title={RDesign: Hierarchical data-efficient representation learning for tertiary structure-based RNA design},
  author={Tan, Cheng and Zhang, Yijie and Gao, Zhangyang and Hu, Bozhen and Li, Siyuan and Liu, Zicheng and Li, Stan Z},
  journal=arxiv,
  year={2023}
}

@article{gao2023knowledge,
  title={Knowledge-design: Pushing the limit of protein design via knowledge refinement},
  author={Gao, Zhangyang and Tan, Cheng and Li, Stan Z},
  journal=arxiv,
  year={2023}
}

@article{li2023semireward,
  title={Semireward: A general reward model for semi-supervised learning},
  author={Li, Siyuan and Jin, Weiyang and Wang, Zedong and Wu, Fang and Liu, Zicheng and Tan, Cheng and Li, Stan Z},
  journal=arxiv,
  year={2023}
}

@article{fan2024decoupling,
  title={Decoupling weighing and selecting for integrating multiple graph pre-training tasks},
  author={Fan, Tianyu and Wu, Lirong and Huang, Yufei and Lin, Haitao and Tan, Cheng and Gao, Zhangyang and Li, Stan Z},
  journal=arxiv,
  year={2024}
}

@article{lin2024cbgbench,
  title={CBGBench: Fill in the blank of protein-molecule complex binding graph},
  author={Lin, Haitao and Zhao, Guojiang and Zhang, Odin and Huang, Yufei and Wu, Lirong and Liu, Zicheng and Li, Siyuan and Tan, Cheng and Gao, Zhifeng and Li, Stan Z},
  journal=arxiv,
  year={2024}
}

@article{tan2024metoken,
  title={Metoken: Uniform micro-environment token boosts post-translational modification prediction},
  author={Tan, Cheng and Cao, Zhenxiao and Gao, Zhangyang and Wu, Lirong and Li, Siyuan and Huang, Yufei and Xia, Jun and Hu, Bozhen and Li, Stan Z},
  journal=arxiv,
  year={2024}
}

@article{ren2025half,
  title={Half-order Fine-Tuning for Diffusion Model: A Recursive Likelihood Ratio Optimizer},
  author={Ren, Tao and Zhang, Zishi and Jiang, Jingyang and Li, Zehao and Qin, Shentao and Zheng, Yi and Li, Guanghao and Sun, Qianyou and Li, Yan and Liang, Jiafeng and others},
  journal=arxiv,
  year={2025}
}

@article{tang2025generative,
  title={Generative human geometry distribution},
  author={Tang, Xiangjun and Zhang, Biao and Wonka, Peter},
  journal=arxiv,
  year={2025}
}

@article{liao2025redteamcua,
  title={Redteamcua: Realistic adversarial testing of computer-use agents in hybrid web-os environments},
  author={Liao, Zeyi and Jones, Jaylen and Jiang, Linxi and Ning, Yuting and Fosler-Lussier, Eric and Su, Yu and Lin, Zhiqiang and Sun, Huan},
  journal=arxiv,
  year={2025}
}

@article{xu2025medagentgym,
  title={MedAgentGym: A Scalable Agentic Training Environment for Code-Centric Reasoning in Biomedical Data Science},
  author={Xu, Ran and Zhuang, Yuchen and Zhong, Yishan and Yu, Yue and Wang, Zifeng and Tang, Xiangru and Wu, Hang and Wang, May D and Ruan, Peifeng and Yang, Donghan and others},
  journal=arxiv,
  year={2025}
}

@article{jiang2025tabstruct,
  title={TabStruct: Measuring Structural Fidelity of Tabular Data},
  author={Jiang, Xiangjian and Simidjievski, Nikola and Jamnik, Mateja},
  journal=arxiv,
  year={2025}
}

@article{gopalan2025dpo,
  title={Why DPO is a Misspecified Estimator and How to Fix It},
  author={Gopalan, Aditya and Chowdhury, Sayak Ray and Banerjee, Debangshu},
  journal=arxiv,
  year={2025}
}

@article{bhalla2025temporal,
  title={Temporal Sparse Autoencoders: Leveraging the Sequential Nature of Language for Interpretability},
  author={Bhalla, Usha and Oesterling, Alex and Verdun, Claudio Mayrink and Lakkaraju, Himabindu and Calmon, Flavio P},
  journal=arxiv,
  year={2025}
}

@article{mishra2025compositional,
  title={Compositional Diffusion with Guided search for Long-Horizon Planning},
  author={Mishra, Utkarsh A and He, David and Chen, Yongxin and Xu, Danfei},
  journal=arxiv,
  year={2025}
}

@article{yang2026extending,
  title={Extending Sequence Length is Not All You Need: Effective Integration of Multimodal Signals for Gene Expression Prediction},
  author={Yang, Zhao and Duan, Yi and Zhu, Jiwei and Ba, Ying and Cao, Chuan and Su, Bing},
  journal=arxiv,
  year={2026}
}

@article{huang2026rain,
  title={RAIN-Merging: A Gradient-Free Method to Enhance Instruction Following in Large Reasoning Models with Preserved Thinking Format},
  author={Huang, Zhehao and Liu, Yuhang and Lin, Baijiong and Lou, Yixin and He, Zhengbao and Tian, Hanling and Li, Tao and Huang, Xiaolin},
  journal=arxiv,
  year={2026}
}

@article{pan2026through,
  title={Through the Lens of Contrast: Self-Improving Visual Reasoning in VLMs},
  author={Pan, Zhiyu and Wu, Yizheng and Hua, Jiashen and Feng, Junyi and Yan, Shaotian and Deng, Bing and Cao, Zhiguo and Ye, Jieping},
  journal=arxiv,
  year={2026}
}

@article{zhuo2026modality,
  title={Modality-free Graph In-context Alignment},
  author={Zhuo, Wei and Luo, Siqiang},
  journal=arxiv,
  year={2026}
}

@article{tan2022deciphering,
  title={Deciphering RNA secondary structure prediction: A probabilistic k-rook matching perspective},
  author={Tan, Cheng and Gao, Zhangyang and Cao, Hanqun and Chen, Xingran and Wang, Ge and Wu, Lirong and Xia, Jun and Zheng, Jiangbin and Li, Stan Z},
  journal=arxiv,
  year={2022}
}

@article{gao2024graph,
  title={A graph is worth $ k $ words: Euclideanizing graph using pure transformer},
  author={Gao, Zhangyang and Dong, Daize and Tan, Cheng and Xia, Jun and Hu, Bozhen and Li, Stan Z},
  journal=arxiv,
  year={2024}
}

@article{huang2024re,
  title={Re-Dock: towards flexible and realistic molecular docking with diffusion bridge},
  author={Huang, Yufei and Zhang, Odin and Wu, Lirong and Tan, Cheng and Lin, Haitao and Gao, Zhangyang and Li, Siyuan and Li, Stan and others},
  journal=arxiv,
  year={2024}
}

@article{li2024vqdna,
  title={Vqdna: Unleashing the power of vector quantization for multi-species genomic sequence modeling},
  author={Li, Siyuan and Wang, Zedong and Liu, Zicheng and Wu, Di and Tan, Cheng and Zheng, Jiangbin and Huang, Yufei and Li, Stan Z},
  journal=arxiv,
  year={2024}
}

@article{wang2024r,
  title={R{\'e}nyi Neural Processes},
  author={Wang, Xuesong and Zhao, He and Bonilla, Edwin V},
  journal=arxiv,
  year={2024}
}

@article{cheng2025unified,
  title={A unified framework for entropy search and expected improvement in Bayesian optimization},
  author={Cheng, Nuojin and Papenmeier, Leonard and Becker, Stephen and Nardi, Luigi},
  journal=arxiv,
  year={2025}
}

@article{liu2025sundial,
  title={Sundial: A family of highly capable time series foundation models},
  author={Liu, Yong and Qin, Guo and Shi, Zhiyuan and Chen, Zhi and Yang, Caiyin and Huang, Xiangdong and Wang, Jianmin and Long, Mingsheng},
  journal=arxiv,
  year={2025}
}

@article{zhang2025lora,
  title={Lora-one: One-step full gradient could suffice for fine-tuning large language models, provably and efficiently},
  author={Zhang, Yuanhe and Liu, Fanghui and Chen, Yudong},
  journal=arxiv,
  year={2025}
}

@article{zhang2025stair,
  title={Stair: Improving safety alignment with introspective reasoning},
  author={Zhang, Yichi and Zhang, Siyuan and Huang, Yao and Xia, Zeyu and Fang, Zhengwei and Yang, Xiao and Duan, Ranjie and Yan, Dong and Dong, Yinpeng and Zhu, Jun},
  journal=arxiv,
  year={2025}
}

@article{chefer2025videojam,
  title={Videojam: Joint appearance-motion representations for enhanced motion generation in video models},
  author={Chefer, Hila and Singer, Uriel and Zohar, Amit and Kirstain, Yuval and Polyak, Adam and Taigman, Yaniv and Wolf, Lior and Sheynin, Shelly},
  journal=arxiv,
  year={2025}
}

@article{helbling2025conceptattention,
  title={Conceptattention: Diffusion transformers learn highly interpretable features},
  author={Helbling, Alec and Meral, Tuna Han Salih and Hoover, Ben and Yanardag, Pinar and Chau, Duen Horng},
  journal=arxiv,
  year={2025}
}

@article{wei2025videorope,
  title={Videorope: What makes for good video rotary position embedding?},
  author={Wei, Xilin and Liu, Xiaoran and Zang, Yuhang and Dong, Xiaoyi and Zhang, Pan and Cao, Yuhang and Tong, Jian and Duan, Haodong and Guo, Qipeng and Wang, Jiaqi and others},
  journal=arxiv,
  year={2025}
}

@article{jha2025itbench,
  title={Itbench: Evaluating ai agents across diverse real-world it automation tasks},
  author={Jha, Saurabh and Arora, Rohan and Watanabe, Yuji and Yanagawa, Takumi and Chen, Yinfang and Clark, Jackson and Bhavya, Bhavya and Verma, Mudit and Kumar, Harshit and Kitahara, Hirokuni and others},
  journal=arxiv,
  year={2025}
}

@article{kim2025train,
  title={Train for the worst, plan for the best: Understanding token ordering in masked diffusions},
  author={Kim, Jaeyeon and Shah, Kulin and Kontonis, Vasilis and Kakade, Sham and Chen, Sitan},
  journal=arxiv,
  year={2025}
}

@article{yang2025embodiedbench,
  title={Embodiedbench: Comprehensive benchmarking multi-modal large language models for vision-driven embodied agents},
  author={Yang, Rui and Chen, Hanyang and Zhang, Junyu and Zhao, Mark and Qian, Cheng and Wang, Kangrui and Wang, Qineng and Koripella, Teja Venkat and Movahedi, Marziyeh and Li, Manling and others},
  journal=arxiv,
  year={2025}
}

@article{babbar2025near,
  title={Near optimal decision trees in a SPLIT second},
  author={Babbar, Varun and McTavish, Hayden and Rudin, Cynthia and Seltzer, Margo},
  journal=arxiv,
  year={2025}
}

@article{tajwar2025training,
  title={Training a generally curious agent},
  author={Tajwar, Fahim and Jiang, Yiding and Thankaraj, Abitha and Rahman, Sumaita Sadia and Kolter, J Zico and Schneider, Jeff and Salakhutdinov, Ruslan},
  journal=arxiv,
  year={2025}
}

@article{wang2025retrieval,
  title={Retrieval-augmented perception: High-resolution image perception meets visual rag},
  author={Wang, Wenbin and Jing, Yongcheng and Ding, Liang and Wang, Yingjie and Shen, Li and Luo, Yong and Du, Bo and Tao, Dacheng},
  journal=arxiv,
  year={2025}
}

@article{chan2025mgd,
  title={MGD$^3$: Mode-Guided Dataset Distillation using Diffusion Models},
  author={Chan-Santiago, Jeffrey A and Tirupattur, Praveen and Nayak, Gaurav Kumar and Liu, Gaowen and Shah, Mubarak},
  journal=arxiv,
  year={2025}
}

@article{lucchese2025learning,
  title={Learning with expected signatures: Theory and applications},
  author={Lucchese, Lorenzo and Pakkanen, Mikko S and Veraart, Almut ED},
  journal=arxiv,
  year={2025}
}

@article{wu2021self,
  title={Self-supervised learning on graphs: Contrastive, generative, or predictive},
  author={Wu, Lirong and Lin, Haitao and Tan, Cheng and Gao, Zhangyang and Li, Stan Z},
  journal=TKDE,
  volume={35},
  pages={4216--4235},
  year={2021},
  publisher={IEEE}
}

@article{cao2024survey,
  title={A survey on generative diffusion models},
  author={Cao, Hanqun and Tan, Cheng and Gao, Zhangyang and Xu, Yilun and Chen, Guangyong and Heng, Pheng-Ann and Li, Stan Z},
  journal=TKDE,
  volume={36},
  pages={2814--2830},
  year={2024},
  publisher={IEEE}
}

@article{tan2023revisiting,
  title={Revisiting the temporal modeling in spatio-temporal predictive learning under a unified view},
  author={Tan, Cheng and Wang, Jue and Gao, Zhangyang and Li, Siyuan and Wu, Lirong and Xia, Jun and Li, Stan Z},
  journal=arxiv,
  year={2023}
}

@article{zhang2024dsgnn,
  title={DSGNN: A dual-view supergrid-aware graph neural network for regional air quality estimation},
  author={Zhang, Xin and Chen, Ling and Tang, Xing and Shi, Hongyu},
  journal=arxiv,
  year={2024}
}

@article{wei2024micro,
  title={Micro-macro spatial-temporal graph-based encoder-decoder for map-constrained trajectory recovery},
  author={Wei, Tonglong and Lin, Youfang and Lin, Yan and Guo, Shengnan and Zhang, Lan and Wan, Huaiyu},
  journal=TKDE,
  volume={36},
  pages={6574--6587},
  year={2024},
  publisher={IEEE}
}

@article{gao2025robgc,
  title={Robgc: Towards robust graph condensation},
  author={Gao, Xinyi and Yin, Hongzhi and Chen, Tong and Ye, Guanhua and Zhang, Wentao and Cui, Bin},
  journal=TKDE,
  year={2025},
  publisher={IEEE}
}

@article{mei2025temporal,
  title={Temporal-Aware Spiking Transformer Hashing Based on 3D-DWT},
  author={Mei, Zihao and Li, Jianhao and Zhang, Bolin and Wang, Chong and Guo, Lijun and Li, Guoqi and Qian, Jiangbo},
  journal=arxiv,
  year={2025}
}

@article{tang2025enhanced,
  title={Enhanced multi-scale cross-attention for person image generation},
  author={Tang, Hao and Shao, Ling and Sebe, Nicu and Van Gool, Luc},
  journal=TPAMI,
  volume={47},
  pages={3377--3393},
  year={2025},
  publisher={IEEE}
}

@article{huang2025causality,
  title={A causality-aware paradigm for evaluating creativity of multimodal large language models},
  author={Huang, Zhongzhan and Zhong, Shanshan and Zhou, Pan and Gao, Shanghua and Zitnik, Marinka and Lin, Liang},
  journal=TPAMI,
  year={2025},
  publisher={IEEE}
}

@article{hornauer2025revisiting,
  title={Revisiting gradient-based uncertainty for monocular depth estimation},
  author={Hornauer, Julia and El-Ghoussani, Amir and Belagiannis, Vasileios},
  journal=TPAMI,
  year={2025},
  publisher={IEEE}
}

@article{guo2025recent,
  title={Recent advances in discrete speech tokens: A review},
  author={Guo, Yiwei and Li, Zhihan and Wang, Hankun and Li, Bohan and Shao, Chongtian and Zhang, Hanglei and Du, Chenpeng and Chen, Xie and Liu, Shujie and Yu, Kai},
  journal=TPAMI,
  year={2025},
  publisher={IEEE}
}

@article{pang2025document,
  title={Document-level tabular numerical cross-checking: A coarse-to-fine approach},
  author={Pang, Chaoxu and Cao, Yixuan and Zhou, Ganbin and Li, Hongwei and Luo, Ping},
  journal=arxiv,
  year={2025}
}

@article{hu2025aefs,
  title={AEFS: Adaptive Early Feature Selection for Deep Recommender Systems},
  author={Hu, Fan and Lu, Gaofeng and Chen, Jun and Guo, Channan and Yang, Yuekui and Li, Xirong},
  journal=TKDE,
  year={2025},
  publisher={IEEE}
}

@article{zhang2026high,
  title={High-utility Sequential Rule Mining Utilizing Segmentation Guided by Confidence},
  author={Zhang, Chunkai and Deng, Jiarui and Lyu, Maohua and Gan, Wensheng and Yu, Philip S},
  journal=arxiv,
  year={2026}
}

@article{malitesta2026training,
  title={Training-free Graph-based Imputation of Missing Modalities in Multimodal Recommendation},
  author={Malitesta, Daniele and Rossi, Emanuele and Pomo, Claudio and Di Noia, Tommaso and Malliaros, Fragkiskos D},
  journal=TKDE,
  year={2026},
  publisher={IEEE}
}

@article{qu2026spatio,
  title={Spatio-temporal Decoupled Knowledge Compensator for Few-Shot Action Recognition},
  author={Qu, Hongyu and Shu, Xiangbo and Yan, Rui and Gao, Hailiang and Wang, Wenguan and Tang, Jinhui},
  journal=TPAMI,
  year={2026},
  publisher={IEEE}
}

@article{li2026ppc,
  title={PPC-MT: Parallel Point Cloud Completion with Mamba-Transformer Hybrid Architecture},
  author={Li, Jie and Tian, Shengwei and Yu, Long and Ning, Xin},
  journal=arxiv,
  year={2026}
}

@article{yin2026benchmarking,
  title={Benchmarking Semantic Segmentation Models via Appearance and Geometry Attribute Editing},
  author={Yin, Zijin and Li, Bing and Liang, Kongming and Sun, Hao and He, Zhongjiang and Ma, Zhanyu and Guo, Jun},
  journal=TPAMI,
  year={2026},
  publisher={IEEE}
}

@article{ye2025deep,
  title={Deep Tabular Representation Corrector},
  author={Ye, Hangting and Wang, Peng and Fan, Wei and Song, Xiaozhuang and Zhao, He and Guo, Dandan and Chang, Yi},
  journal=TPAMI,
  year={2025},
  publisher={IEEE}
}

@article{yang2026stable,
  title={STABLE: Efficient Hybrid Nearest Neighbor Search via Magnitude-Uniformity and Cardinality-Robustness},
  author={Yang, Qianyun and Chen, Zhiwei and Hu, Yupeng and Li, Zixu and Fu, Zhiheng and Nie, Liqiang},
  journal=arxiv,
  year={2026}
}

@article{chen2023eve,
  title={Eve: Efficient zero-shot text-based video editing with depth map guidance and temporal consistency constraints},
  author={Chen, Yutao and Dong, Xingning and Gan, Tian and Zhou, Chunluan and Yang, Ming and Guo, Qingpei},
  journal=arxiv,
  year={2023}
}

@article{chen2023factchd,
  title={Factchd: Benchmarking fact-conflicting hallucination detection},
  author={Chen, Xiang and Song, Duanzheng and Gui, Honghao and Wang, Chenxi and Zhang, Ningyu and Jiang, Yong and Huang, Fei and Lv, Chengfei and Zhang, Dan and Chen, Huajun},
  journal=arxiv,
  year={2023}
}

@article{chow2024imperio,
  title={Imperio: Language-guided backdoor attacks for arbitrary model control},
  author={Chow, Ka-Ho and Wei, Wenqi and Yu, Lei},
  journal=arxiv,
  year={2024}
}

@article{an2024bring,
  title={Bring metric functions into diffusion models},
  author={An, Jie and Yang, Zhengyuan and Wang, Jianfeng and Li, Linjie and Liu, Zicheng and Wang, Lijuan and Luo, Jiebo},
  journal=arxiv,
  year={2024}
}

@article{dong2024bridging,
  title={Bridging generative and discriminative models for unified visual perception with diffusion priors},
  author={Dong, Shiyin and Zhu, Mingrui and Cheng, Kun and Wang, Nannan and Gao, Xinbo},
  journal=arxiv,
  year={2024}
}

@article{kim2024llmem,
  title={Llmem: Estimating gpu memory usage for fine-tuning pre-trained llms},
  author={Kim, Taeho and Wang, Yanming and Chaturvedi, Vatshank and Gupta, Lokesh and Kim, Seyeon and Kwon, Yongin and Ha, Sangtae},
  journal=arxiv,
  year={2024}
}

@article{duan2024group,
  title={Group-aware coordination graph for multi-agent reinforcement learning},
  author={Duan, Wei and Lu, Jie and Xuan, Junyu},
  journal=arxiv,
  year={2024}
}

@article{cao2024peach,
  title={PEACH: Pretrained-embedding Explanation Across Contextual and Hierarchical Structure},
  author={Cao, Feiqi and Han, Caren and Chung, Hyunsuk},
  journal=arxiv,
  year={2024}
}

@article{wei2024sentence,
  title={Sentence-level or token-level? a comprehensive study on knowledge distillation},
  author={Wei, Jingxuan and Sun, Linzhuang and Leng, Yichong and Tan, Xu and Yu, Bihui and Guo, Ruifeng},
  journal=arxiv,
  year={2024}
}

@article{yan2024mas,
  title={MAS-SAM: Segment any marine animal with aggregated features},
  author={Yan, Tianyu and Wan, Zifu and Deng, Xinhao and Zhang, Pingping and Liu, Yang and Lu, Huchuan},
  journal=arxiv,
  year={2024}
}

@article{li2024meta,
  title={Meta in-context learning makes large language models better zero and few-shot relation extractors},
  author={Li, Guozheng and Wang, Peng and Liu, Jiajun and Guo, Yikai and Ji, Ke and Shang, Ziyu and Xu, Zijie},
  journal=arxiv,
  year={2024}
}

@article{tang2024mgcbs,
  title={MGCBS: an optimal and efficient algorithm for solving multi-goal multi-agent path finding problem},
  author={Tang, Mingkai and Li, Yuanhang and Liu, Hongji and Chen, Yingbing and Liu, Ming and Wang, Lujia},
  journal=arxiv,
  year={2024}
}

@article{cui2024fast,
  title={Fast one-stage unsupervised domain adaptive person search},
  author={Cui, Tianxiang and Wang, Huibing and Peng, Jinjia and Deng, Ruoxi and Fu, Xianping and Wang, Yang},
  journal=arxiv,
  year={2024}
}

@article{bianchi2024interpretable,
  title={Interpretable network visualizations: A human-in-the-loop approach for post-hoc explainability of cnn-based image classification},
  author={Bianchi, Matteo and De Santis, Antonio and Tocchetti, Andrea and Brambilla, Marco},
  journal=arxiv,
  year={2024}
}

@article{chen2024boosting,
  title={Boosting single positive multi-label classification with generalized robust loss},
  author={Chen, Yanxi and Li, Chunxiao and Dai, Xinyang and Li, Jinhuan and Sun, Weiyu and Wang, Yiming and Zhang, Renyuan and Zhang, Tinghe and Wang, Bo},
  journal=arxiv,
  year={2024}
}

@article{liu2024prompt,
  title={Prompt learning for generalized vehicle routing},
  author={Liu, Fei and Lin, Xi and Liao, Weiduo and Wang, Zhenkun and Zhang, Qingfu and Tong, Xialiang and Yuan, Mingxuan},
  journal=arxiv,
  year={2024}
}

@article{atienza2024contrastive,
  title={Contrastive learning is not optimal for quasiperiodic time series},
  author={Atienza, Adrian and Bardram, Jakob and Puthusserypady, Sadasivan},
  journal=arxiv,
  year={2024}
}

@article{garmendia2024marco,
  title={Marco: a memory-augmented reinforcement framework for combinatorial optimization},
  author={Garmendia, Andoni I and Cappart, Quentin and Ceberio, Josu and Mendiburu, Alexander},
  journal=arxiv,
  year={2024}
}

@article{tan2025sketchagent,
  title={Sketchagent: Generating structured diagrams from hand-drawn sketches},
  author={Tan, Cheng and Chen, Qi and Wei, Jingxuan and Wu, Gaowei and Gao, Zhangyang and Li, Siyuan and Yu, Bihui and Guo, Ruifeng and Li, Stan Z},
  journal=arxiv,
  year={2025}
}

@article{hu2025multimodal,
  title={Multimodal regression for enzyme turnover rates prediction},
  author={Hu, Bozhen and Tan, Cheng and Li, Siyuan and Zheng, Jiangbin and Qiu, Sizhe and Xia, Jun and Li, Stan Z},
  journal=arxiv,
  year={2025}
}

@article{zhou2025edgesam,
  title={EdgeSAM: Prompt-in-the-loop distillation for SAM},
  author={Zhou, Chong and Li, Xiangtai and Loy, Chen Change and Dai, Bo},
  journal=IJCV,
  volume={133},
  pages={8452--8468},
  year={2025},
  publisher={Springer}
}

@article{xue2024indoor,
  title={Indoor obstacle discovery on reflective ground via monocular camera},
  author={Xue, Feng and Chang, Yicong and Wang, Tianxi and Zhou, Yu and Ming, Anlong},
  journal=IJCV,
  volume={132},
  pages={987--1007},
  year={2024},
  publisher={Springer}
}

@article{guo2024cyclic,
  title={Cyclic refiner: Object-aware temporal representation learning for multi-view 3d detection and tracking},
  author={Guo, Mingzhe and Zhang, Zhipeng and Jing, Liping and He, Yuan and Wang, Ke and Fan, Heng},
  journal=IJCV,
  volume={132},
  pages={6184--6206},
  year={2024},
  publisher={Springer}
}

@article{shi2025easy,
  title={From easy to hard: Learning curricular shape-aware features for robust panoptic scene graph generation},
  author={Shi, Hanrong and Li, Lin and Xiao, Jun and Zhuang, Yueting and Chen, Long},
  journal=IJCV,
  volume={133},
  pages={489--508},
  year={2025},
  publisher={Springer}
}

@article{bi2025agmtr,
  title={AgMTR: Agent mining transformer for few-shot segmentation in remote sensing},
  author={Bi, Hanbo and Feng, Yingchao and Mao, Yongqiang and Pei, Jianning and Diao, Wenhui and Wang, Hongqi and Sun, Xian},
  journal=IJCV,
  volume={133},
  pages={1780--1807},
  year={2025},
  publisher={Springer}
}

@article{li2025one,
  title={One-shot generative domain adaptation in 3d gans},
  author={Li, Ziqiang and Wu, Yi and Wang, Chaoyue and Rui, Xue and Li, Bin},
  journal=IJCV,
  volume={133},
  pages={2371--2391},
  year={2025},
  publisher={Springer}
}

@article{ding2024calibrated,
  title={Calibrated cache model for few-shot vision-language model adaptation},
  author={Ding, Kun and Yu, Qiang and Zhang, Haojian and Meng, Gaofeng and Xiang, Shiming},
  journal=arxiv,
  year={2024}
}

@article{peng2025globally,
  title={Globally correlation-aware hard negative generation},
  author={Peng, Wenjie and Huang, Hongxiang and Chen, Tianshui and Ke, Quhui and Dai, Gang and Huang, Shuangping},
  journal=IJCV,
  volume={133},
  pages={2441--2462},
  year={2025},
  publisher={Springer}
}

@article{li2024multimodal,
  title={Multimodal alignment and fusion: A survey},
  author={Li, Songtao and Tang, Hao},
  journal=arxiv,
  year={2024}
}

@article{li2025rethinking,
  title={Rethinking generalizability and discriminability of self-supervised learning from evolutionary game theory perspective},
  author={Li, Jiangmeng and Zang, Zehua and Ji, Qirui and Sun, Chuxiong and Qiang, Wenwen and Zhang, Junge and Zheng, Changwen and Sun, Fuchun and Xiong, Hui},
  journal=IJCV,
  volume={133},
  pages={3542--3567},
  year={2025},
  publisher={Springer}
}

@article{zhao2024inspiring,
  title={Inspiring the next generation of segment anything models: Comprehensively evaluate sam and sam 2 with diverse prompts towards context-dependent concepts under different scenes},
  author={Zhao, Xiaoqi and Pang, Youwei and Chang, Shijie and Zhao, Yuan and Zhang, Lihe and Yu, Chenyang and Liu, Hanqi and Zuo, Jiaming and Ouyang, Jinsong and Lin, Weisi and others},
  journal=arxiv,
  year={2024}
}

@article{zhang2024contrastive,
  title={Is Contrastive Distillation Enough for Learning Comprehensive 3D Representations?},
  author={Zhang, Yifan and Hou, Junhui},
  journal=arxiv,
  year={2024}
}

@article{zhu2025pointsea,
  title={PointSea: point cloud completion via self-structure augmentation},
  author={Zhu, Zhe and Chen, Honghua and He, Xing and Wei, Mingqiang},
  journal=IJCV,
  volume={133},
  pages={4770--4794},
  year={2025},
  publisher={Springer}
}

@article{qian2025creatively,
  title={Creatively Upscaling Images with Global-Regional Priors},
  author={Qian, Yurui and Cai, Qi and Pan, Yingwei and Yao, Ting and Mei, Tao},
  journal=IJCV,
  volume={133},
  pages={5197--5215},
  year={2025},
  publisher={Springer}
}

@article{wang2025vision,
  title={Vision Generalist Model: A Survey: Z. Wang et al.},
  author={Wang, Ziyi and Rao, Yongming and Sun, Shuofeng and Liu, Xinrun and Wei, Yi and Yu, Xumin and Liu, Zuyan and Wang, Yanbo and Liu, Hongmin and Zhou, Jie and others},
  journal=IJCV,
  volume={133},
  pages={6639--6667},
  year={2025},
  publisher={Springer}
}

@article{wang2025feature,
  title={Feature Hallucination for Self-supervised Action Recognition: L. Wang, P. Koniusz},
  author={Wang, Lei and Koniusz, Piotr},
  journal=IJCV,
  volume={133},
  pages={7612--7646},
  year={2025},
  publisher={Springer}
}

@article{huang2026unsupervised,
  title={Unsupervised Robust Domain Adaptation: Paradigm, Theory and Algorithm: F. Huang etal},
  author={Huang, Fuxiang and Fu, Xiaowei and Ye, Shiyu and Ma, Lina and Li, Wen and Gao, Xinbo and Zhang, David and Zhang, Lei},
  journal=IJCV,
  volume={134},
  pages={5},
  year={2026},
  publisher={Springer}
}

@article{zhang2025free,
  title={Free Lunch to Meet the Gap: Intermediate Domain Reconstruction for Cross-Domain Few-Shot Learning},
  author={Zhang, Tong and Zhao, Yifan and Wang, Liangyu and Li, Jia},
  journal=IJCV,
  volume={133},
  pages={5118--5137},
  year={2025},
  publisher={Springer}
}

@article{kuang2026object,
  title={Object-Scene-Camera Decomposition and Recomposition for Data Efficient Monocular 3D Object Detection},
  author={Kuang, Zhaonian and Ding, Rui and Yang, Meng and Zheng, Xinhu and Hua, Gang},
  journal=IJCV,
  volume={134},
  pages={155},
  year={2026},
  publisher={Springer}
}

@article{wang2026neural,
  title={Neural Discrimination-Prompted Transformers for Efficient UHD Image Restoration and Enhancement: C. Wang et al.},
  author={Wang, Cong and Pan, Jinshan and Wang, Liyan and Wang, Wei and Yang, Yang},
  journal=IJCV,
  volume={134},
  pages={84},
  year={2026},
  publisher={Springer}
}

@article{liu2023harnessing,
  title={Harnessing hard mixed samples with decoupled regularizer},
  author={Liu, Zicheng and Li, Siyuan and Wang, Ge and Wu, Lirong and Tan, Cheng and Li, Stan Z},
  journal=NIPS,
  volume={36},
  pages={52884--52906},
  year={2023}
}

@article{tan2023openstl,
  title={Openstl: A comprehensive benchmark of spatio-temporal predictive learning},
  author={Tan, Cheng and Li, Siyuan and Gao, Zhangyang and Guan, Wenfei and Wang, Zedong and Liu, Zicheng and Wu, Lirong and Li, Stan Z},
  journal=NIPS,
  volume={36},
  pages={69819--69831},
  year={2023}
}

@article{gao2024uniif,
  title={Uniif: Unified molecule inverse folding},
  author={Gao, Zhangyang and Wang, Jue and Tan, Cheng and Wu, Lirong and Huang, Yufei and Li, Siyuan and Ye, Zhirui and Li, Stan Z},
  journal=NIPS,
  volume={37},
  pages={135843--135860},
  year={2024}
}

@article{nie2025large,
  title={Large language diffusion models},
  author={Nie, Shen and Zhu, Fengqi and You, Zebin and Zhang, Xiaolu and Ou, Jingyang and Hu, Jun and Zhou, Jun and Lin, Yankai and Wen, Ji-Rong and Li, Chongxuan},
  journal=arxiv,
  year={2025}
}

@article{cai2025pan,
  title={Pan-lut: Efficient pan-sharpening via learnable look-up tables},
  author={Cai, Zhongnan and Wang, Yingying and Zheng, Hui and Pan, Panwang and Lin, ZiXu and Meng, Ge and Li, Chenxin and He, Chunming and Xie, Jiaxin and Lin, Yunlong and others},
  journal=arxiv,
  year={2025}
}

@article{geng2025mean,
  title={Mean flows for one-step generative modeling},
  author={Geng, Zhengyang and Deng, Mingyang and Bai, Xingjian and Kolter, J Zico and He, Kaiming},
  journal=arxiv,
  year={2025}
}

@article{walder2025pass,
  title={Pass@ k policy optimization: Solving harder reinforcement learning problems},
  author={Walder, Christian and Karkhanis, Deep},
  journal=arxiv,
  year={2025}
}

@article{chae2025web,
  title={Web-shepherd: Advancing prms for reinforcing web agents},
  author={Chae, Hyungjoo and Kim, Sunghwan and Cho, Junhee and Kim, Seungone and Moon, Seungjun and Hwangbo, Gyeom and Lim, Dongha and Kim, Minjin and Hwang, Yeonjun and Gwak, Minju and others},
  journal=arxiv,
  year={2025}
}

@article{tang2025ai,
  title={Ai-researcher: Autonomous scientific innovation},
  author={Tang, Jiabin and Xia, Lianghao and Li, Zhonghang and Huang, Chao},
  journal=arxiv,
  year={2025}
}

@article{dai2025qoq,
  title={Qoq-med: Building multimodal clinical foundation models with domain-aware grpo training},
  author={Dai, Wei and Chen, Peilin and Ekbote, Chanakya and Liang, Paul Pu},
  journal=arxiv,
  year={2025}
}

@article{pourcel2025learning,
  title={Learning long range dependencies through time reversal symmetry breaking},
  author={Pourcel, Guillaume and Ernoult, Maxence},
  journal=arxiv,
  year={2025}
}

@article{tu2025playerone,
  title={Playerone: Egocentric world simulator},
  author={Tu, Yuanpeng and Luo, Hao and Chen, Xi and Bai, Xiang and Wang, Fan and Zhao, Hengshuang},
  journal=arxiv,
  year={2025}
}

@article{daras2025ambient,
  title={Ambient diffusion omni: Training good models with bad data},
  author={Daras, Giannis and Rodriguez-Munoz, Adrian and Klivans, Adam and Torralba, Antonio and Daskalakis, Constantinos},
  journal=arxiv,
  year={2025}
}

@article{jiang2025trident,
  title={Trident: Tri-modal molecular representation learning with taxonomic annotations and local correspondence},
  author={Jiang, Feng and Prakash, Mangal and Ma, Hehuan and Deng, Jianyuan and Guo, Yuzhi and Mollaysa, Amina and Mansi, Tommaso and Liao, Rui and Huang, Junzhou},
  journal=arxiv,
  year={2025}
}

@article{viswanathan2025checklists,
  title={Checklists are better than reward models for aligning language models},
  author={Viswanathan, Vijay and Sun, Yanchao and Ma, Shuang and Kong, Xiang and Cao, Meng and Neubig, Graham and Wu, Tongshuang},
  journal=arxiv,
  year={2025}
}

@article{zhang2025deltaflow,
  title={DeltaFlow: An efficient multi-frame scene flow estimation method},
  author={Zhang, Qingwen and Zhu, Xiaomeng and Zhang, Yushan and Cai, Yixi and Andersson, Olov and Jensfelt, Patric},
  journal=arxiv,
  year={2025}
}

@article{burgert2025imagenet,
  title={ImageNet-trained CNNs are not biased towards texture: Revisiting feature reliance through controlled suppression},
  author={Burgert, Tom and Stoll, Oliver and Rota, Paolo and Demir, Beg{\"u}m},
  journal=arxiv,
  year={2025}
}

@article{liang2025abstain,
  title={Abstain Mask Retain Core: Time Series Prediction by Adaptive Masking Loss with Representation Consistency},
  author={Liang, Renzhao and Xu, Sizhe and Xie, Chenggang and Chen, Jingru and Ren, Feiyang and Yang, Shu and Yabe, Takahiro},
  journal=arxiv,
  year={2025}
}

@article{liu2025infinitystar,
  title={Infinitystar: Unified spacetime autoregressive modeling for visual generation},
  author={Liu, Jinlai and Han, Jian and Yan, Bin and Wu, Hui and Zhu, Fengda and Wang, Xing and Jiang, Yi and Peng, Bingyue and Yuan, Zehuan},
  journal=arxiv,
  year={2025}
}

@article{trivedi2026inner,
  title={Inner Speech as Behavior Guides: Steerable Imitation of Diverse Behaviors for Human-AI coordination},
  author={Trivedi, Rakshit and Sharma, Kartik and Parkes, David C},
  journal=arxiv,
  year={2026}
}

@article{wang_openhands_2024,
  title={{OpenHands}: An Open Platform for {AI} Software Developers as Generalist Agents},
  author={Wang, Xingyao and Li, Boxuan and Song, Yufan and Xu, Frank F. and Tang, Xiangru and Zhuge, Mingchen and Pan, Jiayi and Song, Yueqi and Li, Bowen and Singh, Jaskirat and Tran, Hoang H. and Li, Fuqiang and Ma, Ren and Zheng, Mingzhang and Qian, Bill and Shao, Yanjun and Muennighoff, Niklas and Zhang, Yizhe and Hui, Binyuan and Lin, Junyang and Brennan, Robert and Peng, Hao and Ji, Heng and Neubig, Graham},
  journal=arxiv,
  year={2024},
  url={https://arxiv.org/abs/2407.16741}
}

@misc{gauthier_aider_2023,
  title={Aider: {AI} Pair Programming in Your Terminal},
  author={Gauthier, Paul},
  year={2023},
  howpublished={\url{https://github.com/Aider-AI/aider}},
  note={Accessed: 2026-04-28}
}

@article{yang_swe-agent_2024,
  title={{SWE-agent}: Agent-Computer Interfaces Enable Automated Software Engineering},
  author={Yang, John and Jimenez, Carlos E. and Wettig, Alexander and Lieret, Kilian and Yao, Shunyu and Narasimhan, Karthik and Press, Ofir},
  journal=NIPS,
  volume={37},
  year={2024},
  url={https://arxiv.org/abs/2405.15793}
}

@article{wang_mineru_2024,
  title={{MinerU}: An Open-Source Solution for Precise Document Content Extraction},
  author={Wang, Bin and Xu, Chao and Zhao, Xiaomeng and Ouyang, Linke and Wu, Fan and Zhao, Zhiyuan and Xu, Rui and Liu, Kaiwen and Qu, Yuan and Shang, Fukai and Zhang, Bo and Wei, Liqun and Sui, Zhihao and Li, Wei and Shi, Botian and Qiao, Yu and Lin, Dahua and He, Conghui},
  journal=arxiv,
  year={2024},
  url={https://arxiv.org/abs/2409.18839}
}

@misc{datalab_marker_2024,
  title={Marker: Convert Documents to Markdown, {JSON}, Chunks, and {HTML}},
  author={{Datalab}},
  year={2024},
  howpublished={\url{https://github.com/datalab-to/marker}},
  note={Accessed: 2026-04-28}
}

@misc{blecher_latex-ocr_2022,
  title={{pix2tex}: Using a {ViT} to Convert Images of Equations into {LaTeX} Code},
  author={Blecher, Lukas},
  year={2022},
  howpublished={\url{https://github.com/lukas-blecher/LaTeX-OCR}},
  note={Accessed: 2026-04-28}
}

@book{mittelbach2004latex,
  title={The LATEX companion},
  author={Mittelbach, Frank and Goossens, Michel and Braams, Johannes and Carlisle, David and Rowley, Chris},
  year={2004},
  publisher={Addison-Wesley Professional}
}

@article{saraiva2025rxiv,
  title={Rxiv-Maker: an automated template engine for streamlined scientific publications},
  author={Saraiva, Bruno M and Brito, Ant{\'o}nio D and Jaquemet, Guillaume and Henriques, Ricardo},
  journal={arXiv preprint arXiv:2508.00836},
  year={2025}
}

@book{knuth1984texbook,
  title={The texbook},
  author={Knuth, Donald Ervin and Bibby, Duane},
  volume={15},
  year={1984},
  publisher={Addison-Wesley Reading}
}

@article{hurst2024gpt,
  title={Gpt-4o system card},
  author={Hurst, Aaron and Lerer, Adam and Goucher, Adam P and Perelman, Adam and Ramesh, Aditya and Clark, Aidan and Ostrow, AJ and Welihinda, Akila and Hayes, Alan and Radford, Alec and others},
  journal={arXiv preprint arXiv:2410.21276},
  year={2024}
}

@article{team2023gemini,
  title={Gemini: a family of highly capable multimodal models},
  author={Team, Gemini and Anil, Rohan and Borgeaud, Sebastian and Alayrac, Jean-Baptiste and Yu, Jiahui and Soricut, Radu and Schalkwyk, Johan and Dai, Andrew M and Hauth, Anja and Millican, Katie and others},
  journal={arXiv preprint arXiv:2312.11805},
  year={2023}
}

@article{yang2025qwen3,
  title={Qwen3 technical report},
  author={Yang, An and Li, Anfeng and Yang, Baosong and Zhang, Beichen and Hui, Binyuan and Zheng, Bo and Yu, Bowen and Gao, Chang and Huang, Chengen and Lv, Chenxu and others},
  journal={arXiv preprint arXiv:2505.09388},
  year={2025}
}

@article{madaan2023self,
  title={Self-refine: Iterative refinement with self-feedback},
  author={Madaan, Aman and Tandon, Niket and Gupta, Prakhar and Hallinan, Skyler and Gao, Luyu and Wiegreffe, Sarah and Alon, Uri and Dziri, Nouha and Prabhumoye, Shrimai and Yang, Yiming and others},
  journal={Advances in neural information processing systems},
  volume={36},
  pages={46534--46594},
  year={2023}
}

@article{shinn2023reflexion,
  title={Reflexion: Language agents with verbal reinforcement learning},
  author={Shinn, Noah and Cassano, Federico and Gopinath, Ashwin and Narasimhan, Karthik and Yao, Shunyu},
  journal={Advances in neural information processing systems},
  volume={36},
  pages={8634--8652},
  year={2023}
}

@article{lu2024ai,
  title={The ai scientist: Towards fully automated open-ended scientific discovery},
  author={Lu, Chris and Lu, Cong and Lange, Robert Tjarko and Foerster, Jakob and Clune, Jeff and Ha, David},
  journal={arXiv preprint arXiv:2408.06292},
  year={2024}
}

@article{weng2024cycleresearcher,
  title={Cycleresearcher: Improving automated research via automated review},
  author={Weng, Yixuan and Zhu, Minjun and Bao, Guangsheng and Zhang, Hongbo and Wang, Jindong and Zhang, Yue and Yang, Linyi},
  journal={arXiv preprint arXiv:2411.00816},
  year={2024}
}

@misc{openai_gpt54_2026,
  title={{GPT-5.4 Thinking System Card}},
  author={{OpenAI}},
  year={2026},
  howpublished={\url{https://openai.com/index/gpt-5-4-thinking-system-card}},
  note={Accessed: 2026-05-03}
}

@misc{anthropic_claude_opus_46_2026,
  title={{Claude Opus 4.6}},
  author={{Anthropic}},
  year={2026},
  howpublished={\url{https://www.anthropic.com/news/claude-opus-4-6}},
  note={Accessed: 2026-05-03}
}

@misc{deepseek_v4_pro_2026,
  title={{DeepSeek-V4-Pro Technical Report}},
  author={{DeepSeek-AI}},
  year={2026},
  howpublished={\url{https://huggingface.co/deepseek-ai/DeepSeek-V4-Pro/blob/main/DeepSeek_V4.pdf}},
  note={Accessed: 2026-05-03}
}

@misc{xiaomi_mimo_v25_pro_2026,
  title={{MiMo-V2.5-Pro Model Card}},
  author={{XiaomiMiMo}},
  year={2026},
  howpublished={\url{https://huggingface.co/XiaomiMiMo/MiMo-V2.5-Pro}},
  note={Accessed: 2026-05-03}
}
